\documentclass[journal]{IEEEtran}
\usepackage{amsmath,graphicx,amsthm,amssymb,bbm, subfigure, algorithm, bm, algorithmic, cite,xcolor,fontenc,subcaption,booktabs}
\usepackage{adjustbox}
\usepackage{hyperref}
\usepackage{textcomp}
\usepackage[normalem]{ulem}
\usepackage{cases}
\usepackage{hyperref,url}
\newtheorem{Theorem}{Theorem}
\newtheorem{Lemma}{Lemma}
\newtheorem{Assumption}{Assumption}
\theoremstyle{definition}

\newcommand\mc{\mathcal}
\newcommand\bd{\boldsymbol}

\newcommand{\cblack}[1]{\textcolor{black}{#1}}
\renewcommand{\cblack}[1]{\textcolor{black}{#1}}

\def\bx{\boldsymbol{x}}
\def\by{\boldsymbol{y}}
\def\bm{\boldsymbol{m}}

\begin{document}
\title{Accelerated Stochastic Min-Max Optimization Based on   Bias-corrected Momentum}
\author{Haoyuan Cai, Sulaiman A. Alghunaim, and Ali H.Sayed\\
\thanks{
Haoyuan Cai
and Ali H. Sayed 
are with the \'Ecole Polytechnique F\'ed\'erale de Lausanne, Switzerland
(mails: haoyuan.cai@epfl.ch, ali.sayed@epfl.ch)}
\thanks{Sulaiman A. Alghunaim
is with the Kuwait University, Kuwait
(email: sulaiman.alghunaim@ku.edu.kw)}
}
\renewcommand{\mc}{\mathcal}
\newcommand{\mbE}{\mathbb{E}}
\newcommand{\mbR}{\mathbb{R}}
\renewcommand{\bd}{\boldsymbol}
\newcommand{\mb}{\mathbb}

\newtheorem{Remark}{Remark}

\newtheorem{intassumption}{Assumption}
\numberwithin{intassumption}{Assumption}
\newtheorem{Corollary}{Corollary}
\allowdisplaybreaks
\maketitle

\begin{abstract}
Lower-bound analyses   
 for nonconvex strongly-concave minimax optimization problems
 have shown that stochastic 
 first-order algorithms 
require at least $\mathcal{O}(\varepsilon^{-4})$
sample complexity to find an $\varepsilon$-stationary point.
Some works 
indicate that
this complexity can be improved to 
$\mathcal{O}(\varepsilon^{-3})$
when the stochastic loss gradient is Lipschitz continuous.
The question 
of
achieving enhanced convergence rates under distinct conditions, remains open.
In this work, 
we address this question for optimization problems that are nonconvex in the minimization variable and strongly concave or Polyak-Lojasiewicz (PL) in the maximization variable.
We introduce novel bias-corrected momentum algorithms utilizing efficient Hessian-vector products. 
We establish convergence conditions and demonstrate a lower iteration complexity
of $\mathcal{O}(\varepsilon^{-3})$ for the proposed algorithms.
The effectiveness of the proposed method is validated through applications to robust logistic regression and robust adaptive cruise control.
\end{abstract}

\begin{IEEEkeywords}
Stochastic minimax optimization, 
\cblack{Polyak--Lojasiewicz conditions}, Hessian-vector product, bias-corrected momentum
\end{IEEEkeywords}

\section{Introduction}
\label{sec:introduction}
\IEEEPARstart{T}{he} study of minimization problems involves
identifying a model $w$
that minimizes a risk function $J(w)$ using only stochastic oracles:
\begin{align}
    \min_{w \in \mb{R}^K} J(w)&, \quad  
    \text{ where } J(w) = \mb{E}_{\bd{\zeta}} Q(w; \bd{\zeta}). 
\label{minimization}
\end{align}
Here, 
$\bd{\zeta}$ represents the training samples or streaming data in online learning, and  $Q(w; \bd{\zeta})$ denotes the stochastic loss function.
While this formulation is widely applicable,
many real-world scenarios involves minimizing over some variables while maximizing over other variables. For example, such scenarios appear in robust model predictive control \cite{gao2012explicit, copp2017simultaneous},
AUC maximization \cite{liu2019stochastic}, and reinforcement learning \cite{wen2021probably, pinto2017robust}.
To better address these cases, we study the
stochastic {\em minimax} optimization problem given by
\begin{subequations}
\begin{align}
    &\min_{x \in \mb{R}^{M_1}} \max_{y \in \mb{R}^{M_2}}
\ J(x, y) ,  \label{minimax1}
\\ &\text{ where } J(x, y) = \mb{E}_{\bd{\xi}} Q(x, y; \bd{\xi})  .\label{minimax2}
\end{align}
\end{subequations}
Here,
$\bd{\xi}$ represents the stochastic sample, $J(x,y)$ is the {\em risk} function, and 
$Q(x, y; \bd{\xi})$
is the stochastic {\em loss} function, with the $x$ and $y$ 
denoting
the optimization variables.
This formulation broadens the application scope of optimization methods. However,
solving a general minimax problem can be challenging.
Existing works focus on solving a subclass of this problem.
A common case is when
$J(x,y)$
is nonconvex in $x$ \cblack{but} strongly concave in $y$. In this context, \cite{lin2020gradient} showed that the two-time-scale
stochastic gradient descent-ascent (SGDA) \cblack{algorithm} can find an $\varepsilon$-stationary point.
This formulation is applicable to a wide range of applications, such as
 regularized Wasserstein GANs
 \cite{yang2022faster}
 and robust regression \cite{yan2019stochastic}, among others. 
For these reasons, we will study the stochastic minimax problem under a similar setting and aim to develop provably faster algorithms. Accelerated algorithms are important because they can improve iteration complexity and reduce training time.
Several strategies
have been explored in the literature 
to achieve this goal, 
with momentum methods and variance reduction techniques emerging as particularly prominent.
For instance, 
momentum-based techniques
such as Adam
\cite{kingma2014adam, sayed2022inference}
have become  
the de facto workhorse for training deep neural networks. 
This fact motivates
us to focus on momentum acceleration.
We next review the most closely related works.

\subsection{Related works}

For the minimization problem \eqref{minimization},
earlier studies 
have introduced accelerated methods such as
Polyak's heavy ball (PHB) \cite{polyak1964some}
and Nesterov's accelerated gradient (NAG)
\cite{nesterov2013introductory}
to enhance convergence rates in convex environments.
These methods have been widely investigated for smooth convex objectives \cite{nesterov2013introductory, sayed2022inference}.
Empirical evidence shows that momentum can improve deep learning performance; for example, RNNs trained with PHB or NAG outperform those using standard SGD \cite{sutskever2013importance}.
The success of momentum techniques has inspired extensive research into more effective methods. 
This has led to the development of adaptive momentum methods, such as
Adam \cite{kingma2014adam, sayed2022inference}, AdamW \cite{loshchilov2018decoupled},
AMSgrad \cite{sayed2022inference,reddi2019convergence}, Adan \cite{xie2022adan}.
Other studies \cblack{examined} the convergence of momentum methods under
\cblack{various scenarios (see \cite{li2024convergence,xin2019distributed,sayed2022inference} and references therein)}.

Despite the strong empirical performance of momentum methods, it has been shown that, for nonconvex objectives, stochastic first-order algorithms require at least $\mathcal{O}(\varepsilon^{-4})$ oracle complexity to find an $\varepsilon$-stationary point under a Lipschitz risk-gradient condition \cite{arjevani2023lower}.
Both SGD and stochastic momentum methods are known to achieve this rate of convergence
\cite{ghadimi2013stochastic, yang2016unified}.
Several works \cite{ liu2021noise,yang2016unified, kidambi2018insufficiency, liu2018accelerating} have demonstrated that the standard momentum methods do not necessarily yield theoretical improvements over standard SGD. 
These observations have motivated further investigation into whether more efficient approaches exist. For example, the work \cite{cutkosky2019momentum} proposed STORM, which achieves a lower oracle complexity of
$\mc{O}(\varepsilon^{-3})$ under stronger Lipschitz assumptions.
Adaptive extensions of STORM are subsequently explored in \cite{li2024convergence, levy2021storm}.
Nevertheless, STORM may not improve the convergence rate once its key Lipschitz assumption is relaxed. In particular, it critically relies on the Lipschitz continuity of the stochastic loss gradient
$\nabla_w Q(w;\bd{\zeta})$, an assumption that is not imposed in our analysis.
On the other hand, the works \cite{xie2022adan, cutkosky2020momentum} have shown that an oracle complexity of 
$\mathcal{O}(\varepsilon^{-3.5})$ can be attained under a Lipschitz continuous Hessian. This result can be further improved to 
$\mathcal{O}(\varepsilon^{-3})$ using a second-order momentum approach \cite{tran2022better}. Note that vanilla momentum does not benefit from these improvements, even under more relaxed Lipschitz conditions. Instead, techniques such as those in \cite{tran2022better,cutkosky2019momentum}, restarting schemes \cite{xie2022adan}, and past-gradient transport mechanisms \cite{cutkosky2020momentum} are crucial for achieving enhanced performance.

Many earlier works have focused on the minimization problem \eqref{minimization}. In recent years, however, increasing attention has been devoted to understanding its role in minimax optimization,  which is fundamentally more challenging than single-variable minimization settings.  Indeed, solving a general minimax problem can be intractable \cite{daskalakis2021complexity}, and convergence analyses are therefore typically conducted under specific structural assumptions on the risk function. Moreover, SGDA may suffer from large-batch requirements \cite{lin2020gradient}, which motivates the incorporation of momentum techniques to alleviate these theoretical limitations \cite{sharma2022federated}.
The work \cite{huang2022new} proposed a unified framework that integrates extragradient and momentum strategies for solving stochastic strongly monotone variational inequalities, achieving the optimal convergence rate.
Other works \cite{dou2021one, barazandeh2021solving}  proposed Adam-type algorithms for solving a class of nonmonotone minimax problems that satisfy the Minty variational inequality, obtaining a complexity of $\mathcal{O}(\varepsilon^{-4})$
with a batch size of $\mathcal{O}(\varepsilon^{-2})$.
Another work \cite{xian2021faster}
integrated 
the STORM momentum into  gradient descent ascent (GDA) 
to address nonconvex strongly-concave (SC) minimax problems,
establishing an oracle complexity of 
$\mathcal{O}(\varepsilon^{-3})$ with a mini-batch initialization.
Likewise, the work \cite{huang2023enhanced} relaxed the one-sided strong-concavity assumption to the Polyak-Lojasiewicz (PL) setting, demonstrating an oracle complexity of 
$\tilde{\mathcal{O}}(\varepsilon^{-3}
)$  with a  batch  size of $\mc{O}(1)$.
While similar convergence rates can be achieved using variance reduction techniques \cite{luo2020stochastic, yang2020global}, these methods can face \cblack{challenges} due to the need for periodically computing excessively large batch gradients.

To the best of our knowledge, under alternative assumptions such as Lipschitz continuous Hessians, an $\mathcal{O}(\varepsilon^{-3})$
convergence rate has not yet been established for stochastic nonconvex SC/PL minimax problems. This naturally leads to the following question:

\textbf{Q1:} {\em  
Can we design efficient stochastic minimax algorithms under a Lipschitz Hessian condition, and can such algorithms achieve a reduced convergence rate 
$\mathcal{O}(\varepsilon^{-3})$ with improved batch efficiency?}

In this work, we answer this question by introducing new bias-corrected momentum strategies.
 Our approach is inspired by \cite{tran2022better}, which leverages Hessian–vector products to construct a variance-reduced momentum estimator for minimization (rather than minimax) problems, leading to reduced variance and improved convergence rates. Related ideas have also been explored in stochastic compositional optimization \cite{chen2021solving}. However, these works focus exclusively on minimization setting and do not account for the coupling between primal and dual variables inherent in minimax optimization, which limits their applicability to several robust control and optimization applications considered in this work. In contrast, our method explicitly addresses this issue, requiring a substantial redesign of the algorithm to accommodate the new setting.

Unlike existing minimax analysis frameworks, our proposed algorithm poses substantial theoretical challenges due to several key factors. First, {\em nonlinearity} -- to achieve an improved convergence rate, we incorporate momentum clipping (or normalization) to control the error moments. 
This introduces nonlinearity into the effective learning rates, 
resulting in a mismatch error caused by nonlinear distortion. To address the challenge, we establish Lemma \ref{lemmaclipping} to carefully connect the error between the clipped momentum and the unclipped one.
Second, {\em error moments} -- as a consequence of momentum normalization, the standard second-order moment analysis tool is no longer directly applicable. 
 To address this challenge, we develop a distinct analytical tool based on a first-order error moment analysis.
Third, {\em hyperparameter selection} -- selecting hyperparameters to close the analysis is nontrivial. We address this challenge through a step-by-step construction of the hyperparameters in steps~\eqref{stepsize:choise:start}--\eqref{stepsize:choice:last}, ensuring that all required inequalities are satisfied simultaneously.
Fourth, the {\em PL condition} -- to the best of our knowledge, the PL condition has not been investigated in conjunction with the nonlinear momentum method of this work, thereby necessitating new theoretical techniques.
To overcome this difficulty, Lemma~\ref{dual_gap_ncpl} establishes a new bound on the optimality gap of the maximization variable.
\vspace{-0.5em}

\subsection{Contributions}
\begin{table*}[t]
\centering
\caption{Oracle complexity comparison of relevant stochastic minimax algorithms.
Here, $\varepsilon$ denotes the target solution accuracy, $n$ is the number of samples in the finite-sum setting, and $\kappa$ represents the ratio between the Lipschitz constant and the strong concavity/PL parameter. The order notation $\widetilde{\mathcal{O}}(\cdot)$ suppresses logarithmic factors.
We note that the Lipschitz risk gradient assumption is weaker than the (expected) Lipschitz loss gradient assumption.}
\label{tab:comparison}
\resizebox{\textwidth}{!}{
\begin{tabular}{lcccc}
\toprule
\textbf{Algorithm}  
& \textbf{Cost Structure}
& \textbf{Lipschitz Condition} 
& \textbf{Learning Rates} 
& \textbf{Sample Complexity} \\
\midrule
\cite{huang2022new} Extra-Momentum 
& Strongly Monotone
& Lipschitz risk gradient
& Constant
& $\mathcal{O}\!\left(\kappa \log\!\left(\frac{1}{\varepsilon}\right)\right)$ \\

\cite{yang2020global} VR-AGDA & PL-PL &  Lipschitz loss gradient &Constant &
$\mathcal{O}( \max \{(n+\kappa^9),  n +n^{\frac{2}{3}}\kappa^3\}\log(\frac{1}{\varepsilon}))$
\\

\cite{lin2020gradient} \, SGDA
& Nonconvex--SC
& Lipschitz risk gradient 
& Constant
& $\mathcal{O}(\varepsilon^{-4})$ \\

\cite{luo2020stochastic} SREDA
&Nonconvex-SC& Expected Lipschitz loss gradient&Constant& $\mathcal{O}(\varepsilon^{-3})$
\\

\cite{xian2021faster} DM-HSGD & Nonconvex-SC &
Lipschitz loss gradient & Constant & $\mathcal{O}(\varepsilon^{-3})$\\

\cite{yang2022faster} \, Stoc-Smoothed-AGDA
& Nonconvex--PL
& Lipschitz risk gradient
& Constant
& $\mathcal{O}(\varepsilon^{-4})$ \\

\cite{huang2023enhanced} MSGDA
& Nonconvex--PL
& Lipschitz loss gradient
& Diminishing
& $\widetilde{\mathcal{O}}(\varepsilon^{-3})$ 
\\
\textbf{HCMM-1} (\textbf{This Work})
&  Nonconvex--PL
& Lipschitz risk gradient and  Hessian
& Nonlinear via clipping
& $\mathcal{O}(\varepsilon^{-3})$ \\
\textbf{HCMM-2} (\textbf{This Work})
&  Nonconvex--PL
& Lipschitz risk gradient and  Hessian
& Nonlinear via normalization
& $\mathcal{O}(\varepsilon^{-3})$ \\
\bottomrule
\end{tabular}
}
\vspace{-1em}
\end{table*}

The contributions of our work are summarized as follows:

1) 
 We propose a novel momentum method for accelerating stochastic nonconvex-SC/PL minimax optimization under a Lipschitz Hessian condition. To the best of our knowledge, this momentum-based approach has not previously been investigated in the minimax optimization setting, unlike existing results for single-variable optimization. While inspired by the bias-correction techniques similar to \cite{tran2022better,chen2021solving}, our approach introduces a critical new feature to address the coupling between optimization variables in minimax settings. Compared with these minimization settings, our analysis is more challenging because it must simultaneously track the optimality gaps of both variables. The proposed minimax algorithms are particularly well-suited for applications with inherent robust or adversarial components, such as adversarial control \cite{gao2012explicit}, where traditional stochastic minimization algorithms are not directly applicable.

2) 
We develop new analyses by constructing a novel potential function and introducing a first-order moment analysis to handle the theoretical challenges. We show that the proposed algorithm achieves a reduced convergence rate
$\mc{O}(\varepsilon^{-3})$.
To highlight our contributions, we provide a comparison with existing methods in Table~\ref{tab:comparison}. As shown in Table~\ref{tab:comparison}, 
compared with accelerated algorithms (e.g., \cite{luo2020stochastic,xian2021faster,huang2023enhanced}),  our methods rely on a distinct set of assumptions and achieve the best-known convergence rate, demonstrating a unique advantage. In contrast to non-momentum methods, such as SGDA \cite{lin2020gradient} and stoc-smoothed-AGDA \cite{yang2022faster}, our approach improves the complexity by an order of  $\mathcal{O}(\varepsilon^{-1})$, which provides a significant speedup when seeking a high-accuracy solution.

3)
To demonstrate the practical value of the proposed methods, we evaluate them on two practically relevant applications: robust logistic regression and robust adaptive cruise control for a driving system. The results show that the proposed algorithms achieve superior empirical performance, further highlighting their practical effectiveness.


\emph{ Notation and preliminary assumptions:}
Lowercase letters (e.g., 
\(x\)) denote deterministic scalars or vectors, while boldface letters (e.g., \(\bd{x}\))
denote random variables.
The notation 
$\|\cdot\|$ represents the $\ell_2$-norm 
for vectors or the spectral norm for matrices, and
$\langle \cdot, \cdot \rangle$ represents the inner product.
For convenience, we denote the concatenated vector
 $z=\mbox{\rm cat}\{x,y\} \in \mb{R}^{M}$, \cblack{where} $M= {M_1}+M_2$ a\textcolor{black}{nd $x\in \mathbb{R}^{M_1}, y\in \mathbb{R}^{M_2}$}. Moreover, the notation
\begin{align*}
    \nabla_z J (x, y) &= [\nabla_x J (x, y);\nabla_y J (x, y)] \in \mb{R}^{M} 
\\
    \nabla^2_{zz} J (x, y) &= \begin{bmatrix}
 \nabla^2_{xx} J (x, y),\nabla^2_{xy} J (x, y)\\
 \nabla^2_{yx} J (x, y),
 \nabla^2_{yy} J (x, y)
\end{bmatrix} \in \mb{R}^{M\times M}
\end{align*}
denotes the true gradient and Hessian  of the risk function $J(x, y)$
relative to the concatenated variable $z$, respectively.
{\cblack   Their stochastic realizations are denoted by
$\nabla_z Q (x, y; \bd{\xi})
$ and $\nabla^2_{zz} Q (x, y; \bd{\xi})$.
Since the true gradient and Hessian are unavailable, we use stochastic approximations based on the loss value. They are assumed to be unbiased with bounded variance in expectation, which is standard in the stochastic optimization literature \cite{ghadimi2013stochastic}.
\begin{Assumption}[\textbf{Unbiased and bounded-variance constructions}] \label{unbiased}
We denote the $\sigma$-algebra generated by the random processes as
$\boldsymbol{\mathcal{F}}_{i} = \{\bd{z}_j \mid 
j =0, \dots, i \}$.
We assume
the  stochastic gradient evaluated at the block variable $\bd{z}_i = \mbox{\rm cat}\{\bd{x}_{i}, \bd{y}_{i}\}$ is unbiased  with bounded variance conditioned
on
$\boldsymbol{\mathcal{F}}_{i}$, i.e.,
  \begin{align}
 &  \mathbb{E}\Big\{ \nabla_{z} Q(\bd{x}_{i}, \bd{y}_{i};\boldsymbol{\xi}_{ i}) \mid       \boldsymbol{\mathcal{F}}_{i}\Big\} = \nabla_{z} J(\bd{x}_{i},\bd{y}_{i}) ,\notag \\
&      \mathbb{E}
      \Big\{\|\nabla_{z} Q(\bd{x}_{i}, \bd{y}_{i};\boldsymbol{\xi}_{i}) -  \nabla_{z} J(\bd{x}_{i}, \bd{y}_{i})\|^2 \mid \boldsymbol{\mathcal{F}}_{i} \Big\} \le {\sigma^2}  .
\end{align}
for some nonnegative constant $\sigma^2$. Likewise, the stochastic Hessian at location $\bd{z}_i$ is unbiased  with bounded variance conditioned
on
$\boldsymbol{\mathcal{F}}_{i}$, i.e.,
  \begin{align}
 &  \mathbb{E}\Big\{ \nabla^2_{zz} Q(\bd{x}_{i}, \bd{y}_{i};\boldsymbol{\xi}_{i}) \mid       \boldsymbol{\mathcal{F}}_{i}\Big\} = \nabla^2_{zz} J(\bd{x}_{i},\bd{y}_{i}) ,\notag \\
&      \mathbb{E}
      \Big\{\|\nabla^2_{zz} Q(\bd{x}_{ i}, \bd{y}_{i};\boldsymbol{\xi}_{i}) -  \nabla^2_{zz} J(\bd{x}_{i}, \bd{y}_{i})\|^2 \mid \boldsymbol{\mathcal{F}}_{i} \Big\} \le
      {\sigma^2_h}   .
    \label{boundHessian} 
\end{align}
where $\sigma_h^2$ is a nonnegative constant. Moreover,
we assume the data samples 
$\boldsymbol{\xi}_{i}$
are independent of each other 
for all  $i$.
\end{Assumption}
}
\section{Algorithm Development}
As shown in nonconvex optimization studies \cblack{(e.g., \cite{ghadimi2013stochastic})}, the performance bound of stochastic algorithms includes a deterministic term (from initialization) and a noisy term (from gradient variance). The step size affects both terms—large values reduce the deterministic part but amplify noise, and vice versa. An optimal step size needs to balance these components. Reducing the noise term enables using a larger step size and achieving faster convergence. Typical strategies to achieve this include variance reduction techniques that rely on large batch sizes to reduce variance, such as \cite{johnson2013accelerating}. Another approach is the use of variance-reduced momentum, which allows for obtaining a noise-reduced momentum with a batch size of
$\mathcal{O}(1)$ \cite{cutkosky2019momentum,tran2022better}.
To motivate our bias-corrected momentum, we first show that stochastic PHB is a biased gradient estimator, then introduce the correction technique to reduce this bias.

\vspace{-1em}
\subsection{Bias-corrected momentum}





The PHB method modifies gradient descent by using exponentially weighted averages of past gradients, helping stabilize the algorithm against sudden gradient fluctuations.
In practice, the PHB method employs a stochastic implementation \cite{sutskever2013importance, sayed2022inference}, which admits the following \cblack{form} at each iteration 
$i$:
\begin{subequations}
\begin{align}
\bd{m}_i &= (1-\beta) \bd{m}_{i-1}
    +\beta \nabla_w Q(\bd{w}_{i}; \bd{\zeta}_i) \label{HBmomentuma} \\
\bd{w}_{i+1} & = \bd{w}_{i}
    - \mu \bd{m}_i \label{HBmomentumb}
\end{align}
\end{subequations}
Here, $\bd{m}_i$ represents the momentum vector, 
$\bd{\zeta}_i$ denotes the stochastic sample, $\mu$ is the learning rate and 
$\beta$ is the smoothing factor. 
As previously discussed, the convergence rate of a stochastic algorithm is generally influenced by two components.
When the risk value $J(x,y)$ is directly accessible, the 
\cblack{gradient}
noise component vanishes, allowing the recovery of a rapidly decaying bound. Otherwise, a variance-reducing technique is important to improve the convergence rate.
Therefore, 
$\bd{m}_i$ is expected to estimate the risk gradient $\nabla_w J(\bd{w}_i)$ at point $\bd{w}_i$ as accurately as possible,  to approximate the performance of its deterministic counterpart. 
To achieve this,
$\bd{m}_i$ should be designed to be as accurate an estimator for the true gradient at that moment as possible.
The PHB momentum vector, which leverages a great number of past samples, is conceptually similar to variance reduction methods that utilize large-batch samples for computing the update direction. This similarity suggests that PHB momentum may help mitigate gradient noise. However, through mathematical induction, we \cblack{can verify} that the momentum vectors $\bd{m}_i, \bd{m}_{i-1}$ at any successive moments can not be 
unbiased simultaneously
in the transient regime.
Intuitively, 
if the \cblack{expected} momentum term
$\bd{m}_{i-1}$ \cblack{exactly matches}
$\nabla_w {J}(\bd{w}_{i-1})$,
such that $\mb{E}\bd{m}_{i-1} = \mb{E}\nabla_w {J}(\bd{w}_{i-1})$,
then it follows that 
\begin{align}
    \mb{E}\bd{m}_{i} &= (1 -\beta) \mb{E}\bd{m}_{i-1} 
    + \beta \mb{E} \nabla_w Q(\bd{w}_{i}; \bd{\zeta}_i)
      \notag \\
   &= (1-\beta) \mb{E} \nabla_w {J}(\bd{w}_{i-1})
    + \beta \mb{E} \nabla_w {J}(\bd{w}_{i}) \notag \\
   & \not =\mb{E}  \nabla_w {J}(\bd{w}_{i}) \label{biasedeffect}
\end{align}
Equation \eqref{biasedeffect} indicates that \cblack{an unbiased} $\bd{m}_{i-1}$ always results in a biased estimator 
$\bd{m}_{i}$. 
On the other hand, the
residual of this momentum term
\begin{align}
  \widetilde{\bd{m}}_i \triangleq \bd{m}_i - \nabla_w J(\bd{w}_{i}) 
\end{align}
could suffer from large variance. Subtracting $\nabla_w J(\bd{w}_{i})$ from both sides of 
\eqref{HBmomentuma} to get 
\begin{align}
\mbE\widetilde{\bd{m}}_i &=(1-\beta)\mbE\widetilde{\bd{m}}_{i-1}+(1-\beta)\mbE\big(\nabla_w J(\bd{w}_{i-1})-\nabla_w J(\bd{w}_{i})\big) \notag \\
 &\quad +\beta\mbE\big(\nabla_w Q(\bd{w}_i;\bd{\zeta}_i) - \nabla_w J(\bd{w}_{i})\big)
 \label{residual_term}
\end{align}
Assume that 
$\bd{m}_{i-1}$ has zero variance so that $\widetilde{\bd{m}}_{i-1}=0$.
Then,
 $\widetilde{\bd{m}}_i$
will be influenced by the last two terms, one of which can be controlled by selecting an appropriately small $\beta$.
However, $\widetilde{\bd{m}}_i$
still suffers from the deviation 
  $\nabla_w J(\bd{w}_{i-1}) -\nabla_w J(\bd{w}_{i})$
  that is accumulated over iterations.

These \cblack{bias} effects prevent the stochastic momentum 
\eqref{HBmomentuma}-\eqref{HBmomentumb} from consistently achieving an
accurate estimator for the true gradient.
To improve the quality of the momentum estimation, the works
\cite{cutkosky2019momentum, tran2022better} use a bias-correction approach.
Assuming $\bd{m}_{i-1}$ 
is \cblack{an unbiased} estimator for $\nabla_w J(\bd{w}_{i-1})$,
they propose subtracting the momentum term $\bd{m}_{i-1}$
from $\nabla_w J(\bd{w}_{i-1})$
and compensating  by adding the true gradient $\nabla_w J(\bd{w}_{i})$
 to ensure $\bd{m}_{i}$ remains unbiased
 \cblack{and close to the risk gradient}. This idea is described as follows:
\begin{align}
    \bd{m}_i = &(1 -\beta) [\bd{m}_{i-1}-
    \nabla_w J(\bd{w}_{i-1}) +
    \nabla_w J(\bd{w}_{i})]  \notag
    \\ &+\beta \nabla_w Q(\bd{w}_{i}; \bd{\zeta}_i) 
    \label{STORMder}
\end{align}
\cblack{By doing so and repeating the above argument, 
we have} 
\begin{align}
 \hspace{-2mm}   \mathbb{E} \bd{m}_i &= \mathbb{E} \nabla_w J(\bd{w}_i) \\
 \hspace{-2mm}  \mbE\widetilde{\bd{m}}_i
 &=(1-\beta)\mbE\widetilde{\bd{m}}_{i-1} +\beta\mbE\big(\nabla_w Q(\bd{w}_i;\bd{\zeta}_i) - \nabla_w J(\bd{w}_{i})\big)
 \label{improve_residual_term}
\end{align}
\cblack{Here, we observe that the large deviation terms have been removed in \eqref{improve_residual_term}, making the momentum term $\bd{m}_i$ more accurate, provided the preceding momentum  $\bd{m}_{i-1}$ is accurate}.
Note that the bias-correction term \cblack{satisfies}
\begin{align}
    &\nabla_w J(\bd{w}_{i-1})
    - \nabla_w J(\bd{w}_{i}) \notag \\ 
    &= 
    \nabla^2_w J(\bd{w}_{i})(\bd{w}_{i-1} - \bd{w}_{i})
    +\mc{O}(\|\bd{w}_i -\bd{w}_{i-1}\|^2)
\end{align}
The remainder can typically be managed by selecting an appropriate step size. Therefore,
omitting the higher-order terms, \cblack{expression} \eqref{STORMder}
can be approximated by using
\begin{align}
        \bd{m}_i =& (1 - \beta) [\bd{m}_{i-1} + \nabla^2_w
        J(\bd{w}_i) (\bd{w}_i - \bd{w}_{i-1})] 
     \notag\\ & + \beta \nabla_w Q(\bd{w}_{i}; \bd{\zeta}_i) 
     \label{SGDHessDer}
\end{align}
In a noisy environment,
we use the 
stochastic realization of 
$\nabla^2_w J(\bd{w}_i)$
to approximate the true \cblack{Hessian}. Hence, \cblack{relation} 
\eqref{SGDHessDer}
inspires the following recursion used in \cite{tran2022better}:
\begin{align}
    \bd{m}_i =& (1 - \beta) [\bd{m}_{i-1} + \nabla^2_w
        Q(\bd{w}_i; \bd{\zeta}_i) (\bd{w}_i - \bd{w}_{i-1})] 
     \notag\\ & + \beta \nabla_w Q(\bd{w}_{i}; \bd{\zeta}_i) 
     \label{SGDHessSto}
\end{align}
Using
\eqref{SGDHessSto} can improve the convergence rate of SGD
 from $\mathcal{O}(\varepsilon^{-4})$ to $\mathcal{O}(\varepsilon^{-3})$ with \cblack{a} Lipschitz Hessian \cite{tran2022better}.

In this work, we focus on
solving stochastic minimax optimization \cblack{problems}.
A natural question arises: can we leverage the benefits of \eqref{SGDHessSto} to achieve a faster convergence rate in the minimax optimization setting?
In the context of minimax optimization, we need to determine update directions for both
 $x$
and $y$.
A natural approach is to apply the same momentum recursion to both variables simultaneously, \cblack{say}
\begin{subequations}
\begin{align}
{\bd{m}}_{x,i} =&
 (1-\beta_x)[ \bd{m}_{x,i-1}
 +\nabla^2_{xx}Q(\bd{x}_i, \bd{y}_{i}; \bd{\xi}_i)
 (\bd{x}_{i}- \bd{x}_{i-1})]  \notag   \\
 & 
 + \beta_x \nabla_x Q(\bd{x}_i, \bd{y}_{i}; \boldsymbol{\xi}_{i}) \label{directmyupdate1} \\
{\bd{m}}_{y,i} =&
 (1-\beta_y)[ \bd{m}_{y,i-1}
 +\nabla^2_{yy}Q(\bd{x}_i, \bd{y}_{i}; \bd{\xi}_i)
 (\bd{y}_{i}- \bd{y}_{i-1})]  \notag \\
 & 
 + \beta_y \nabla_y Q(\bd{x}_i, \bd{y}_{i};
 \boldsymbol{\xi}_{i}) \label{directmyupdate2}
\end{align}
\end{subequations}
Here, $\bd{m}_{x,i}, \bd{m}_{y,i}$ are momentum vectors,
$\beta_x, \beta_y$
are smoothing factors relative to each variable. While \eqref{directmyupdate1}--\eqref{directmyupdate2} are straightforward for use, they serve as an {\em  inaccurate} approximation for the true gradient due to the missing information that couples
$x$ and $y$.
To demonstrate this,
we 
can use Taylor's expansion
for the block variable $z= \mbox{\rm cat}\{x, y\}$ of $J(x,y)$,
yielding
\begin{align}
 &\begin{bmatrix}
 \nabla^2_{xx} J (\bd{x}_{i}, \bd{y}_{i}),\nabla^2_{xy} J (\bd{x}_{i}, \bd{y}_{i})\\
 \nabla^2_{yx} J (\bd{x}_{i}, \bd{y}_{i}),
 \nabla^2_{yy} J (\bd{x}_{i}, \bd{y}_{i})
\end{bmatrix}
    \begin{bmatrix}
        \bd{x}_{i-1} - \bd{x}_{i}\\
        \bd{y}_{i-1}-\bd{y}_{i}
    \end{bmatrix}
    \label{approximateBias}  \\
    &=\nabla_z J(\bd{x}_{i-1}, \bd{y}_{i-1}) - 
     \nabla_z J(\bd{x}_{i}, \bd{y}_{i})
     +\mc{O}(\|\bd{z}_i -\bd{z}_{i-1}\|^2) \notag
\end{align}
It is clear that the cross-correlation terms
 have to be taken into consideration
 to obtain a more accurate approximation for $\nabla_z J(\bd{x}_{i}, \bd{y}_{i})-\nabla_z J(\bd{x}_{i-1}, \bd{y}_{i-1})$. 
Inspired by this \cblack{observation},
we propose the following stochastic recursion to update the momentum vectors:
\begin{subequations}
\begin{align}
{\bd{m}}_{x,i} =&
 (1-\beta_x)[ \bd{m}_{x,i-1}
 +\nabla^2_{xx}Q(\bd{x}_i, \bd{y}_{i};\bd{\xi}_i)
 (\bd{x}_{i}- \bd{x}_{i-1})  \notag   \\
   +& \nabla^2_{xy}Q(\bd{x}_i, \bd{y}_{i};\bd{\xi}_i)
 (\bd{y}_{i}- \bd{y}_{i-1})]
 + \beta_x \nabla_x Q(\bd{x}_i, \bd{y}_{i}; \boldsymbol{\xi}_{i}) \label{correctdirectmyupdate1} \\
{\bd{m}}_{y,i} =&
 (1-\beta_y)[ \bd{m}_{y,i-1}
 +\nabla^2_{yy}Q(\bd{x}_i, \bd{y}_{i};\bd{\xi}_i)
 (\bd{y}_{i}- \bd{y}_{i-1})  \notag \\
 + & \nabla^2_{yx}Q(\bd{x}_i, \bd{y}_{i};\bd{\xi}_i)
 (\bd{x}_{i}- \bd{x}_{i-1})]
 + \beta_y \nabla_y Q(\bd{x}_i, \bd{y}_{i};
 \boldsymbol{\xi}_{i}) \label{correctdirectmyupdate2}
\end{align}
\end{subequations}
The Hessian-vector products involved \cblack{in these relations} can be efficiently computed using a fast numerical approach such as \cite{pearlmutter1994fast}, without directly forming and storing the Hessian. Moreover,
this approach  has linear complexity in time and space
similar to that in querying a stochastic gradient.

Note that  the bias-correction term
$\nabla_w J(\bd{w}_{i-1})-\nabla_w J(\bd{w}_{i})$
 in \eqref{STORMder} can also be approximated by their stochastic realizations,  resulting in the STORM momentum \cite{cutkosky2019momentum}.
This strategy relies on the condition  $\|\nabla_w Q(w_1;\bd{\zeta}) -
 \nabla_w Q(w_2;\bd{\zeta})\| \le L_w\|w_1 -w_2\|$ for some constant $L_w$ to demonstrate  enhanced convergence.
As this condition is not the focus of our work, we refer readers to \cite{cutkosky2019momentum} for a detailed comparison.

In the next subsection, we introduce strategies based on
\eqref{correctdirectmyupdate1}--\eqref{correctdirectmyupdate2}
for
solving the stochastic minimax problem.

\subsection{Hessian corrected momentum methods}

\begin{algorithm}[t]
\caption{:
\textbf{H}essian \textbf{C}orrected 
\textbf{M}omentum
\textbf{M}ethod}
\label{example-algorithm}
\begin{algorithmic}[1]
\footnotesize
\renewcommand{\REQUIRE}{\item[\textbf{Initialize:}]}
\REQUIRE $\boldsymbol{x}_{0}, 
{\boldsymbol{m}}^c_{x, 0} \in \mb{R}^{M_1},
\boldsymbol{y}_{0},{\boldsymbol{m}}^c_{y, 0} \in \mb{R}^{M_2}$, step { sizes} $\mu_x$, $\mu_y$,
$\bx_1 = \boldsymbol{x}_{0}- \mu_x {\boldsymbol{m}}^c_{x, 0}, \bx_y = \boldsymbol{y}_{0}+ \mu_y{\boldsymbol{m}}^c_{y, 0}$ for \textbf{HCMM-1} or 
$\bx_1 = \boldsymbol{x}_{0}- \mu_x {\boldsymbol{m}}^c_{x, 0}/\|{\boldsymbol{m}}^c_{x, 0}\|, \bx_y = \boldsymbol{y}_{0}+ \mu_y{\boldsymbol{m}}^c_{y, 0}/\|{\boldsymbol{m}}^c_{y, 0}\|$ for \textbf{HCMM-2}, smoothing factors $\beta_x$, $\beta_y$, clipping threshold $N$ and clipping factor $N_1$.
\FOR{ $i = 1, \dots$}    
\STATE \underline{\text{Momentum update using random sample $\boldsymbol{\xi}_{i}$}:} 
\vspace{1em}
  
 $\displaystyle 
 \begin{aligned}
\bd{m}_{x,i} &=
 (1-\beta_x)\Big[ \bd{m}^c_{x,i-1}
 +\nabla^2_{xx}Q(\bd{x}_i, \bd{y}_{i};
 \boldsymbol{\xi}_{i})
 (\bd{x}_{i} - \bd{x}_{i-1}) \\
 & +
 \nabla^2_{xy}Q(\bd{x}_i, \bd{y}_{i};
 \boldsymbol{\xi}_{i})
 (\bd{y}_{i} - \bd{y}_{i-1})\Big]
 + \beta_x \nabla_x Q(\bd{x}_i, \bd{y}_{i};
 \boldsymbol{\xi}_{i})
 \end{aligned}
 $ 
 \vspace{1em}
  $\displaystyle \begin{aligned}
 \bd{m}_{y,i} &=
 (1-\beta_y)\Big[ \bd{m}^{c}_{y,i-1}
 +\nabla^2_{yy}Q(\bd{x}_i, \bd{y}_{i};
 \boldsymbol{\xi}_{i})(\bd{y}_{i} - \bd{y}_{i-1})
 \\
 & +
 \nabla^2_{yx}Q(\bd{x}_i, \bd{y}_{i};
 \boldsymbol{\xi}_{i}) (\bd{x}_{i} - \bd{x}_{i-1})
 \Big]
 + \beta_y \nabla_y Q(\bd{x}_i, \bd{y}_{i};
 \boldsymbol{\xi}_{i})
 \end{aligned}
 $
\STATE 
Choose one of the following options:\\
\underline{Gradient Clipping (\textbf{HCMM-1}):}
\\ if $\|\bd{m}_{x,i}\| \ge N \Longrightarrow$  $\bd{m}^c_{x,i} = N_1\frac{\bd{m}_{x,i}}{\|\bd{m}_{x,i}\|}$ otherwise
$\bd{m}^c_{x,i} = \bd{m}_{x,i}$\\
if $\|\bd{m}_{y,i}\| \ge N \Longrightarrow$ $\bd{m}^c_{y,i} = N_1\frac{\bd{m}_{y,i}}{\|\bd{m}_{y,i}\|}$
otherwise $\bd{m}^c_{y,i}=\bd{m}_{y,i}$
$$\displaystyle  
\bd{x}_{i+1}  = \boldsymbol{x}_{i} 
     - \mu_x \boldsymbol{m}^c_{x,i}, \quad  
\bd{y}_{i+1}  = \boldsymbol{y}_{i} 
     + \mu_y \boldsymbol{m}^c_{y,i}$$  \underline{\text{Normalization (\textbf{HCMM-2}):}} 
$$\displaystyle  
\bd{x}_{i+1}  = \boldsymbol{x}_{i} 
     - \mu_x \boldsymbol{m}_{x,i}/\|\boldsymbol{m}_{x,i}\|, \quad \bd{y}_{i+1}  = \boldsymbol{y}_{i} 
     + \mu_y \boldsymbol{m}_{y,i}/\|\boldsymbol{m}_{y,i}\|
     $$ 
$$
\bd{m}^c_{x,i}=\bd{m}_{x,i}, \quad \bd{m}^c_{y,i}=\bd{m}_{y,i}$$
\ENDFOR
\end{algorithmic}
\end{algorithm}
We present our algorithm, namely
\textbf{H}essian \textbf{C}orrected 
\textbf{M}omentum
\textbf{M}ethod
in \textbf{Algorithm \ref{example-algorithm}}.
The algorithm first initializes some appropriate quantities. At each iteration $i \ge 1$, we draw a random sample $\bd{\xi}_i$ to update the momentum vectors.
In step 2,
we need to compute accelerated momentum vectors using
the Hessian-vector product.
For small-scale problems, this can be done directly. For large-scale problems, we can use automatic differentiation techniques, such as \cite{paszke2017automatic}, without explicitly forming the full Hessian.
To stabilize the momentum vector, we next consider two clipping strategies: gradient clipping based on a predefined threshold (corresponding to \textbf{HCMM-1}) and consistent normalization (corresponding to \textbf{HCMM-2}).
Note that these two clipping operators introduce nonlinearity into the effective learning rates; they are essential for achieving faster convergence rates.
In the final step of both options, we employ a two-time-scale scheme, similar to SGDA \cite{lin2020gradient}, to update the model parameters.

\section{Convergence analysis}
In this section, we present convergence results for the proposed algorithms,
demonstrating their rates of convergence and the necessary conditions to \cblack{convergence toward an $\varepsilon$-stationary point of the function $P(x)$, to be defined in \eqref{maxfunction}}. 
We consider two important \cblack{cases for the} risk function. We begin with a basic setting of the nonconvex strongly-concave formulation and then address a more relaxed scenario of nonconvex-PL risk functions. The key assumptions supporting the proofs are listed below.


\vspace{-1em}
\setcounter{equation}{17}
\subsection{Assumptions}
The convergence analysis of minimax optimization problems has been pursued in the literature under conditions similar in spirit to those we list below. The main difference is that we will be relying on the Hessian condition and applying it to establish a better convergence of \textbf{HCMM-1} and \textbf{HCMM-2}.
To begin with, the convergence results for the proposed algorithms
are presented 
under the following two assumptions on the risk function.
\stepcounter{Assumption}
\begin{intassumption}[\textbf{Nonconvex strongly-concave}]
\label{NC_SC}
The risk function
$J(x, y)$
is  nonconvex in $x$ \cblack{and} $\nu$-strongly concave in $y$, 
where $\nu$ is a strictly positive constant. 
\end{intassumption}
\begin{intassumption}[\textbf{Nonconvex-PL}]
\label{NC_PL}
The risk function
$J(x, y)$
is  nonconvex in $x$ while 
$-J(x, y)$ is $\nu$-PL in $y$, 
i.e.,  $\forall x \in \mb{R}^{M_1}, y \in \mb{R}^{M_2}$, it holds that
\begin{equation} 
 \|\nabla_{y} J(x, y)\|^2
 \ge 2\nu(\underset{y} {\max} \ J(x, y) -  J(x, y))
\end{equation}
where $\nu$ is a strictly positive constant. 
\end{intassumption}
\textcolor{black}{
It is worth noting that the PL condition holds in certain over-parameterized neural network settings
\cite{liu2020linearity, liu2022loss}.}
We further \cblack{introduce} the following objective: 
\begin{equation}
\label{maxfunction}
  P(x) = \max_y J(x, y)  
\end{equation}
\textcolor{black}{
To avoid solving a trivial problem, we impose a condition on $P(x)$ similar to  \cite{lin2020gradient, yang2022faster}
}
\begin{Assumption}[\textbf{Lower-boundedness}] \label{boundPhi}
The  objective $P(x)$ is lower bounded, i.e.,
$P^\star =\operatorname{inf}_{x} P(x) > -\infty$.
\end{Assumption}
We \cblack{also} assume the risk gradient and Hessian are Lipschitz continuous.
\begin{Assumption}[\textbf{Lipschitz condition}]\label{LipschitzGradient}
 \cblack{The gradient vector of the risk function}
is $L_f$-Lipschitz, i.e.,
\begin{align} 
& \| \nabla_z J (x_1, y_1) - \nabla_z J (x_2, y_2)\| 
\le  
L_f \Big\|\begin{bmatrix}
    x_1 -x_2\\
    y_2-y_2
\end{bmatrix}\Big\|  
\end{align}
while the Hessian matrix
 is $L_h$-Lipschitz, i.e.,
\begin{align} 
\label{LipschitzHessian}
& \| \nabla^2_z J (x_1, y_1) - \nabla^2_z J (x_2, y_2)\| 
\le  
L_h
\Big\|\begin{bmatrix}
    x_1 -x_2\\
    y_2-y_2
\end{bmatrix}\Big\|  
\end{align}
\end{Assumption}


\textcolor{black}{
Condition \eqref{LipschitzHessian}
is {\em essential} for establishing a stronger convergence rate, as indicated by various works \cite{cutkosky2020momentum,tran2022better}.}

The following assumption may be stringent but is {\em only} needed for the analysis of \textbf{HCMM-1} to handle fourth-order error moments. Assumptions of this type are used in nonconvex optimization problems (see, e.g., \cite{xie2022adan, dou2021one,swenson2019annealing, vlaski2019diffusion}).
Moreover, it may hold locally along the optimization trajectory, as the gradient clipping helps stabilize the algorithm.

\begin{Assumption}[\textbf{Bounded gradient norm}]
\label{BoundedGradient}
    The norm of the \cblack{gradient vector} is bounded, i.e.,
    \begin{align}
        \|\nabla_z J(x, y)\| \le G
    \end{align}
\end{Assumption}
We will only use this result to \cblack{show that}  
the deviation between the clipped momentum and the true gradient is smaller than that of the nonclipped \cblack{momentum} in \textbf{HCMM-1} (see Lemma \ref{lemmaclipping}). 

\subsection{Convergence Metrics}
Similar to the minimax works \cite{xian2021faster, gao2022decentralized, chen2022simple, lin2020gradient},
we focus on 
finding an $\varepsilon$-stationary point $\bd{x}^\star$ such that 
\begin{align}
\mb{E}\|\nabla P(\bd{x}^\star)\| \le \varepsilon 
\label{criterion}.
\end{align}
This convergence criterion is 
suitable to the robust regression application considered in this work,
where the value function 
$P(x)$ represents some worst-case 
construction 
and 
$\bd{x}^\star$
is a robust model that minimizes the worst-case cost.
Note that we can also translate the above stationary point into a game-stationary point by fixing the found $\bd{x}^\star$ and applying gradient ascent to the inner maximization problem \cite{cai2025communication}.

\subsection{Main Results}
 The following theorem gives the convergence results of \textbf{HCMM-1} (see Appendix \ref{proofMaintheorem} for proof). 
\begin{Theorem}[\textbf{HCMM-1 convergence}]\label{Maintheorem}
Let Assumptions \ref{unbiased}, \ref{NC_SC} and \ref{boundPhi}--\ref{BoundedGradient} hold. The stability condition for the hyperparameters in \textbf{HCMM-1} is given by
\begin{subequations}
    \begin{align}
 \label{mainboy_hyperparamter_condition1}
\beta_x &= \beta_y \le \frac{1}{2} 
\\
 \mu_y &\le  \min \Big\{ \frac{\sigma_h \sqrt{2\beta_y}}{L_h N_1}, \sqrt{\frac{C\beta_y}{2}}, \sqrt{\frac{C\beta_y}{30\kappa^2}}, \frac{2}{\nu}, \pi_1\Big\}, 
  \label{mainboy_hyperparamter_condition2}\\
\mu_x &\le  \min \Big\{\mu_y, \frac{1}{480\kappa^4}\mu_y, \frac{1}{2L_1}\Big\}  
 \label{mainboy_hyperparamter_condition3}
\end{align}
\end{subequations}
where 
$\kappa = \frac{L_f}{\nu}, L_1 =L_f +\kappa L_f$, while
 $C, \pi_1$ are constants given by
\begin{align}
C = \min \Big\{\frac{5\pi_1L^2_f}{8\nu \sigma^2_h }, \frac{4\sigma^2_h}{N_1^2L^2_h}, \frac{1}{128 \sigma^2_h}\Big\}, \pi_1 = \frac{1}{2L_f+\nu} 
\end{align} 
We choose the smoothing factors as $\beta_x = \beta_y = \mathcal{O}(\frac{1}{T^{2/3}})$, and 
$\mu_x = c_1\sqrt{\beta_x},  \mu_y = c_2\sqrt{\beta_y}$
for some small constants $c_1 < c_2$. Then,
for sufficiently large $T$,
the convergence rate of \textbf{HCMM-1} is given by 
\begin{equation}
\begin{aligned}
 \frac{1}{T}   \sum_{i=0}^{T-1} \mathbb{E}\|\nabla P(\bd{x}_i)\| \le
\mathcal{O}\Big(\frac{1}{T^{1/3}}\Big)
\end{aligned}
\end{equation}
That is, 
{\normalfont \textbf{HCMM-1}} outputs an $\varepsilon$-stationary point $\bd{x}^\star$
after $T = \mathcal{O}(\varepsilon^{-3})$ iterations
and oracle complexity.
\end{Theorem}
The convergence of \textbf{HCMM-1} is guaranteed by choosing the smoothing factors and step sizes from the stability ranges defined by \eqref{mainboy_hyperparamter_condition1}--\eqref{mainboy_hyperparamter_condition3}.
However,
to achieve a theoretically optimal bound,
the smoothing factors
and step sizes are set based on $T$, 
specifically as $\beta_x= \beta_y =\mathcal{O}(\frac{1}{T^{2/3}})$, with 
the step sizes set as $\mu_x = c_1 \sqrt{\beta_x}, \mu_y = c_2 \sqrt{\beta_y}$.
Such a step size policy is similar to those used in \cite{lin2020gradient,yang2022faster,xian2021faster,huang2023enhanced}.
Note that the stability condition is easily satisfied when $T$ is large enough and  the constants $c_1$
  and 
$c_2$
  are properly tuned.
The condition  \eqref{mainboy_hyperparamter_condition3} theoretically requires $c_1 < c_2$, and we can tune them empirically.
Theorem \ref{Maintheorem} further implies that the convergence of \textbf{HCMM-1} relies on a two-time-scale step size policy. This reflects the unbalanced structure of the risk function and necessitates asymmetric step sizes to ensure convergence, as also discussed in \cite{lin2020gradient}.
\begin{Corollary}\label{Corollary}
Let Assumptions \ref{unbiased}, \ref{NC_PL} and \ref{boundPhi}--\ref{BoundedGradient} hold. Choose $\beta_x=\beta_y = \mathcal{O}(\frac{1}{T^{2/3}})$
and $\mu_x = c_1 \sqrt{\beta_x}, \mu_y = c_2 \sqrt \beta_y$, for some small constant $c_1 < c_2$. Then, the convergence rate of \textbf{HCMM-1} is given by $\mathcal{O}(1/{T^{1/3}})$.
See Appendix \ref{appendices:corollary} for proof.

\end{Corollary}
The above corollary shows that the strong convergence rate $(\mathcal{O}(1/T^{1/3}))$ also holds under weaker assumptions on the risk functions. Therefore, we focus on this weaker setting below.

The following theorem gives the convergence results of \textbf{HCMM-2} (see Appendix \ref{proof_Maintheorem2} for proof.)
\begin{Theorem}[\textbf{HCMM-2 convergence}] \label{Maintheorem2}
 Let  Assumptions \ref{unbiased}, \ref{NC_PL} and \ref{boundPhi}--\ref{LipschitzGradient} hold. The stability condition for the hyperparameters in \textbf{HCMM-2} is given by
  \begin{subequations}
        \begin{align}
    \beta_x &= \beta_y  \le 1\\
    \mu_x & \le \min\Big\{\mu_y, \frac{\mu_y}{6\kappa}\Big\} \label{mainbbody_theorem2_hyper}
    \end{align}
  \end{subequations}
    where $\kappa = L_f/\nu$ represents the condition number.
We choose the smoothing factors 
as $\beta_x =\beta_y = \mathcal{O}\Big(\frac{1}{T^{2/3}}\Big)$
and $\mu_y = \mathcal{O}\Big(\frac{1}{T^{2/3}}\Big), \mu_x = c_3 \mu_y$ for a small constant $c_3 < 1$.
Then, the convergence rate of \textbf{HCMM-2} is given by 
\begin{equation}
\begin{aligned}
 \frac{1}{T}   \sum_{i=0}^{T-1} \mathbb{E}\|\nabla P(\bd{x}_i)\| \le
\mc{O}\Big( \frac{1}{T^{1/3}}\Big)
    + \mc{O}\Big( \frac{1}{T}\Big) +\mc{O}\Big( \frac{1}{T^{2/3}}\Big)
\end{aligned}
\end{equation}
That is,
\textbf{HCMM-2} outputs an $\varepsilon$-stationary point $\bd{x}^\star$
after $T = \mathcal{O}(\varepsilon^{-3})$ iterations
and oracle complexity.
\end{Theorem}
Note that 
the stability condition for \textbf{HCMM-2}
is simpler compared to 
\textbf{HCMM-1}
with a constant normalization step. 
To ensure the convergence 
of \textbf{HCMM-2},
we 
only need to satisfy the two-time-scale step size policy indicated by \eqref{mainbbody_theorem2_hyper}.
However,
to achieve the optimal convergence rate, 
the step sizes $\mu_x,\mu_y$
need to be selected with a smaller order in $T$ than those used in \textbf{HCMM-1},
following a similar order of smoothing factors.
This is understandable because a small momentum norm in the steady-state regime can result in a large learning rate, which leads to reduced solution accuracy.
\begin{Remark}
We remark that the proof of Theorem \ref{Maintheorem2}
does not rely on the bounded gradient  
condition due to the consistent normalization step introduced in \textbf{HCMM-2}.
This normalization facilitates the analysis 
by ensuring that the actual update vector at each iteration has unit norm, thereby effectively controlling the weight increment throughout the analysis.
\end{Remark}

\color{black}
\begin{figure*}[htbp!]
\vspace{-2em}
\label{figureresult}
\label{figure}
  \begin{minipage}[b]{0.32\textwidth}
    \centering
\includegraphics[scale = 0.33]{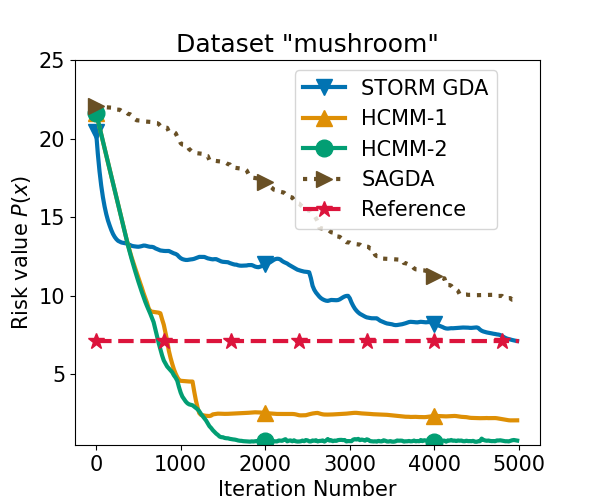} 
 \end{minipage}
  \hfill
  \begin{minipage}[b]{0.32\textwidth}
    \centering
\includegraphics[scale = 0.33]{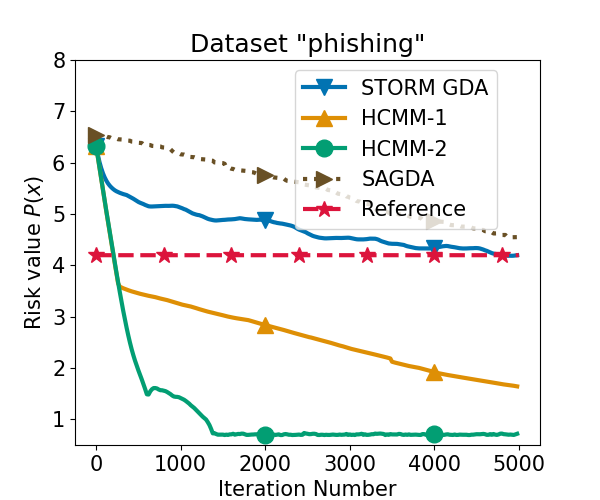}  
  \end{minipage}
  \hfill
  \begin{minipage}[b]{0.32\textwidth}
    \centering
\includegraphics[scale = 0.33]{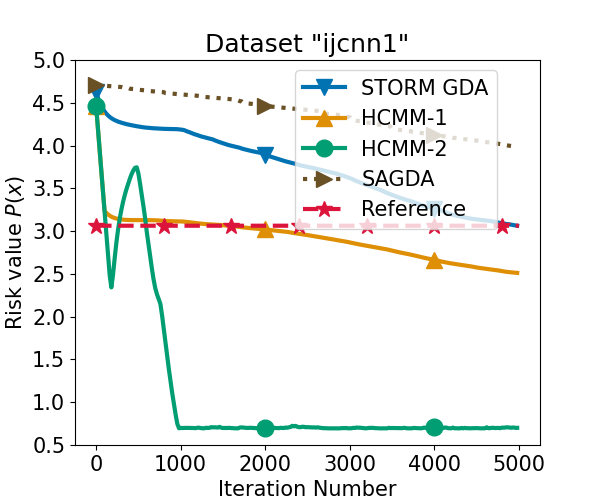}  
\end{minipage}
  \vfill
  \begin{center}
    \begin{minipage}[b]{0.33\textwidth}
    \centering
\includegraphics[scale = 0.33]{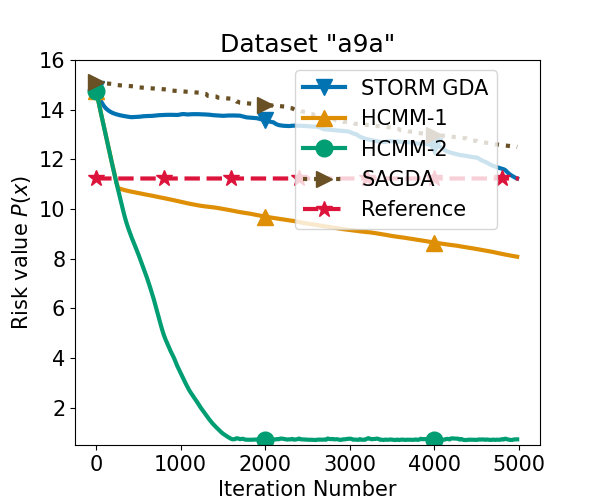} 
\end{minipage}
\hspace{-1em}
\begin{minipage}[b]{0.32\textwidth}
    \centering
\includegraphics[scale = 0.33]{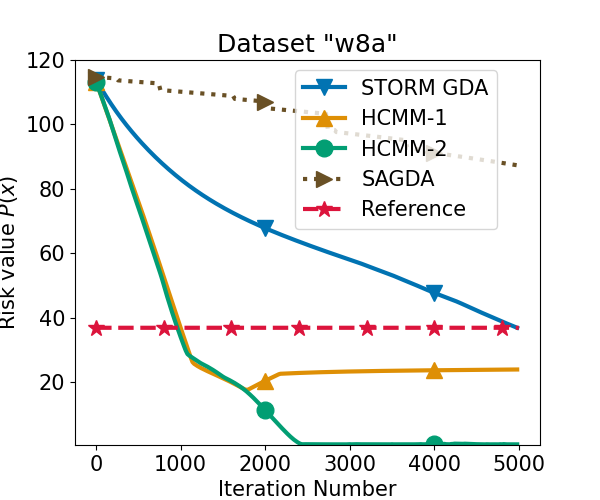} 
\end{minipage}
  \end{center}
\caption{
The figures, arranged from top to bottom and left to right, represent the results on the datasets "mushrooms", "phishing", "ijcnn1", "a9a", and "w8a", respectively. \cblack{These figures illustrate the worst-case risk value  $P(x)$
versus the number of iterations.}}
\end{figure*}
\begin{table*}[htbp!]
\centering
\renewcommand{\arraystretch}{1.2}
\begin{tabular}{|c||cc|cc|cc|}
\hline
\textbf{Dataset} & \multicolumn{2}{c|}{\textbf{STORM GDA}} & \multicolumn{2}{c|}{\textbf{HCMM-1}} & \multicolumn{2}{c|}{\textbf{HCMM-2}} \\
\cline{2-7}
 & \#Gradients & Time (s) & \#Gradients & Time (s) & \#Gradients & Time (s) \\
\hline
mushroom & 403.92k & 6.16 & 55.54k & 1.16 & 49.54k & 1.09 \\
phishing & 403.92k & 7.03 & 18.34k & 0.40 & 18.34k & 0.40 \\
ijcnn1   & 403.92k & 16.93 & 99.94k & 5.38 & 41.14k & 1.89 \\
a9a      & 403.92k & 26.82 & 18.34k & 1.16 & 18.34k & 1.21 \\
w8a      & 403.92k & 34.76 & 65.14k & 6.64 & 63.94k & 7.56 \\
\hline
\end{tabular}
\caption{
\normalfont Comparison of stochastic oracle calls and runtime of the momentum algorithms at the reference risk value within a finite-time horizon $T=5000$. Each cell reports stochastic oracle calls (in thousands) / time (in seconds) of a specific algorithm when it reaches the reference value within the simulated time horizon.}
\label{tab:oracle_runtime}
\end{table*}
\subsection{Proof Outline}
\color{black}
Our convergence analysis for \textbf{HCMM-1} relies on constructing a potential function, whereas the analysis for \textbf{HCMM-2} involves analyzing Euclidean norms of some terms.
We first examine the performance 
of \textbf{HCMM-1} under the nonconvex–strongly concave and nonconvex–PL 
settings. 
We then focus on a weaker setting for \textbf{HCMM-2}.


To prove the convergence of \textbf{HCMM-1} under  Assumption~\ref{NC_SC},  the following potential function is introduced:
\begin{align} \label{potential}
\bd{\Omega}_{i}
&= \mathbb{E}
\Big[P(\bd{x}_i) 
+ \eta \|\bd{y}^o(\bd{x}_i) - \bd{y}_i \|^2+ \gamma
\|\bd{m}^c_{x,i} 
\notag \\&  \quad  - \nabla_x J(\bd{x}_{i}, \bd{y}_{i})\|^2+\gamma\|\bd{m}^c_{y,i} - \nabla_y J(\bd{x}_{i}, \bd{y}_{i})\|^2
\Big]
\end{align}
where $\bd{\Omega}_i$ is a function of $\bx_{i}, \by_{i}, \bm^c_{x,i}, \bm^c_{y,i}$, and $\eta, \gamma$
are auxiliary parameters introduced solely for the purpose of the analysis.
The choice of these parameters and 
the role of $\bd{\Omega}_{i}$
will be clear in
the proof of Theorem \ref{Maintheorem}.

We use a different approach to establish the convergence of \textbf{HCMM-2}.
 This is 
because the momentum is normalized over iterations, making it challenging to establish the descent relation for the squared terms appearing in \eqref{potential}.
The key challenge in proving the convergence of 
\textbf{HCMM-2}
lies in finding \cblack{a} 
descent relation for the
deviation between $\bd{y}_i$
and $\bd{y}^o(\bd{x}_i)$, where
$\bd{y}^o(\bd{x}_i)$ is dependent on $\bd{x}_i$ and both are drifting over time.
If we consider starting from the squared norm
$\mb{E}\|\bd{y}^o(\bd{x}_i) - \bd {y}_i \|^2$,
we may only
establish a descent 
relation that is  meaningful in the asymptotic case
\cite{mai2021stability}.
We avoid this asymptotic approach because it obscures the convergence rate information, and practical training is usually performed over a finite number of iterations.
To address the challenge, we adopt a new non-asymptotic analysis to establish a descent relation for the first-order error moment $\mb{E}\|\bd{y}^o(\bd{x}_i) - \bd {y}_i\|$, which facilitates bounding target gradient norms.

\section{Computer Simulations}
In this section, we study two applications: robust logistic regression and robust adaptive cruise control (ACC) to illustrate the performance of the proposed algorithms.

\subsection{robust logistic regression}

We first consider the example of
distributionally robust logistic regression \cite{xian2021faster,luo2020stochastic}.
Suppose the dataset is given by $\{(r_i, l_i)\}_{i=1}^{n}$, 
where $r_i \in \mb{R}^{d}$
is the regression vector and 
$l_i \in \{+1, -1\}$
is the associated label.
We aim to find a robust model by solving the following minimax problem:
\begin{align}
\min _{x \in \mathbb{R}^d} \max _{y \in \Delta_n} J(x, y)=\sum_{i=1}^n y_i Q_i(x)-V(y)+g(x).
\end{align}
where 
\begin{subequations}
    \begin{align}
Q_i(x) &=\log \left(1+\exp \left(-l_i r_i^T x\right)\right) ,\\
g(x)&=\lambda_2 \sum_{i=1}^d \frac{\rho x_i^2}{1+\rho x_i^2},  \quad 
V(y)=\frac{1}{2} \lambda_1\|n y-\mathbf{1}_n\|^2,\\
\Delta_n &= \{y \in \mathbb{R}^n: 0 \leq y_i \leq 1, \sum_{i=1}^n y_i=1 \}.
\end{align}
\end{subequations}
Here, $Q_i(x)$ is the logistic loss function, $g(x)$ 
is a nonconvex regularizer suggested by \cite{antoniadis2011penalized},
$V(y)$ is the divergence measure,  $\Delta_n$
is the simplex set, and
$\mathbf{1}_n$
is the $n$-dimensional vector with all $1$.
Intuitively, the objective of this task is to find a robust model $x$
that minimizes 
the worst-case construction $P(x)$.
 This is achieved by maximizing $J(x,y)$ over the weight vector 
$y$ to determine a linear combination of loss values that yields the worst-case risk value $P(x)$.
All our experiments were run on an iMac with a 4.2 GHz Intel i7 CPU, 16 GB RAM.

We use five real-world datasets of
``mushrooms",  ``phishing", 
``ijcnn1", ``a9a" and ``w8a"
that can be downloaded from \cblack{the}
LIBSVM repository\footnote{\url{ https://www.csie.ntu.edu.tw/~cjlin/libsvmtools/datasets/}}.
Following the experimental setting in \cite{xian2021faster}, 
we use $\lambda_1 = \frac{1}{n^2}, \lambda_2 = 0.001$,
and $\rho = 10$.
We compare our algorithms
with STORM momentum-based GDA
\cite{xian2021faster,huang2023enhanced} (STORM GDA),
as well as the stochastic alternating GDA (SAGDA)
\cite{yang2022faster}.
For all algorithms,
we tune the step sizes
$\mu_x, \mu_y$
from $\{0.1, 0.01, 0.001\}$, smoothing factors 
$\beta_x, \beta_y$
from $\{0.01, 0.001\}$,
and then plot the best simulation results.
The other settings 
follow the same as in \cite{xian2021faster}.
To improve the empirical performance,
we directly update 
$\bd{m}_{x,i}, \bd{m}_{y,i}$
from the nonclipped momentum in step 2
and we tune $N, N_1$
from \{0.1, 0.01\}
as it is necessary for this algorithm to 
manage some higher-order moments.


The simulation results of the  algorithms are shown in Figure 1, \cblack{where 
we plot the worst-case risk value $P(x)$ over iterations}. 
From all figures,
we observe that the momentum methods outperform  SAGDA. Specifically, both \textbf{HCMM-1} and \textbf{HCMM-2} outperform the other algorithms in finite-time training. Notably, \textbf{HCMM-2} significantly outperforms the other algorithms in terms of convergence speed.
However, \textbf{HCMM-1} can be more robust than \textbf{HCMM-2}, as indicated   by the simulation results on the ``ijcnn1" dataset.
 This can be attributed to the presence of outliers
 and the smooth curvature of the loss landscape.
 Specifically, 
 when \textbf{HCMM-2} encounters outliers, its gradient direction may conflict with previous updates. Moreover, its amplified effective learning rate $\mu_x/\|\bm_{x,i}\|$  in smooth regions can cause an abrupt shift in the weights, leading to a sharp spike in the risk value.
 On the other hand,
although computing the Hessian-vector product incurs additional computation in practice, our algorithms require fewer iterations to achieve the risk value achieved by the best baselines, as shown in Table \ref{tab:oracle_runtime}.

\textbf{Synthesized data with outliers.}
\begin{figure*}[htbp!]
\centering
  \begin{minipage}[b]{0.3\textwidth}
    \centering
\includegraphics[scale = 0.35]{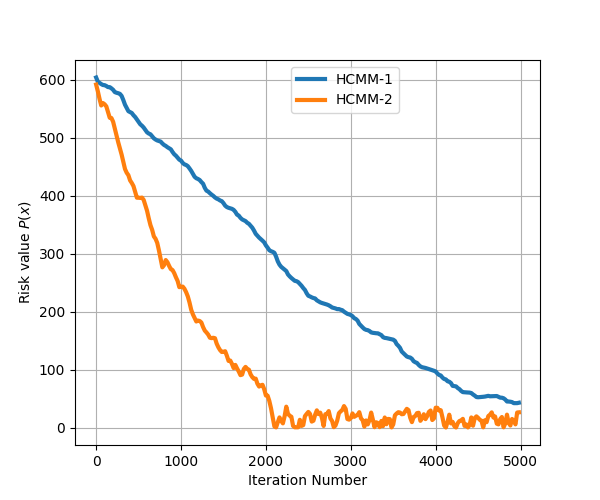} 
 \end{minipage}
  \begin{minipage}[b]{0.3\textwidth}
    \centering
\includegraphics[scale = 0.35]{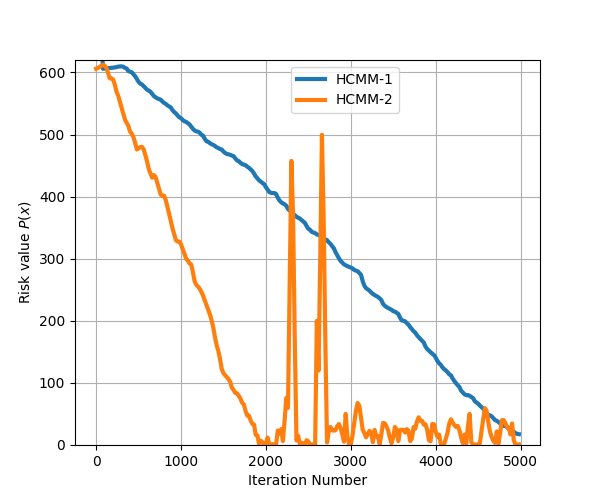}  
  \end{minipage}
\caption{
Comparison of algorithms \textbf{HCMM-1}
and \textbf{HCMM-2}
under synthesized data in the presence of outliers.
In the left figure, the algorithms are trained on linearly separable data.
In the right  figure,
 $10\%$ of the training data is comprised of the synthesized outliers.}
\label{figure:outlier}
\vspace{-1em}
\end{figure*}
Simulation results on the dataset  ``ijcnn1"
exhibit a sharp spike 
in the worst-case risk value produced by running \textbf{HCMM-2}.
We hypothesize that this phenomenon arises from a strong misalignment between the gradient direction at outlier samples and the directions of previous gradients.
To certify this, we test the proposed algorithms using synthetic data under two scenarios: one with linearly separable data and another where a subset of the training samples are outliers.
We randomly generate the training samples $[r^\top_1;
r^\top_2; \cdots; r^\top_n] \in \mathbb{R}^{n \times d}$, where each entry is independently drawn from the Gaussian distribution $\mathcal{N}(0, 20)$.
The entries of the ground truth weight vector  $x^\star$
is drawn from the Gaussian distribution
$\mathcal{N}(0, 1)$.
Given the feature vector $r_i$, the associated label $l_i \in \{-1, +1\}$
is generated according to the sign of the inner
product, i.e.,
\begin{align}
l_i  = {\rm sign}(r^\top_i x^\star), \forall i =1, \dots, n.
\end{align}
Furthermore, we generate a cluster of concentrated outlier features whose entries are independently drawn from the Gaussian distribution $\mathcal{N}\Big(\frac{1000x^\star_i}{\|x^\star\|}, 0.1\Big)$ $\forall i \in \{1, \dots, d\}$. They are located in a region that is distant from the high-density area of the clean data.
The simulation setups are set as $n= 30000, d =50$. 
For other hyperparameters, we tune them similarly as real-world dataset scenarios.
For the outlier scenario, we consider a setting where 
$10\%$ of the training data consists of outlier samples.
The simulation results are shown 
in Fig. \ref{figure:outlier}. It can be seen that \textbf{HCMM-2} is more susceptible to the influence of outliers compared to \textbf{HCMM-1}. This suggests that \textbf{HCMM-1} is a more robust approach in scenarios involving strong outliers.

\begin{figure*}[htbp!]
\centering
  \begin{minipage}[b]{0.4\textwidth}
    \centering
\includegraphics[scale = 0.45]{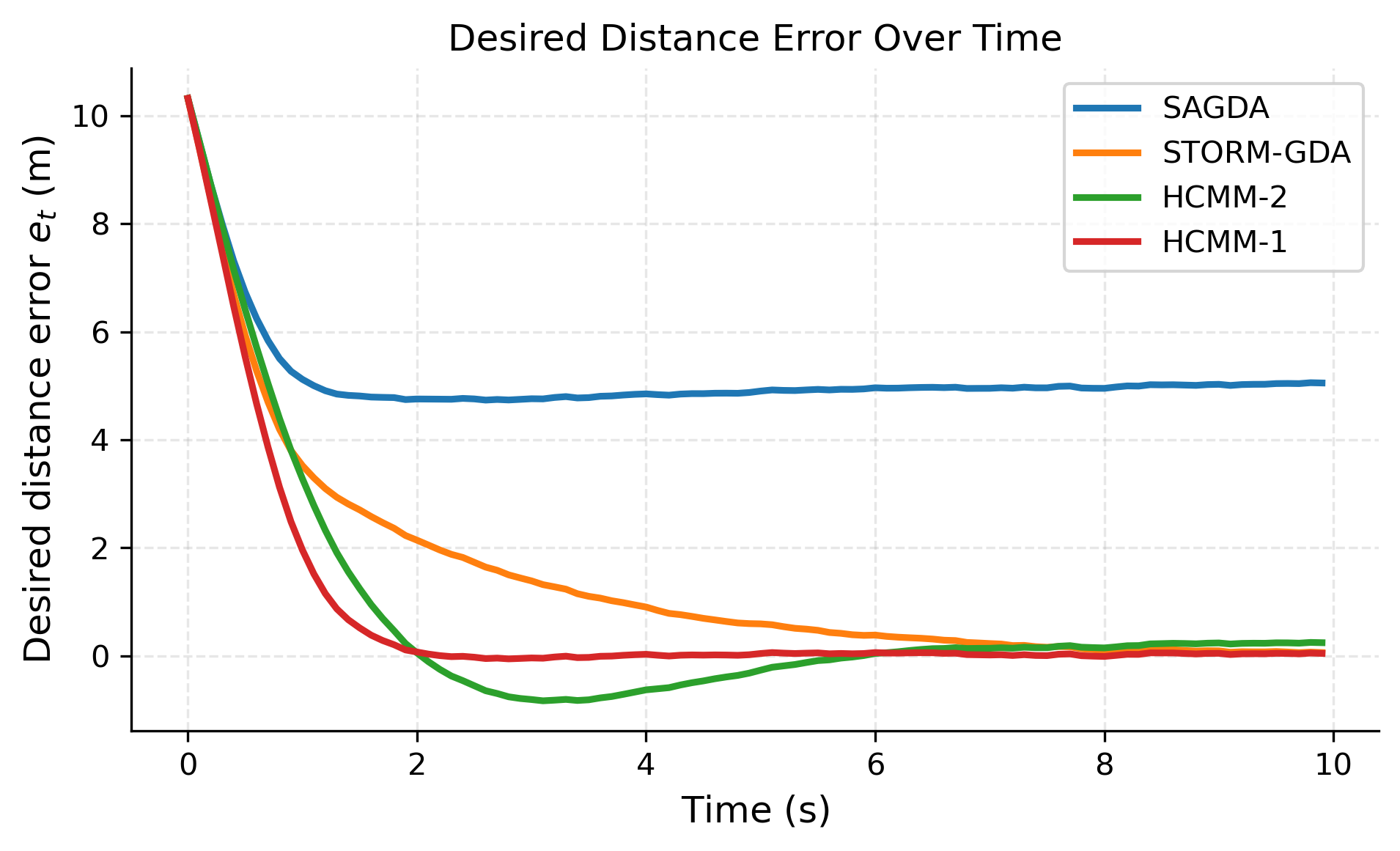} 
 \end{minipage}
  \begin{minipage}[b]{0.4\textwidth}
    \centering
\includegraphics[scale = 0.45]{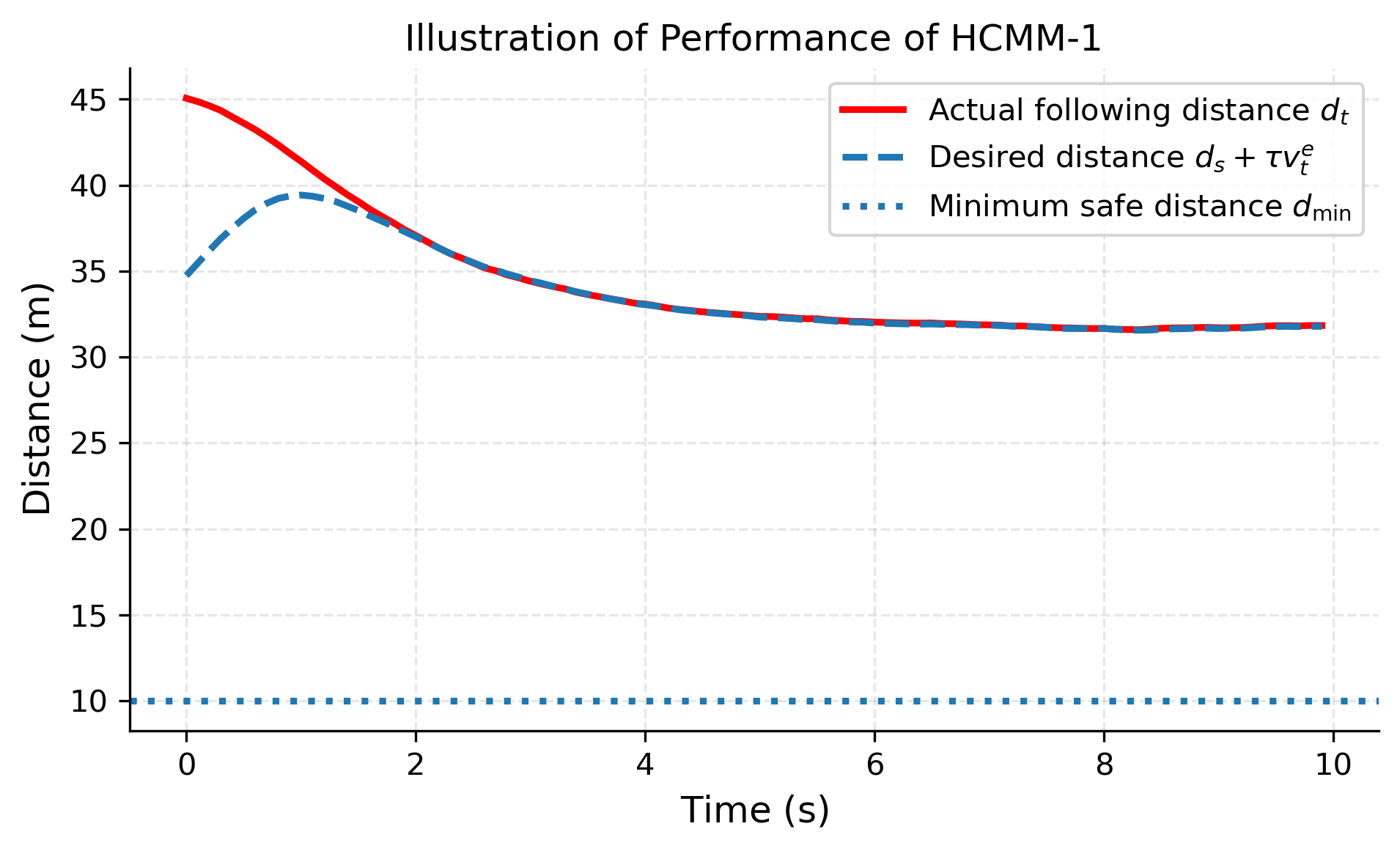}  
  \end{minipage}
\caption{
Robust ACC performance under adversarial lead-vehicle acceleration. Left: spacing error \(e_t\), where values closer to zero indicate better tracking of the desired following distance. Right: actual following distance, desired following distance, and minimum safe distance for \textbf{HCMM-1}.}
\label{figure:ACC}
\vspace{-1em}
\end{figure*}

\begin{table}[t]
\centering
\caption{Derivation of the stochastic ACC dynamics. Notation: $v^{\ell}_t$: velocity of lead vehicle. $v^{\ell}_e$: velocity of ego vehicle. 
$u_t$: acceleration command of the ego car. $w_t$: acceleration command of the lead car.
$h$: time step length. $d_{\min}$: distance at standstill. $d_t$: actual following distance of the ego vehicle. $e_t$: spacing error.}
\label{tab:acc_dynamics_derivation}
\small
\begin{tabular}{p{0.28\linewidth}p{0.64\linewidth}}
\toprule
Quantity & Derivation \\
\midrule
Distance & 
$
d_{t+1}=d_t+h(v_t^\ell-v_t^e).
$
\\
\midrule
Velocities &
 $
v_{t+1}^e=v_t^e+h u_t,\
v_{t+1}^\ell=v_t^\ell+h w_t, 
$
\\
\midrule
Spacing error &
Using \(e_t=d_t-d_{s}-\tau v_t^e\), we have \[ \begin{aligned} &e_{t+1} =d_{t+1}-d_{s}-\tau v_{t+1}^e\\ &=\bigl(d_t+h(v_t^\ell-v_t^e)\bigr) -d_{s} -\tau\bigl(v_t^e+h u_t\bigr)\\ &=\bigl(d_t-d_{s}-\tau v_t^e\bigr) +h(v_t^\ell-v_t^e)-\tau h u_t\\ &=e_t+h\Delta v_t-\tau h u_t. \end{aligned} \]
\\
\midrule
Relative velocity &
Using \(\Delta v_t=v_t^\ell-v_t^e\), we have
$
\Delta v_{t+1}
=v_{t+1}^\ell-v_{t+1}^e
=\Delta v_t+h(w_t-u_t). $\\
\midrule
Dynamics &
Adding random perturbations
\[
\begin{aligned}
e_{t+1}&=e_t+h\Delta v_t-\tau h u_t+\sigma_e\xi_t^e,\\
\Delta v_{t+1}&=\Delta v_t+h(w_t-u_t)+\sigma_v\xi_t^v,\\
v_{t+1}^e&=v_t^e+h u_t+\sigma_u\xi_t^u.
\end{aligned}
\]
\\
\bottomrule
\end{tabular}
\end{table}

\subsection{Stochastic robust adaptive  cruise control}

\begin{figure*}
\centering
\includegraphics[width=0.5\linewidth]{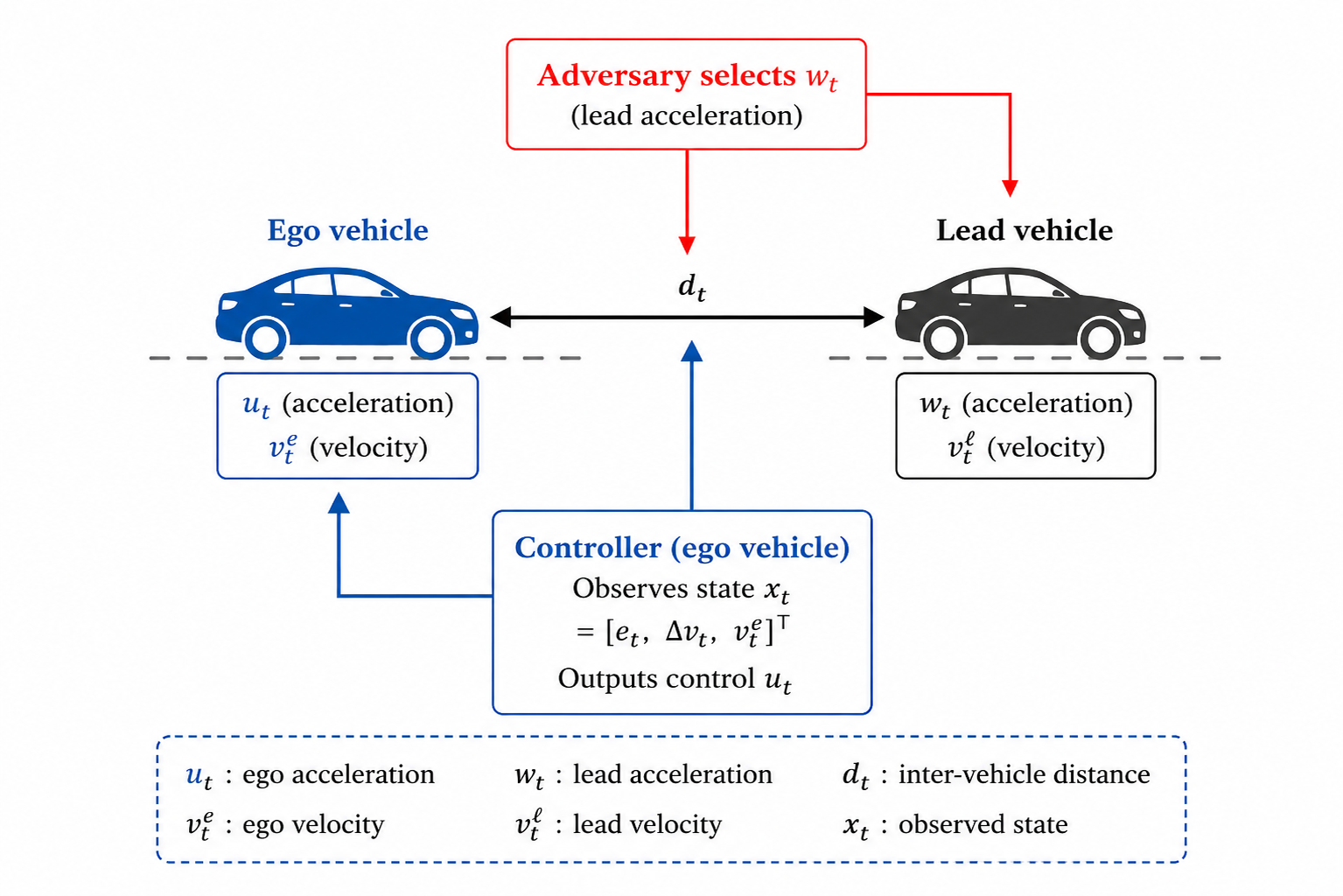}
    \caption{Illustration of the robust ACC experiment. In this application, the ego car aims to safely follow a lead vehicle that may exhibit aggressive behavior, such as sudden braking. The goal is to learn a controller gain $K$ that maps the observed state vector $x_t$ to an acceleration command, allowing the ego vehicle to speed up or slow down as needed to maintain a safe and comfortable following distance $d_t$.}
    \label{fig:ACC_illustration}
\end{figure*}

To further evaluate the practical performance of the proposed algorithm, we consider a stochastic robust ACC problem motivated by \cite{li2011model, gao2012explicit}.
In this problem, the ego vehicle aims to maintain a safe and smooth following distance from the lead vehicle, while an adversary selects the lead-vehicle acceleration to create difficult scenarios such as sudden braking.
We provide Figure \ref{fig:ACC_illustration} to illustrate this example.
Let \(d_t\) be the distance between the lead and ego vehicles, and let \(v_t^e\) and \(v_t^\ell\) be the velocities of the ego and lead vehicles, respectively. The desired following distance is
$
d_{\mathrm{des},t}=d_s+\tau v_t^e,
$
where \(d_s>0\) is the standstill distance and \(\tau>0\) is the time-headway parameter, controlling how much extra distance needs to be added when the ego vehicle moves faster. We define the spacing error and relative velocity as
$
e_t=d_t-d_s-\tau v_t^e,
\Delta v_t=v_t^\ell-v_t^e,
$
and obtain the state vector $x_t=[e_t,\Delta v_t,v_t^e]^\top$. The ego acceleration is generated by a saturated linear feedback controller, i.e.,
$
u_t=u_{\max}\tanh\left(-Kx_t/u_{\max}\right),
$
where \(K\) is the learnable controller gain and \(u_{\max}\) is the bound of the acceleration command. The adversary chooses a finite-horizon lead-vehicle acceleration sequence \(w=(w_0,\ldots,w_{T-1})\) to generate challenging lead-vehicle behaviors that increase the closed-loop cost of the ego vehicle.
The resulting stochastic min--max ACC problem can be formulated as
\begin{align}
\min_{K}\max_{w}
J(K,w)
&:= \mathbb{ E}_{\xi}\left[Q(K,w;\xi)\right],\\
Q(K,w;\xi)
&=
\sum_{t=0}^{T-1}
\Big(
q_e e_t^2
+q_v(\Delta v_t)^2
+q_u u_t^2
 \\
&+q_s [\operatorname{softplus}(d_{\min}-d_t)]^2
-\frac{\mu}{2}w_t^2\Big)
,\notag
\end{align}
subject to the stochastic ACC dynamics (see Table \ref{tab:acc_dynamics_derivation} for derivation)
\[
\begin{aligned}
e_{t+1}
&=e_t+h\Delta v_t-\tau h u_t+\sigma_e \xi_t^e,\\
\Delta v_{t+1}
&=\Delta v_t+h(w_t-u_t)+\sigma_v \xi_t^v,\\
v_{t+1}^e
&=v_t^e+h u_t+\sigma_u \xi_t^u,
\end{aligned}
\]
where \(h>0\) is the time step and \(\xi_t=[\xi_t^e,\xi_t^v,\xi_t^u]^\top\) is sampled independently from a standard Gaussian distribution. The terms \(q_e e_t^2\), \(q_v(\Delta v_t)^2\), and \(q_u u_t^2\) penalize spacing error, velocity mismatch, and aggressive behaviour, respectively. The term $q_s[\operatorname{softplus}(d_{\min}-d_t)]^2=q_s [\ln(1 + e^{(d_{\min}-d_t)})]^2$ is a smooth safety penalty that increases when the following distance is below the safety threshold \(d_{\min}\). The term \(-\frac{\mu}{2}w_t^2\) prevents the adversary from using unrealistically large accelerations.

 We simulate a \(10\)-second driving episode with \(h=0.1 s\) and \(T=100\). 
 The initial conditions are sampled as
$
d_0\sim 45+3\mathcal N(0,1),
v_0^e\sim 20+\mathcal N(0,1),
v_0^\ell\sim 18+\mathcal N(0,1),
$
with \(x_0=[d_0-d_s-\tau v_0^e,\ v_0^\ell-v_0^e,\ v_0^e]^\top\). 
We use \(d_s=5\), \(\tau=1.5\), \(d_{\min}=10\), \(u_{\max}=5\), \(w_t\in[-5,5]\), \((q_e,q_v,q_u,q_s,\mu)=(1,0.5,0.05,30,50)\), and \(\sigma_e=\sigma_v=\sigma_u=0.1\). All methods use minibatch size \(16\). We tune the stepsizes over \(\{0.1,0.01,0.001\}\), smoothing parameters over \(\{0.9,0.5,0.1\}\), and clipping thresholds over \(\{1.0,0.5,0.1\}\). For each iteration, we obtain the full trajectory \(\{x_t\}_{t=0}^{T-1}\) by unrolling the stochastic ACC dynamics. After training, we freeze the learned controller \(K\) and solve the inner maximization problem to obtain a worst-case adversarial sequence \(w^\star\) for evaluation.

Figure~\ref{figure:outlier} reports the performance of the learned ACC controllers. The left panel shows that SAGDA has difficulty tracking the desired following distance, while the proposed methods reduce the spacing error more rapidly and smoothly. The right panel shows that the best-performing algorithm \textbf{HCMM-1} closely tracks the desired following distance while staying above the safety threshold for the ego car. These results highlight the practical value of the proposed stochastic min--max algorithms in real-world control applications.

\section{Conclusion}
We proposed bias-corrected momentum methods for stochastic \cblack{nonconvex strongly-concave/PL} minimax problems, achieving  
$\mc{O}(\varepsilon^{-3})$
complexity under a Lipschitz Hessian and using $\mc{O}(1)$ batch size.
Experiments on real-world problems demonstrate strong performance. Future work includes adaptive variants and applications to larger-scale problems.

\section{Acknowledgement}
To enhance readability, some text \cblack{in} this article \cblack{has} undergone revision using ChatGPT 4.0.

\appendices

\section{Basic Lemmas for Nonconvex Strongly-Concave Risk Functions}

\label{appendice1}
We present the theoretical analysis hereafter. The structure of the proofs is organized as follows: the main technical lemmas for the nonconvex strongly-concave and nonconvex-PL setups are provided in Appendices \ref{appendice1} and \ref{appendice2}, respectively. The convergence proofs for \textbf{HCMM-1} are deferred to Appendices \ref{proofMaintheorem} 
and \ref{appendices:corollary}. 
The convergence proofs for \textbf{HCMM-2} are deferred to  \ref{proof_Maintheorem2}.

\setcounter{equation}{0}

\begin{Lemma}
{(\cite[Lemma 4.3]{lin2020gradient})}
\label{Danskin}
    Under Assumptions \ref{NC_SC} and \ref{LipschitzGradient},
    if $-J(x,y)$ is $L_f$-smooth \cblack{over the block variable $z= \mbox{\rm cat}\{x, y\}$} and $J(x,y)$ is
    $\nu$-strongly concave in $y$ for any fixed $x$,
    then:
    \begin{itemize}
        \item $P(x)$ is $L_1 \triangleq (L_f +\kappa L_f)$-smooth 
    and 
    \begin{align}
           \nabla P(x) = \nabla_x J(x , y^o(x)) 
    \end{align}
     where $\kappa = \frac{L_f}{\nu}$ is the condition number
     and $y^o(x)$ is the maximum 
point of $J(x, y)$ for a fixed $x$, i.e,
$y^o(x) =\operatorname{argmax}_y \  J(x, y)$.
\item Furthermore, $y^o(x)$ is $\kappa$-Lipschitz,
i.e., 
\begin{align}
    \|y^o(x_1) - y^o(x_2)\| \le \kappa \|x_1-x_2\|
\end{align}
    \end{itemize}
\end{Lemma}

\begin{Lemma} \label{Metrics}
Under Assumptions \ref{NC_SC} and \ref{LipschitzGradient}, the following result holds when running {\normalfont \textbf{HCMM-1}:}
    \begin{align}
    &\|\nabla P(\bd{x}_i)\| 
    \\
    &\le L_f \| \bd{y}^o_i(\bd{x}_i) - \bd{y}_i\|
    +\|\nabla_x J(\bd{x}_i, \bd{y}_i) - \bd{m}^c_{x,i}\| +\|\bd{m}^c_{x, i}\| \notag
    \end{align}
\end{Lemma}
\noindent
\noindent\textbf{Proof:}
Inserting $\bd{m}^c_{x,i}$
and $\nabla_x J(\bd{x}_i, \bd{y}_i)$  into 
$\| \nabla P(\bd{x}_i)\| $ sequentially and using the triangle inequality, we can deduce 
\begin{align}
        &\| \nabla P(\bd{x}_i)\| 
        \\
        &\le \|\nabla P(\bd{x}_i) - \bd{m}^c_{x,i}\| +\|\bd{m}^c_{x,i}\| \notag
        \\
        &\le 
         \|\nabla P(\bd{x}_i) - \nabla_x J(\bd{x}_i, \bd{y}_i)\| +
         \|  \nabla_x J(\bd{x}_i, \bd{y}_i) -\bd{m}^c_{x,i}\|
          \notag  \\
         &\quad +\|\bd{m}^c_{x,i}\| \notag \\
         &\overset{(a)}{=} 
          \|\nabla_x J(\bx_i, \by^o(\bx_i))- \nabla_x J(\bd{x}_i, \bd{y}_i)\| 
          \notag  \\
         &\quad +
         \|  \nabla_x J(\bd{x}_i, \bd{y}_i) -\bd{m}^c_{x,i}\|+\|\bd{m}^c_{x,i}\| \notag
         \\
        &\overset{(b)}{\le} 
        L_f \|\bd{y}^{o}_i(\bd{x}_i)-\bd{y}_i\|
        +\|\nabla_x J(\bd{x}_i, \bd{y}_i) -\bd{m}^c_{x,i}\|
    +\|\bd{m}^c_{x,i}\| \notag
\end{align}
where $(a)$ follows from Lemma \ref{Danskin}, $(b)$ is derived by using the $L_f$-smooth assumption.
\qed

\begin{Lemma}
\label{lemmaclipping}
    Under Assumption \ref{BoundedGradient}, choosing 
    $N \ge N_1 \ge G$, the following result holds \cblack{for} {\normalfont \textbf{HCMM-1}:}
    \begin{align} 
    \label{deviationineq}
    \|\bd{m}^c_{u,i} - \nabla_u J(\bd{x}_i, \bd{y}_i)\|^2 \le 
\|\bd{m}_{u,i} - \nabla_u J(\bd{x}_i, \bd{y}_i)\|^2
\end{align}
where $u = x \text{ or } y$.
\end{Lemma}
\noindent \textbf{Proof:}
The proof for $u=x$ 
or $u=y$
is similar.
If 
no gradient clipping happens, i.e.,
$\|\bd{m}_{x,i}\| \le N$,
we always have  
\begin{align}
\|\bd{m}^c_{x,i} - \nabla_x J(\bd{x}_i, \bd{y}_i)\|^2 = 
\|\bd{m}_{x,i} - \nabla_x J(\bd{x}_i, \bd{y}_i)\|^2
\end{align}
Let us then consider $\|\bd{m}_{x,i}\| \ge N$.
Expanding the squared norm
in \eqref{deviationineq}, we notice that
    proving \eqref{deviationineq}
    is equivalent to 
    proving  
    \begin{align}
    2 \langle  \bd{m}_{x,i} - \bd{m}^c_{x,i}, \nabla_x J(\bd{x}_i, \bd{y}_i)\rangle 
    \le   \|\bd{m}_{x,i}\|^2
-\|\bd{m}^c_{x,i}\|^2 
\label{deviationineqstep1}
\end{align}
For the term $\|\bd{m}_{x,i}\|^2
-\|\bd{m}^c_{x,i}\|^2\ge 0  
 $, where $\|\bm_{x,i}\|^2 \ge N \ge N_1 =\|\bd{m}^c_{x,i}\|$,
we can deduce that 
\begin{align}
&\|\bd{m}_{x,i}\|^2
-\|\bd{m}^c_{x,i}\|^2   \notag \\
&=(\|\bd{m}_{x,i}\|
+\|\bd{m}^c_{x,i}\|)\underbrace{(\|\bd{m}_{x,i}\|
-\|\bd{m}^c_{x,i}\|)}_{\ge 0} \notag 
\\
&\overset{(a)}{\ge} 2N_1\underbrace{(\|\bd{m}_{x,i}\|
-\|\bd{m}^c_{x,i}\|)}_{\ge 0} \notag \\
&\overset{(b)}{=} 
2N_1 \Big(\|\bm_{x,i}\| - \frac{N_1}{\|\bm_{x,i}\| }\|\bm_{x,i}\|\Big) \notag \\
&= 
2N_1 \Big(1 -  \frac{N_1}{\|\bm_{x,i}\| }\Big) \|\bm_{x,i}\| \\
&\overset{(c)}{=}2N_1 \Big\|\Big(1- \frac{N_1}{\|\bm_{x,i}\|} \Big) \bm_{x,i}\Big\| \notag 
\\
&\overset{(d)}{\ge}
2N_1 \|\bd{m}_{x,i}-\bd{m}^c_{x,i}\| \notag \\
&\overset{(e)}{\ge}
2\frac{N_1}{G}
\|\nabla_x J(\bd{x}_i, \bd{y}_i)\|\|\bd{m}_{x,i}-\bd{m}^c_{x,i}\| \notag \\
&\overset{(f)}{\ge}
2
\|\nabla_x J(\bd{x}_i, \bd{y}_i)\|\|\bd{m}_{x,i}-\bd{m}^c_{x,i}\| \notag \\
&\overset{(g)}{\ge}
2 \langle
\nabla_x J(\bd{x}_i, \bd{y}_i),
\bd{m}_{x,i}-\bd{m}^c_{x,i}
\rangle
\end{align}
where $(a)$
follows from 
$\|\bm_{x,i}\|^2 \ge N \ge N_1 =\|\bd{m}^c_{x,i}\|$;
(b) follows from the gradient clipping step 
$\bm^c_{x,i} = \frac{N_1}{\|\bm_{x,i}\|} \bm_{x,i}$
and the fact that $\frac{N_1}{\|\bm_{x,i}\|} $ is positive; $(c)$ follows from 
$(1- \frac{N_1}{\|\bm_{x,i}\|}) \ge 0$;
$(d)$
follows from the gradient clipping step 
$\bm^c_{x,i} = \frac{N_1}{\|\bm_{x,i}\|} \bm_{x,i}$; $(e)$ follows from the fact that 
$\|\nabla_x J(\bx_{i}, \by_{i})\| \le G$;
$(f)$ follows from $N_1 \ge G$;
$(g)$ follows from the
Cauchy-Schwarz inequality.
Thus, \eqref{deviationineqstep1}
is satisfied by choosing appropriate $N_1$.
\qed

Lemma \ref{lemmaclipping}
implies that the deviation between the clipped stochastic momentum and the true gradient is smaller than that of the nonclipped one
by choosing appropriate $N, N_1$.

\begin{Lemma}\label{FunctionIncremental}
Under Assumptions \ref{NC_SC}
and 
\ref{LipschitzGradient},
choosing  $\mu_x \le \frac{1}{2L_1}$,
the following result holds for {\normalfont \textbf{HCMM-1}}:
\begin{align} \label{LemmaP}
&\mb{E}[P(\bd{x}_{i+1})
- P(\bd{x}_{i})] \notag \\
&\le \mu_x L^2_f\mb{E}\|\bd{y}^o(\bd{x}_i) - \bd{y}_i\|^2 +  \mu_x \mb{E}\| \nabla_x J(\bd{x}_i, \bd{y}_i) -\bd{m}^c_{x,i}\|^2 \notag  \hspace{10em}\\
& \quad 
-\frac{\mu_x}{4}\mb{E}\|\bd{m}^c_{x,i}\|^2 
\end{align}
\end{Lemma}
\noindent\textbf{Proof:}
From \textbf{Lemma \ref{Danskin}},
$P(x)$
is $L_1$-smooth, \cblack{so} we have 
\begin{align}\label{startsmooth1}
 & \ {{P}}(\boldsymbol{x}_{i+1}) \notag\\
  &  \le  
  {{P}}(\boldsymbol{x}_{i})
  + \langle 
  \nabla {{P}}(\boldsymbol{x}_{i}),
  \boldsymbol{x}_{i+1}
  -
  \boldsymbol{x}_{i}
  \rangle
  + \frac{L_1}{2}\|\boldsymbol{x}_{i+1}-  \boldsymbol{x}_{i}\|^2  \notag\\
&\le   
{{P}}(\boldsymbol{x}_{i}) -
\mu_x \langle 
  \nabla {{P}}(\boldsymbol{x}_{i}),
  \boldsymbol{m}^c_{x,i}
  \rangle
  + 
  \frac{L_1\mu^2_x}{2}\|\boldsymbol{m}^c_{x,i}\|^2 \notag
  \\
  &
\le 
P(\boldsymbol{x}_{i})
\underbrace{-\frac{\mu_x}{2}\|\nabla P(\boldsymbol{x}_{i})\|^2}_{\le 0}
- \frac{\mu_x}{2}
\|\boldsymbol{m}^c_{x,i}\|^2 
+\frac{\mu_x}{2}\|\nabla {{P}}(\boldsymbol{x}_{i})
\notag  \\
&\quad
-\boldsymbol{m}^c_{x,i}\|^2+ 
  \frac{L_1\mu^2_x}{2}\|\boldsymbol{m}^c_{x,i}\|^2
\notag
\\
&\overset{(a)}{\le} 
P(\boldsymbol{x}_{i})
-\frac{\mu_x}{2}\|\boldsymbol{m}^c_{x,i}\|^2
+\mu_x \|\nabla P(\boldsymbol{x}_{i})
- \nabla_x J(\boldsymbol{x}_{i}, \bd{y}_i)\|^2 \notag \\
&\quad + \mu_x \|\nabla_x J(\boldsymbol{x}_{i}, \bd{y}_i) - \boldsymbol{m}^c_{x,i}\|^2
+   \frac{L_1\mu^2_x}{2}\|\boldsymbol{m}^c_{x,i}\|^2 \notag \\
&\overset{(b)}{\le} 
P(\boldsymbol{x}_{i})
-\frac{\mu_x}{2}\|\boldsymbol{m}^c_{x,i}\|^2
+ \mu_x L^2_f\|\bd{y}^o(\bd{x}_i) - \bd{y}_i\|^2 \notag 
\\
&\quad  
+ \mu_x \|\nabla_x J(\boldsymbol{x}_{i}, \bd{y}_i) - \boldsymbol{m}^c_{x,i}\|^2
+   \frac{L_1\mu^2_x}{2}\|\boldsymbol{m}^c_{x,i}\|^2
\end{align}
where $(a)$ follows by adding and subtracting $\nabla_x J(\bx_{i}, \by_{i})$
and using Jensen's inequality; $(b)$ is due to $L_f$-Lipschitz assumption.
Moving $P(\boldsymbol{x}_{i})$ to the left-hand side (LHS) of \eqref{startsmooth1}, taking expectations and choosing
$\mu_x \le \frac{1}{2L_1} \Longrightarrow \frac{L_1\mu^2_x}{2} \le \frac{\mu_x}{4}$,
we arrive at \eqref{LemmaP}.
\qed

 \begin{Lemma}\label{meanvalue}
Under Assumption \ref{LipschitzGradient},
for either algorithm, the following result holds: 
\begin{align}
&\Bigg\|
\nabla_z J (x_2, y_2)
- \nabla_z J (x_1, y_1)+ \nabla^2_{z}J(x_1, y_1)
\begin{bmatrix}
    x_1- x_2\\
    y_1-y_2
\end{bmatrix} 
\Bigg\| \notag \\ 
&\le 
\frac{L_h}{2}\|z_1 -z_2\|^2
\end{align}
\end{Lemma}
\noindent\textbf{Proof:}
Using the mean value theorem
for the concatenated variable $z = \mbox{\rm cat}\{x, y\}$,
we obtain 
\begin{align}\label{meanvaluestep1}
& \nabla_z J(x_2, y_2)
    - \nabla_z J(x_1, y_1) \notag \\
& =  \int_{0}^{1}
\nabla^2_{z} J\Big(x_1+ t(x_2-x_1), y_1+ t(y_2-y_1)\Big)
 \begin{bmatrix}
     x_2 - x_1 \\
     y_2 - y_1
 \end{bmatrix} dt
\end{align}
Adding $\nabla^2_{z}J(x_1, y_1)
\begin{bmatrix}
    x_1- x_2\\
    y_1-y_2
\end{bmatrix}$
into both sides of \eqref{meanvaluestep1} and taking the $\ell_2$-norm, we have
\begin{align}
&
\Bigg\| \nabla_z J (x_2, y_2)
- \nabla_z J (x_1, y_1)+ \nabla^2_{z}J(x_1, y_1)
\begin{bmatrix}
    x_1- x_2\\
    y_1-y_2
\end{bmatrix} \Bigg\|  \notag
\\
& = \Bigg\|
\int_{0}^{1} \Big[\nabla^2_{z}J(x_1, y_1) 
 \notag
\\&-\nabla^2_{z} J\Big(x_1+ t(x_2-x_1), y_1+ t(y_2-y_1)\Big)
 \Big]
\begin{bmatrix}
    x_1- x_2\\
    y_1-y_2
\end{bmatrix} dt 
\Bigg\|\notag 
\\
& \overset{(a)}{\le}
\int_{0}^{1}\Bigg\| \Big[ \nabla^2_{z}J(x_1, y_1) \notag
\\& -
\nabla^2_{z} J\Big(x_1+ t(x_1- x_2), y_1+ t(y_1-y_2)\Big)
 \Big]
\begin{bmatrix}
    x_1- x_2\\
    y_1-y_2
\end{bmatrix}
\Bigg\|\notag dt \notag 
\\
\label{meanvaluestep2}
& \overset{(b)}{\le}
\int_{0}^{1}\Big\| \nabla^2_{z}J(x_1, y_1)
 \notag
\\&-\nabla^2_{z} J\Big(x_1+ t(x_1-x_2), y_1+ t(y_1-y_2)\Big)
\Big\|\Bigg\|
\begin{bmatrix}
    x_1- x_2\\
    y_1-y_2
\end{bmatrix}
\Bigg\|\notag dt \notag
\\
&\overset{(c)}{\le}
\int_{0}^{1} t L_h \Bigg\|
\begin{bmatrix}
    x_1- x_2\\
    y_1-y_2
\end{bmatrix} \Bigg\|^2dt \notag \hspace{20em}
\\
& \le  \frac{L_h}{2}\|
z_1 -z_2 \|^2 
\end{align}
where $(a)$ follows from the triangle inequality of \cblack{the} $\ell_2$-norm,
$(b)$ follows from \cblack{the} sub-multiplicative property of norms, \cblack{and}
$(c)$ follows from Assumption \ref{LipschitzGradient}.
\qed

\begin{Lemma} \label{GradientIncremental}
Under Assumptions \ref{unbiased} and \ref{LipschitzGradient}, choosing $\beta_u \le\frac{1}{2}$ ($u = x \text{ or } y$),
the following result holds \cblack{for} {\normalfont \textbf{HCMM-1}:}
\begin{align} \label{gradientincremenal_eq}
\mbE&\Big[\|\bd{m}^c_{u,i+1} - \nabla_u J(\bd{x}_{i+1}, \bd{y}_{i+1})\|^2-\|\bd{m}^c_{u,i}- \nabla_u J(\bd{x}_i, \bd{y}_{i})\|^2\Big]\notag 
  \\
&\le 
-\beta_u\mbE\|\bd{m}^c_{u,i} - \nabla_u J(\bd{x}_i, \bd{y}_{i})\|^2
+ \frac{L^2_h}{2\beta_u}\mbE
\Bigg\|\begin{bmatrix}
    \bd{x}_{i+1} - \bd{x}_i \\
    \bd{y}_{i+1} - \bd{y}_i
\end{bmatrix}\Bigg\|^4
 \notag \\
& 
+ 2 (1-\beta)^2 \sigma^2_h\mbE
\Bigg\|\begin{bmatrix}
    \bd{x}_{i+1} - \bd{x}_i \\
    \bd{y}_{i+1} - \bd{y}_i
\end{bmatrix}\Bigg\|^2 + 2 \beta^2_u \sigma^2
\end{align}
\end{Lemma}
\noindent\textbf{Proof:}
The proof is similar for $x$ and $y$. 
We focus on the $x$-variable.
  It is noted that
if $\|\bd{m}_{x,i}\| \le  N$, we have 
${\bd{m}}^c_{x,i} =\bd{m}_{x,i}$, which leads to
\begin{align}
    \|\bd{m}^c_{x,i} - \nabla_x J(\bd{x}_i, \bd{y}_i)\|^2 =
\|\bd{m}_{x,i} - \nabla_x J(\bd{x}_i, \bd{y}_i)\|^2
\end{align}
On the other hand, if 
$\|\bd{m}_{x,i}\| \ge  N$,
we can show that
\begin{align}
\|\bd{m}^c_{x,i} - \nabla_x J(\bd{x}_i, \bd{y}_i)\|^2 \le 
\|\bd{m}_{x,i} - \nabla_x J(\bd{x}_i, \bd{y}_i)\|^2 
\end{align}
using the argument from Lemma \ref{lemmaclipping}.
Inserting the recursion of $\bd{m}_{x,i}$ into 
$\|\bd{m}_{x,i} - \nabla_x J(\bd{x}_i, \bd{y}_i)\|^2$,
we get 
\begin{align}
\label{GradientIncrementalstep1}
&\|\bd{m}^c_{x,i} - \nabla_x J(\bd{x}_i, \bd{y}_i)\|^2 \notag \\
&\le \|\bd{m}_{x,i} - \nabla_x J(\bd{x}_i, \bd{y}_i)\|^2
\notag \\
&= 
\Big\|
(1-\beta_x)\Big[ \bd{m}^c_{x,i-1}
 +\nabla^2_{xx}Q(\bd{x}_i, \bd{y}_{i};
 \boldsymbol{\xi}_{i})
 (\bd{x}_{i} - \bd{x}_{i-1})  \notag\\
 & \quad +
 \nabla^2_{xy}Q(\bd{x}_i, \bd{y}_{i};
 \boldsymbol{\xi}_{i})
 (\bd{y}_{i} - \bd{y}_{i-1})\Big]
 + \beta_x \nabla_x Q(\bd{x}_i, \bd{y}_{i};
 \boldsymbol{\xi}_{i}) \notag\\&
 \quad - \nabla_x J(\bd{x}_i, \bd{y}_i)
\Big\|^2 \notag \\
&= \Big\|
(1-\beta_x)[\bd{m}^c_{x,i-1} - \nabla_x J(\bd{x}_{i-1}, \bd{y}_{i-1})]
+(1-\beta_x) \notag\\
& \quad \times \Big[ 
\nabla_x J(\bd{x}_{i-1}, \bd{y}_{i-1})
-\nabla_x J(\bd{x}_{i}, \bd{y}_{i})
+ \nabla^2_{xx}Q(\bd{x}_i, \bd{y}_{i};
 \boldsymbol{\xi}_{i}) \notag
 \\& \quad \times (\bd{x}_{i} - \bd{x}_{i-1})+
  \nabla^2_{xy}Q(\bd{x}_i, \bd{y}_{i};
 \boldsymbol{\xi}_{i})
 (\bd{y}_{i} - \bd{y}_{i-1})
\Big] \notag \\
& \quad
+\beta_x\Big[\nabla_x Q(\bd{x}_i, \bd{y}_{i};
 \boldsymbol{\xi}_{i})- 
 \nabla_x J(\bd{x}_i, \bd{y}_{i})\Big]
\Big\|^2  \notag
\\
&\overset{(a)}{=} 
(1-\beta_x)^2\| 
\bd{m}^c_{x,i-1} - \nabla_x J(\bd{x}_{i-1}, \bd{y}_{i-1})
\|^2
+   2\Big\langle  
(1-\beta_x)
\notag 
\\
& \quad \times [\bd{m}^c_{x,i-1} - \nabla_x J(\bd{x}_{i-1}, \bd{y}_{i-1})], (1-\beta_x)
\Big[ 
\nabla_x J(\bd{x}_{i-1}, \bd{y}_{i-1}) \notag\\
& \quad  
-\nabla_x J(\bd{x}_{i}, \bd{y}_{i})
+ \nabla^2_{xx}Q(\bd{x}_i, \bd{y}_{i};
 \boldsymbol{\xi}_{i})(\bd{x}_{i} - \bd{x}_{i-1}) \notag
 \\& \quad +
  \nabla^2_{xy}Q(\bd{x}_i, \bd{y}_{i};
 \boldsymbol{\xi}_{i})
 (\bd{y}_{i} - \bd{y}_{i-1})
\Big]+\beta_x\Big[\nabla_x Q(\bd{x}_i, \bd{y}_{i};
 \boldsymbol{\xi}_{i}) \notag \\
 &\quad - 
 \nabla_x J(\bd{x}_i, \bd{y}_{i})\Big]
\Big\rangle 
+ \Big\| 
(1-\beta_x)\Big[ 
\nabla_x J(\bd{x}_{i-1}, \bd{y}_{i-1}) \notag\\
& \quad  
-\nabla_x J(\bd{x}_{i}, \bd{y}_{i})
+ \nabla^2_{xx}Q(\bd{x}_i, \bd{y}_{i};
 \boldsymbol{\xi}_{i})(\bd{x}_{i} - \bd{x}_{i-1}) \notag
 \\& \quad +
  \nabla^2_{xy}Q(\bd{x}_i, \bd{y}_{i};
 \boldsymbol{\xi}_{i})
 (\bd{y}_{i} - \bd{y}_{i-1})
\Big]+\beta_x\Big[\nabla_x Q(\bd{x}_i, \bd{y}_{i};
\boldsymbol{\xi}_{i}) \notag \\
 &\quad - 
 \nabla_x J(\bd{x}_i, \bd{y}_{i})\Big]
\Big\|^2 
\end{align}
where $(a)$ follows from expanding the squared term
$\|a+b\|^2 = \|a\|^2 + 2\langle a, b \rangle +\|b\|^2$.
Taking expectations of 
\eqref{GradientIncrementalstep1}  over the sample 
$\bd{\xi}_i$ conditioned 
on $\bd{\mc{F}}_i$
and using Assumption \ref{unbiased},
we obtain 
\begin{align} \label{GradientIncrementalstep2}
&\mbE[\|\bd{m}^c_{x,i} - \nabla_x J(\bd{x}_i, \bd{y}_i)\|^2 \mid \bd{\mathcal{F}}_{i}] \notag \\
&\le 
(1-\beta_x)^2\| 
\bd{m}^c_{x,i-1} - \nabla_x J(\bd{x}_{i-1}, \bd{y}_{i-1})
\|^2 + \notag \\
& \quad 
2(1-\beta_x)^2
\Big\langle  
\bd{m}^c_{x,i-1} - \nabla_x J(\bd{x}_{i-1}, \bd{y}_{i-1}),
\nabla_x J(\bd{x}_{i-1}, \bd{y}_{i-1})\notag 
\\
& \quad 
-\nabla_x J(\bd{x}_{i}, \bd{y}_{i}) + \nabla^2_{xx}J(\bd{x}_i, \bd{y}_{i})(\bd{x}_{i} - \bd{x}_{i-1}) +
  \notag\\
& \quad  
\nabla^2_{xy}J(\bd{x}_i, \bd{y}_{i})
 (\bd{y}_{i} - \bd{y}_{i-1}) 
\Big\rangle 
+ \mbE\Big[\Big\| 
(1-\beta_x)\Big[ 
\nabla_x J(\bd{x}_{i-1} \notag\\
& \quad  , \bd{y}_{i-1})
-\nabla_x J(\bd{x}_{i}, \bd{y}_{i})
+ \nabla^2_{xx}Q(\bd{x}_i, \bd{y}_{i};
 \boldsymbol{\xi}_{i})(\bd{x}_{i} - \bd{x}_{i-1}) \notag
 \\& \quad +
  \nabla^2_{xy}Q(\bd{x}_i, \bd{y}_{i};
 \boldsymbol{\xi}_{i})
 (\bd{y}_{i} - \bd{y}_{i-1})
\Big]+\beta_x\Big[\nabla_x Q(\bd{x}_i, \bd{y}_{i};
 \boldsymbol{\xi}_{i}) \notag \\
 &\quad - 
 \nabla_x J(\bd{x}_i, \bd{y}_{i})\Big]
\Big\|^2 \mid \bd{\mc{F}}_{i}\Big]
\end{align}
In the following, we will establish the bounds for the 
cross-term and the last squared-term, respectively.
For the cross-term, we \cblack{have} 
\begin{align}\label{GradientIncrementalstep3}
&\Big\langle  
\bd{m}^c_{x,i-1} - \nabla_x J(\bd{x}_{i-1}, \bd{y}_{i-1}),
\nabla_x J(\bd{x}_{i-1}, \bd{y}_{i-1})-\nabla_x J(\bd{x}_{i},\notag 
\\
&  \bd{y}_{i})
 + \nabla^2_{xx}J(\bd{x}_i, \bd{y}_{i})(\bd{x}_{i} - \bd{x}_{i-1}) +
\nabla^2_{xy}J(\bd{x}_i, \bd{y}_{i})
 (\bd{y}_{i} - \bd{y}_{i-1}) 
\Big\rangle  \notag \\
& \overset{(a)}{\le}
\|\bd{m}^c_{x,i-1} - \nabla_x J(\bd{x}_{i-1}, \bd{y}_{i-1})\|
\Big\|
\nabla_x J(\bd{x}_{i-1}, \bd{y}_{i-1})-\notag 
\\
&  \quad \nabla_x J(\bd{x}_{i},\bd{y}_{i})
 + \nabla^2_{xx}J(\bd{x}_i, \bd{y}_{i})(\bd{x}_{i} - \bd{x}_{i-1})+
\nabla^2_{xy}J(\bd{x}_i, \bd{y}_{i}) \notag \\
 &\quad \times
 (\bd{y}_{i} - \bd{y}_{i-1}) 
\Big\| 
\end{align}
where $(a)$ follows from \cblack{the} Cauchy–Schwarz inequality.
From Lemma \ref{meanvalue}, we deduce that 
\begin{align}\label{GradientIncrementalstep4}
&\Big\|
\nabla_x J(\bd{x}_{i-1}, \bd{y}_{i-1})-\nabla_x J(\bd{x}_{i},\bd{y}_{i})
 + \nabla^2_{xx}J(\bd{x}_i, \bd{y}_{i})(\bd{x}_{i} - \bd{x}_{i-1})\notag 
\\
&   +
\nabla^2_{xy}J(\bd{x}_i, \bd{y}_{i})
 (\bd{y}_{i} - \bd{y}_{i-1}) 
\Big\|^2 \notag \\
&
=\Bigg\| \begin{bmatrix}
\nabla_x J(\bx_{i-1}, \by_{i-1})\\
0_{M_2}
\end{bmatrix} -
\begin{bmatrix}
\nabla_x J(\bx_{i}, \by_{i})\\
0_{M_2}
\end{bmatrix}
\notag \\
&\quad +\begin{bmatrix}
\nabla^2_{xx} J(\bx_{i}, \by_{i})& \nabla^2_{xy} J(\bx_{i}, \by_{i})\\
0_{M_2 \times M_1} & 0_{M_2 \times M_2}
\end{bmatrix}
\begin{bmatrix}
    \bx_{i} - \bx_{i-1}\\
   \by_{i} - \by_{i-1}
\end{bmatrix}
\Bigg\|^2 
\notag 
\\
&\le 
\Bigg\| \begin{bmatrix}
\nabla_x J(\bx_{i-1}, \by_{i-1})\\
\nabla_y J(\bx_{i-1}, \by_{i-1})
\end{bmatrix} -
\begin{bmatrix}
\nabla_x J(\bx_{i}, \by_{i})\\
\nabla_y J(\bx_{i}, \by_{i})
\end{bmatrix}
\notag \\
&\quad +\begin{bmatrix}
\nabla^2_{xx} J(\bx_{i}, \by_{i})& \nabla^2_{xy} J(\bx_{i}, \by_{i})\\
\nabla^2_{yx} J(\bx_{i}, \by_{i}) & 
\nabla^2_{yy} J(\bx_{i}, \by_{i})
\end{bmatrix}
\begin{bmatrix}
    \bx_{i} - \bx_{i-1}\\
    \by_{i} -\by_{i-1}
\end{bmatrix}
\Bigg\|^2 
\notag \\
& 
= 
\Big\|
\nabla_z J(\bd{x}_{i-1}, \bd{y}_{i-1})-\nabla_z J(\bd{x}_{i},\bd{y}_{i})
+ \nabla^2_{z}J(\bd{x}_i, \bd{y}_i) \notag \\
&\quad \times 
\begin{bmatrix}
    \bd{x}_{i} - \bd{x}_{i-1} \\
  \bd{y}_{i} - \bd{y}_{i-1}   
\end{bmatrix}
\Big\|^2 \notag  \\
&\le 
\frac{L^2_h}{4}
\Big\| 
\begin{bmatrix}
    \bd{x}_{i} - \bd{x}_{i-1} \\
  \bd{y}_{i} - \bd{y}_{i-1}   
\end{bmatrix}
\Big\|^4
\end{align}
Therefore, the cross-term in 
\eqref{GradientIncrementalstep2}
can be bounded as follows:
\begin{align}\label{GradientIncrementalstep5}
&2(1- \beta_x)^2\Big\langle  
\bd{m}^c_{x,i-1} - \nabla_x J(\bd{x}_{i-1}, \bd{y}_{i-1}),
\nabla_x J(\bd{x}_{i-1}, \bd{y}_{i-1})\notag 
\\
&  -\nabla_x J(\bd{x}_{i},\bd{y}_{i})
 + \nabla^2_{xx}J(\bd{x}_i, \bd{y}_{i})(\bd{x}_{i} - \bd{x}_{i-1})+
\nabla^2_{xy}J(\bd{x}_i, \bd{y}_{i}) \notag  \\& 
\times (\bd{y}_{i} - \bd{y}_{i-1}) 
\Big\rangle  \notag \\
& \le  L_h(1- \beta_x)^2
\|\bd{m}^c_{x,i-1} - \nabla_x J(\bd{x}_{i-1},\bd{y}_{i-1})\| \notag \\
&\quad \times
\Big\| 
\begin{bmatrix}
    \bd{x}_{i} - \bd{x}_{i-1} \\
  \bd{y}_{i} - \bd{y}_{i-1}   
\end{bmatrix}
\Big\|^2  \notag  \\
& \overset{(a)}{\le}
\beta_x(1-\beta_x)^2
\|\bd{m}^c_{x,i-1} - \nabla_x J(\bd{x}_{i-1},\bd{y}_{i-1})\|^2   \notag \\
& \quad +\frac{L^2_h(1-\beta_x)^2}{4\beta_x} \Big\|
\begin{bmatrix}
    \bd{x}_{i} - \bd{x}_{i-1} \\
  \bd{y}_{i} - \bd{y}_{i-1}   
\end{bmatrix}\Big\|^4  \hspace{8em}
\end{align}
where $(a)$ follows from $u^\top v \le \frac{\tau\|u\|^2}{2} + \frac{\|v\|^2}{2\tau}$
and we choose $\tau =2\beta_x$.
On the other hand, for the \cblack{squared term}, we can bound it as follows:
\begin{align}
&\mbE\Big[\Big\| 
(1-\beta_x)\Big[ 
\nabla_x J(\bd{x}_{i-1} , \bd{y}_{i-1})
-\nabla_x J(\bd{x}_{i}, \bd{y}_{i})+ \nabla^2_{xx}Q(\bd{x}_i, \notag\\
& \quad  \bd{y}_{i};
 \boldsymbol{\xi}_{i})
(\bd{x}_{i} - \bd{x}_{i-1}) +
  \nabla^2_{xy}Q(\bd{x}_i, \bd{y}_{i};
 \boldsymbol{\xi}_{i})
 (\bd{y}_{i}  - \bd{y}_{i-1})
\Big] \notag \\&\quad +\beta_x\Big[\nabla_x Q(\bd{x}_i, \bd{y}_{i};
 \boldsymbol{\xi}_{i}) - 
 \nabla_x J(\bd{x}_i, \bd{y}_{i})\Big]
\Big\|^2 \mid \bd{\mc{F}}_{i}\Big]\notag
\\
&\overset{(a)}{\le}
2(1-\beta_x)^2 \mbE\Big[\Big\|
\nabla_x J(\bd{x}_{i-1} , \bd{y}_{i-1})
-\nabla_x J(\bd{x}_{i}, \bd{y}_{i}) \notag\\
& \quad  + \nabla^2_{xx}Q(\bd{x}_i,\bd{y}_{i};
 \boldsymbol{\xi}_{i})
(\bd{x}_{i} - \bd{x}_{i-1}) +
  \nabla^2_{xy}Q(\bd{x}_i, \bd{y}_{i};
 \boldsymbol{\xi}_{i})
  \notag \\&\quad \times (\bd{y}_{i}  - \bd{y}_{i-1})
\Big\|^2 \mid \bd{\mc{F}}_{i}\Big] +2 \beta^2_x \mbE[\|\nabla_x Q(\bd{x}_i, \bd{y}_{i};
 \boldsymbol{\xi}_{i})\notag \\
& \quad - 
 \nabla_x J(\bd{x}_i, \bd{y}_{i})\|^2 \mid \bd{\mc{F}}_{i}] \notag
\\
&\overset{(b)}{\le}
2(1-\beta_x)^2 \mbE\Big[\Big\|
\nabla_x J(\bd{x}_{i-1} , \bd{y}_{i-1})
-\nabla_x J(\bd{x}_{i}, \bd{y}_{i}) \notag\\
& \quad  + \nabla^2_{xx}Q(\bd{x}_i,\bd{y}_{i};
 \boldsymbol{\xi}_{i})
(\bd{x}_{i} - \bd{x}_{i-1}) +
  \nabla^2_{xy}Q(\bd{x}_i, \bd{y}_{i};
 \boldsymbol{\xi}_{i})
  \notag \\&\quad \times (\bd{y}_{i}  - \bd{y}_{i-1})
\Big\|^2 \mid \bd{\mc{F}}_{i}\Big]
+2 \beta^2_x \sigma^2 \notag
\\
\label{GradientIncrementalstep6}
   & \overset{(c)}{\le}
   2(1-\beta_x)^2 \mbE\Big[\Big\| 
\nabla_x J(\bd{x}_{i-1} , \bd{y}_{i-1})
-\nabla_x J(\bd{x}_{i}, \bd{y}_{i})+  \notag\\
& \quad  \nabla^2_{xx}J(\bd{x}_i,\bd{y}_{i})
(\bd{x}_{i} - \bd{x}_{i-1}) +
  \nabla^2_{xy}J(\bd{x}_i, \bd{y}_{i})(\bd{y}_{i}  - \bd{y}_{i-1})\Big\|^2
  \notag \\&\quad   
 \mid \bd{\mc{F}}_{i}\Big]
+
2(1-\beta_x)^2 \mbE\Big[\Big\|
\Big(\nabla^2_{xx}Q(\bd{x}_i,\bd{y}_{i};\bd{\xi}_i)- \nabla^2_{xx}J(\bd{x}_i, \notag \\& \quad  \bd{y}_{i})\Big)
(\bd{x}_{i} - \bd{x}_{i-1})+
\Big(\nabla^2_{xy}Q(\bd{x}_i, \bd{y}_{i};\bd{\xi}_i)- \nabla^2_{xy}J(\bd{x}_i, \bd{y}_{i})\Big)\notag\\
& \quad \times(\bd{y}_{i}  - \bd{y}_{i-1}) \Big\|^2\mid \bd{\mc{F}}_{i}
\Big] +
2 \beta^2_x \sigma^2
\end{align}
where $(a)$ follows from Jensen's inequality,
$(b)$ follows from Assumption \ref{unbiased},
\cblack{and in} $(c)$ we add and subtract the true Hessian  and 
use Assumption \ref{unbiased}.
Similar to 
\eqref{GradientIncrementalstep4}, we get
\begin{align}\label{GradientIncrementalstep7}
\mbE&\Big[\Big\|
\Big(\nabla^2_{xx}Q(\bd{x}_i,\bd{y}_{i};\bd{\xi}_i)- \nabla^2_{xx}J(\bd{x}_i,   \bd{y}_{i})\Big)
(\bd{x}_{i} - \bd{x}_{i-1})+ \notag 
\\ & 
\Big(\nabla^2_{xy}Q(\bd{x}_i, \bd{y}_{i};\bd{\xi}_i)- \nabla^2_{xy}J(\bd{x}_i, \bd{y}_{i})\Big)(\bd{y}_{i}  - \bd{y}_{i-1}) \Big\|^2\mid \bd{\mc{F}}_{i}
\Big] \notag \\
\le& 
\mbE\Big[\Big\|
\Big( \nabla^2_{z}Q(\bd{x}_i,\bd{y}_{i};\bd{\xi}_i)- \nabla^2_{z}J(\bd{x}_i,   \bd{y}_{i})\Big) 
 \begin{bmatrix}
     \bd{x}_i -\bd{x}_{i-1} \\
    \bd{y}_i -\bd{y}_{i-1} 
 \end{bmatrix}
  \Big\|^2
  \notag\\
 &\quad \mid \bd{\mc{F}}_{i}
\Big] \notag \\
\overset{(a)}{\le}& 
\mbE\Big[\Big\|
\Big(\nabla^2_{z}Q(\bd{x}_i,\bd{y}_{i};\bd{\xi}_i)- \nabla^2_{z}J(\bd{x}_i,   \bd{y}_{i})\Big) \Big\|^2\Big\|
 \begin{bmatrix}
     \bd{x}_i -\bd{x}_{i-1} \\
    \bd{y}_i -\bd{y}_{i-1} 
 \end{bmatrix}
  \Big\|^2
  \notag\\
 &\quad \mid \bd{\mc{F}}_{i}
\Big] \notag \\
\overset{(b)}{\le}&
\sigma^2_h \Big\|
 \begin{bmatrix}
     \bd{x}_i -\bd{x}_{i-1} \\
    \bd{y}_i -\bd{y}_{i-1} 
 \end{bmatrix}
  \Big\|^2
\end{align}
where $(a)$ follows from \cblack{the} submultiplicative property of norms, \cblack{and} $(b)$ follows from Assumption \ref{unbiased}.
\cblack{Combining} the results of \eqref{GradientIncrementalstep6}
and \eqref{GradientIncrementalstep7}, we obtain
\begin{align}\label{GradientIncrementalstep8}
\mbE&\Big[\Big\| 
(1-\beta_x)\Big[ 
\nabla_x J(\bd{x}_{i-1} , \bd{y}_{i-1})
-\nabla_x J(\bd{x}_{i}, \bd{y}_{i})+ \nabla^2_{xx}Q(\bd{x}_i, \notag\\
&   \bd{y}_{i};
 \boldsymbol{\xi}_{i})
(\bd{x}_{i} - \bd{x}_{i-1}) +
  \nabla^2_{xy}Q(\bd{x}_i, \bd{y}_{i};
 \boldsymbol{\xi}_{i})
 (\bd{y}_{i}  - \bd{y}_{i-1})
\Big]+\beta_x \notag \\& \times \Big[\nabla_x Q(\bd{x}_i, \bd{y}_{i};
 \boldsymbol{\xi}_{i}) - 
 \nabla_x J(\bd{x}_i, \bd{y}_{i})\Big]
\Big\|^2 \mid \bd{\mc{F}}_{i-1}\Big]\notag\\
\le& 2(1-\beta_x)^2 \mbE\Big[\Big\| 
\nabla_x J(\bd{x}_{i-1} , \bd{y}_{i-1})
-\nabla_x J(\bd{x}_{i}, \bd{y}_{i})+  \notag\\
& \nabla^2_{xx}J(\bd{x}_i,\bd{y}_{i})
(\bd{x}_{i} - \bd{x}_{i-1}) +
  \nabla^2_{xy}J(\bd{x}_i, \bd{y}_{i})(\bd{y}_{i}  - \bd{y}_{i-1})\Big\|^2
  \notag \\&\quad   
 \mid \bd{\mc{F}}_{i-1}\Big]
 +
 2(1-\beta_x)^2\sigma^2_h
 \Big\|
 \begin{bmatrix}
     \bd{x}_i -\bd{x}_{i-1} \\
    \bd{y}_i -\bd{y}_{i-1} 
 \end{bmatrix}
  \Big\|^2
  +2\beta^2_x \sigma^2
\notag  \\
\overset{(a)}{\le} & 
\frac{(1-\beta_x)^2L^2_h}{2}
\Big\|
 \begin{bmatrix}
     \bd{x}_i -\bd{x}_{i-1} \\
    \bd{y}_i -\bd{y}_{i-1} 
 \end{bmatrix}
  \Big\|^4 + 
 2(1-\beta_x)^2\sigma^2_h
  \notag \\ 
  &\times \Big\|
 \begin{bmatrix}
     \bd{x}_i -\bd{x}_{i-1} \\
    \bd{y}_i -\bd{y}_{i-1} 
 \end{bmatrix}
  \Big\|^2+2\beta^2_x \sigma^2
\end{align}
where $(a)$ follows from \eqref{GradientIncrementalstep4}.
Finally, combining the results of 
\eqref{GradientIncrementalstep2}, \eqref{GradientIncrementalstep5}, \eqref{GradientIncrementalstep8}, and taking expectation again, we obtain
\begin{align}
\mbE&\|\bd{m}^c_{x,i} - \nabla_x J(\bd{x}_i, \bd{y}_i)\|^2 \notag \\
\le& 
(1-\beta_x)^2\mbE\| 
\bd{m}^c_{x,i-1} - \nabla_x J(\bd{x}_{i-1}, \bd{y}_{i-1})
\|^2 +\beta_x(1-\beta_x)^2 \notag \\
& \times 
\mbE\|\bd{m}^c_{x,i-1} - \nabla_x J(\bd{x}_{i-1}, \bd{y}_{i-1})\|^2 + \frac{L^2_h(1-\beta_x)^2}{4\beta_x}  \notag \\
&  \times \mbE\Big\|
\begin{bmatrix}
    \bd{x}_{i} - \bd{x}_{i-1} \\
  \bd{y}_{i} - \bd{y}_{i-1}   
\end{bmatrix}\Big\|^4+ \frac{(1-\beta_x)^2L^2_h}{2}
\mbE\Big\|
 \begin{bmatrix}
     \bd{x}_i -\bd{x}_{i-1} \\
    \bd{y}_i -\bd{y}_{i-1} 
 \end{bmatrix}
  \Big\|^4   
  \notag \\ 
  &+ 2(1-\beta_x)^2\sigma^2_h\mbE\Big\|
 \begin{bmatrix}
     \bd{x}_i -\bd{x}_{i-1} \\
    \bd{y}_i -\bd{y}_{i-1} 
 \end{bmatrix}
  \Big\|^2+2\beta^2_x \sigma^2\notag
\\
\label{GradientIncrementalstep9}
\overset{(a)}{\le}  &
(1-\beta_x)\mbE\|\bd{m}^c_{x,i-1} - \nabla_x J(\bd{x}_{i-1}, \bd{y}_{i-1})\|^2
+ \frac{L^2_h}{2\beta_x}  \notag \\ & \times  \mbE\Big\|
 \begin{bmatrix}
     \bd{x}_i -\bd{x}_{i-1} \\
    \bd{y}_i -\bd{y}_{i-1} 
 \end{bmatrix}
  \Big\|^4 + 2(1-\beta_x)^2\sigma^2_h\mbE\Big\|
 \begin{bmatrix}
     \bd{x}_i -\bd{x}_{i-1} \\
    \bd{y}_i -\bd{y}_{i-1} 
 \end{bmatrix}
  \Big\|^2 \notag \\&
 + 2 \beta^2_x \sigma^2
\end{align}
where $(a)$ follows from 
$(1+\beta_x)(1-\beta_x)^2 \le 1 - \beta_x$
and parameter choice $\beta_x \le \frac{1}{2}$ such that  $\frac{(1-\beta_x)^2L^2_h}{2} \le \frac{L^2_h}{4\beta_x}$.
Moving the $\mbE\|\bd{m}^c_{x,i-1} - \nabla_x J(\bd{x}_{i-1}, \bd{y}_{i-1})\|^2$
to the LHS of \eqref{GradientIncrementalstep9},
we arrive at \eqref{gradientincremenal_eq}.

\qed

\begin{Lemma}\label{IterateIncremental}
    Under Assumptions \ref{NC_SC} and  \ref{LipschitzGradient},
    choosing the step size $\mu_y \le \min \{
    \pi_1, \frac{2}{\nu}
    \}$,
    the following result holds \cblack{for} {\normalfont \textbf{HCMM-1}:}
\begin{align} \label{IterateIncremental_eq}
&\mbE \Big[\|\bd{y}_{i+1} -\bd{y}^o(\bd{x}_{i+1}) \|^2
- \|\bd{y}_{i} -\bd{y}^o(\bd{x}_{i}) \|^2 \Big]
\notag \\
&\le
-\frac{\nu\mu_y}{4}
\mbE\| \bd{y}_i - \bd{y}^o(\bd{x}_i)\|^2 -
\pi_1\mu_y
\mbE\|\bd{m}^c_{y,i}\|^2 +
\frac{6\mu_y}{\nu}
\notag \\
& \quad \times \mbE\| \nabla_y J(\bd{x}_i, \bd{y}_i) - \bd{m}^c_{y,i}\|^2
+\frac{6\kappa^2\mu^2_x}{\nu\mu_y} \mbE\|\bd{m}^c_{x,i}\|^2
\end{align}
where $\pi_1 = \frac{1}{2L_f+\nu}$
is a constant.
\end{Lemma}
\noindent\textbf{Proof:}
Adding and subtracting $\bd{y}^o(\bd{x}_i)$
in $\|\bd{y}_{i+1} -\bd{y}^o(\bd{x}_{i+1}) \|^2$,
we have 
\begin{align}\label{IterateIncrementalStep1}
&\|\bd{y}_{i+1} -\bd{y}^o(\bd{x}_{i+1}) \|^2  \\ 
&\overset{(a)}{\le}  (1+ \frac{\nu\mu_y}{4})
\|\bd{y}_{i+1} - \bd{y}^o(\bd{x}_i)\|^2
+ (1+ \frac{4}{\nu\mu_y})
\|\bd{y}^o(\bd{x}_{i+1})\notag\\&
\quad - \bd{y}^o(\bd{x}_{i})\|^2 
\notag \\
&\overset{(b)}{\le} 
(1+ \frac{\nu\mu_y}{4})
\|\bd{y}_{i+1} - \bd{y}^o(\bd{x}_i)\|^2+ 
(1 + \frac{4}{\nu\mu_y})\mu^2_x\kappa^2\|\bd{m}^c_{x,i}\|^2 \notag \hspace{5em}
\end{align}
where $(a)$ follows from the inequality $\|u+v\|^2 \le (1 +\tau)\|u\|^2 + (1+\frac{1}{\tau})\|v\|^2$, \cblack{and} $(b)$ follows from 
Lemma \ref{Danskin}.
For the term $\|\bd{y}_{i+1} - \bd{y}^o(\bd{x}_i)\|^2$, 
using the recursion \cblack{for} $\bd{y}_{i+1}$,
we deduce that 
\begin{align}\label{IterateIncrementalStep2}
&\|\bd{y}_{i+1} - \bd{y}^o(\bd{x}_i)\|^2  \\
& = 
\|\bd{y}_{i} - \bd{y}^o(\bd{x}_i)\|^2 +
2\langle \bd{y}_{i} - \bd{y}^o(\bd{x}_i),
\mu_y\bd{m}^c_{y,i}\rangle
+ \mu^2_y\|\bd{m}^c_{y,i}\|^2 \notag
\end{align}
We proceed to bound the cross-term
$\langle \bd{y}_{i} - \bd{y}^o(\bd{x}_i),
\bd{m}^c_{y,i}\rangle$.
Note that $J(x,y)$ is $\nu$-strongly concave in $y$, thus we have 
\begin{align}
&J(\bd{x}_i, \bd{y}^o(\bd{x}_i)) \notag \\
&\le 
J(\bd{x}_i, \bd{y}_i) 
+ \langle \nabla_y J(\bd{x}_i, \bd{y}_i),
\bd{y}^o(\bd{x}_i) - \bd{y}_i\rangle
- \frac{\nu}{2}\|\bd{y}^o(\bd{x}_i) - \bd{y}_i\|^2 \notag 
\\
&\overset{(a)}{\le} 
J(\bd{x}_i, \bd{y}_i) 
+ \langle \nabla_y J(\bd{x}_i, \bd{y}_i) -\bd{m}^c_{y,i},
\bd{y}^o(\bd{x}_i) - \bd{y}_i - \alpha   \bd{m}^c_{y,i}\rangle \notag \\
& \quad 
+\langle \bd{m}^c_{y,i},
\bd{y}^o(\bd{x}_i) - \bd{y}_i - \alpha \bd{m}^c_{y,i}  \rangle  +
\langle \nabla_y J(\bd{x}_i, \bd{y}_i),
\alpha \bd{m}^c_{y,i}\rangle
\notag \\ &
\quad - \frac{\nu}{2}\|\bd{y}^o(\bd{x}_i) - \bd{y}_i\|^2 \notag
\\
\label{IterateIncrementalStep3}
&\overset{(b)}{\le}
J(\bd{x}_i, \bd{y}_i) 
+ \frac{2}{\nu}\| \nabla_y J(\bd{x}_i, \bd{y}_i) - \bd{m}^c_{y,i}\|^2
+ \frac{\nu}{8}\|\bd{y}_i - \bd{y}^o(\bd{x}_i)  \notag
\\&\quad -  \alpha \bd{m}^c_{y,i}\|^2
- \alpha \|\bd{m}^c_{y,i}\|^2
+\langle \bd{m}^c_{y,i},
\bd{y}^o(\bd{x}_i) - \bd{y}_i  \rangle \notag \\
& \quad +
\langle \nabla_y J(\bd{x}_i, \bd{y}_i),
\alpha \bd{m}^c_{y,i}\rangle
- \frac{\nu}{2}\| \bd{y}_i - \bd{y}^o(\bd{x}_i)\|^2 \notag
\\
&\overset{(c)}{\le}
J(\bd{x}_i, \bd{y}_i) 
+ \frac{2}{\nu}\| \nabla_y J(\bd{x}_i, \bd{y}_i) - \bd{m}^c_{y,i}\|^2
+ \frac{\nu}{4}\|\bd{y}_i - \bd{y}^o(\bd{x}_i)\|^2 \notag
\\&\quad+ \frac{\nu \alpha^2}{4}
\|\bd{m}^c_{y,i}\|^2 
- \alpha \|\bd{m}^c_{y,i}\|^2
+\langle \bd{m}^c_{y,i},
\bd{y}^o(\bd{x}_i) - \bd{y}_i  \rangle \notag \\
& \quad +
\langle \nabla_y J(\bd{x}_i, \bd{y}_i),
\alpha \bd{m}^c_{y,i}\rangle
- \frac{\nu}{2}\| \bd{y}_i - \bd{y}^o(\bd{x}_i)\|^2
\end{align}
where \cblack{in} $(a)$ 
we add and subtract 
$\bd{m}^c_{y,i}, \alpha \bd{m}^c_{y,i}$ ($\alpha$ is a constant) into the cross-term,
 $(b)$ follows from  
$u^\top v \le \frac{\tau}{2}\|u\|^2 + \frac{1}{2\tau} \|v\|^2 $
and we choose $\tau= \frac{4}{\nu}$ for the first cross-term, \cblack{and in}
$(c)$ we use Jensen's inequality.
By Assumption \ref{LipschitzGradient}, 
$-J(x,y)$
in $L_f$-smooth in $y$ for a given $x$,
therefore we have 
\begin{align}\label{IterateIncrementalStep4}
    &- J(\bd{x}_i, \bd{y}_i + \alpha \bd{m}^c_{y,i})  \\
    &\le 
    -J(\bd{x}_i, \bd{y}_i)
    - \langle 
    \nabla_y J(\bd{x}_i, \bd{y}_i),
    \alpha \bd{m}^c_{y,i}
    \rangle+\frac{L_f\alpha^2}{2}\| 
    \bd{m}^c_{y,i}\|^2 \notag
\end{align}
Adding the results  
\eqref{IterateIncrementalStep3}-\eqref{IterateIncrementalStep4} together
and using the fact that 
$J(\bd{x}_i, \bd{y}^o(\bd{x}_i))\ge J(\bd{x}_i, \bd{y}_i + \alpha \bd{m}^c_{y,i})
$,
we obtain 
\begin{align}\label{IterateIncrementalStep5}
&\langle \bd{m}^c_{y,i},
\bd{y}_i - \bd{y}^o(\bd{x}_i) \rangle \\
&  \le \frac{2}{\nu}\| \nabla_y J(\bd{x}_i, \bd{y}_i) - \bd{m}^c_{y,i}\|^2
-\frac{\nu}{4}
\| \bd{y}_i - \bd{y}^o(\bd{x}_i)\|^2
+ (-\alpha  
\notag \\&\quad+ \frac{L_f\alpha^2 }{2}  + \frac{\nu \alpha^2}{4})\|\bd{m}^c_{y,i}\|^2 \notag \\
& \overset{(a)}{=}
\frac{2}{\nu}\| \nabla_y J(\bd{x}_i, \bd{y}_i) - \bd{m}^c_{y,i}\|^2
-\frac{\nu}{4}
\| \bd{y}_i - \bd{y}^o(\bd{x}_i)\|^2
 - \pi_1 \|\bd{m}^c_{y,i}\|^2 \notag
\end{align}
where $(a)$ is obtained by setting
 $\alpha = \frac{2}{2L_f+\nu}$ to get
$-\alpha + \frac{L_f\alpha^2 }{2}  + \frac{\nu \alpha^2}{4}
= -\frac{1}{2L_f +\nu} \triangleq - \pi_1$.
For the coefficient
$-\alpha + \frac{L_f\alpha^2 }{2}  + \frac{\nu \alpha^2}{4}$,
we can verify that it is
 is negative in the interval $\alpha \in (0, \frac{4}{2L_f+\nu})$. 
Inserting the results of 
\eqref{IterateIncrementalStep5}
and \eqref{IterateIncrementalStep2}
into 
\eqref{IterateIncrementalStep1},
we obtain
\begin{align}\label{IterateIncrementalStep6}
 &\|\bd{y}_{i+1} -\bd{y}^o(\bd{x}_{i+1}) \|^2 \notag \\ 
&\le  
(1+\frac{\nu\mu_y}{4})
\Big[ 
(1-\frac{\nu\mu_y}{2})
\| \bd{y}_i - \bd{y}^o(\bd{x}_i)\|^2 
+ \mu^2_y\|\bd{m}^c_{y,i}\|^2 \notag \\
& \quad 
+\frac{4\mu_y}{\nu}\| \nabla_y J(\bd{x}_i, \bd{y}_i) - \bd{m}^c_{y,i}\|^2
 - 2\pi_1\mu_y \|\bd{m}^c_{y,i}\|^2
\Big] +  (1 \notag \\ & \quad 
 +\frac{4}{\nu\mu_y}) \mu^2_x\kappa^2\|\bd{m}^c_{x,i}\|^2 \notag\\
&\le
(1-\frac{\nu\mu_y}{4})
\| \bd{y}_i - \bd{y}^o(\bd{x}_i)\|^2
+ (1+\frac{\nu\mu_y}{4})(-2\pi_1\mu_y + \mu^2_y) \notag \\
& \quad \times \|\bd{m}^c_{y,i}\|^2
+ (1+\frac{\nu\mu_y}{4})\frac{4\mu_y}{\nu}\| \nabla_y J(\bd{x}_i, \bd{y}_i) - \bd{m}^c_{y,i}\|^2 \notag \\
& \quad 
+(1 + \frac{4}{\nu\mu_y})\mu^2_x\kappa^2\|\bd{m}^c_{x,i}\|^2 
\end{align}
\cblack{We} choose the step size $\mu_y$ such that 
\begin{align}
   &(s1)\ \frac{\nu\mu_y}{4} \le \frac{1}{2} \Longrightarrow 
    \mu_y \le \frac{2}{\nu} \notag \\
    & (s1) \
  -2\pi_1\mu_y + \mu^2_y \le -\pi_1\mu_y 
   \Longrightarrow
   \mu_y \le \pi_1 \notag \\
  &(s3)\ 1 \le \frac{2}{\nu \mu_y} \Longrightarrow 
  \mu_y \le \frac{2}{\nu}
   \notag 
\end{align}
where
$(s3)$ implies $1+ \frac{4}{\nu\mu_y} \le \frac{6}{\nu\mu_y}$.
\cblack{Then, relation} \eqref{IterateIncrementalStep6}
can be simplified into 
\begin{align} \label{IterateIncrementalStep7}
&\|\bd{y}_{i+1} -\bd{y}^o(\bd{x}_{i+1}) \|^2 \notag \\
&\le
(1-\frac{\nu\mu_y}{4})
\| \bd{y}_i - \bd{y}^o(\bd{x}_i)\|^2 -
\pi_1\mu_y
\|\bd{m}^c_{y,i}\|^2 +
\frac{6\mu_y}{\nu}
\notag \\
& \quad \times \| \nabla_y J(\bd{x}_i, \bd{y}_i) - \bd{m}^c_{y,i}\|^2
+\frac{6\mu^2_x\kappa^2}{\nu\mu_y} \|\bd{m}^c_{x,i}\|^2
\end{align}
Moving $\| \bd{y}_i - \bd{y}^o(\bd{x}_i)\|^2$ 
to the 
LHS of \eqref{IterateIncrementalStep7} and taking expectations,
we arrive at \eqref{IterateIncremental_eq}.
\qed

In Lemmas \ref{GradientIncremental} and \ref{IterateIncremental}, we established the descent relation for certain incremental terms. These relations are useful for establishing the descent relation \cblack{for} $\bd{\Omega}_{i}$.

\section{Basic Lemmas for Nonconvex-PL Risk Functions}
\label{appendice2}
\begin{Lemma}[\cite{nouiehed2019solving}] \label{Danskin2}
Under\ Assumptions \ref{NC_PL} and \ref{LipschitzGradient},
if $-J(x,y)$ is $L_f$-smooth \cblack{over the block variable $z= \mbox{\rm cat}\{x, y\}$} and $\nu$-PL in $y$
for any fixed $x$,
then:
\begin{itemize}
\item The primal objective  $P(x)$ 
is   \scalebox{0.9}{$L_2 \triangleq (L_f + \frac{\kappa L_f}{2})$}-smooth and 
\begin{equation}
\nabla P(x)
= \nabla_{x} J(x, y^o(x))
\end{equation}
where $\kappa \triangleq \frac{L_f}{\nu}$ is the condition number and  $y^o(x)$ is a maximum 
point of $J(x, y)$ for a fixed $x$, i.e,
$y^o(x)\in { \operatorname{argmax}}_y \  J(x, y)$.
\item $J(x,y)$ satisfies the quadratic growth property in $y$ for any fixed $x$, i.e.,
\begin{equation}
\begin{aligned}
    \max_y J(x,y) - J(x,y)
    \ge \frac{\nu}{2}
    \|y - y^o(x)\|^2, \forall \  y
\end{aligned}
\end{equation}
\end{itemize}
\end{Lemma}
\begin{Lemma}
\label{proof_hcmm1_costvalue_gap}
Under Assumptions \ref{NC_PL}
and 
\ref{LipschitzGradient}, the following result holds \cblack{for} {\normalfont \textbf{HCMM-1}:}
\begin{align}
\label{ncpl_hcmm1}
&\mbE[P(\bd{x}_{i+1}) - P(\bd{x}_{i})] \notag \\
&\le 
-\frac{\mu_x}{2}
\mbE\|\nabla P(\bd{x}_i)\|^2
-\frac{\mu_x}{2}(1-L_2\mu_x)
\mbE\|\bd{m}^c_{x,i}\|^2 \notag \\
&\quad 
+\frac{2\mu_xL^2_f}{\nu} \mbE\Delta_i
+\mu_x \mbE\|\nabla_x J(\bd{x}_i, \bd{y}_i) - \bd{m}^c_{x,i}\|^2
\end{align}
where $\Delta_i \triangleq P(\bx_i) - J(\bx_i,\by_i)$ is the optimality gap for the $y$-variable for a fixed $\bx_i$.
\end{Lemma}
\noindent
\textbf{Proof:}
Because 
$P(x)$ is $L_2$-smooth,
we get 
\begin{align}
&P(\bd{x}_{i+1}) \notag \\
&\le 
P(\bd{x}_{i})
- \mu_x \langle \nabla P(\bd{x}_i),
\bd{m}^c_{x,i}
\rangle
+\frac{L_2\mu^2_x}{2}
\|\bd{m}^c_{x,i}\|^2 \notag \\
&\le
P(\bd{x}_{i})
-\frac{\mu_x}{2}
\|\nabla P(\bd{x}_i)\|^2
-\frac{\mu_x}{2}
\|\bd{m}^c_{x,i}\|^2
\notag \\
&\quad 
+\frac{\mu_x}{2}
\|\nabla P(\bd{x}_i)-
\bd{m}^c_{x,i}\|^2 + \frac{L_2\mu^2_x}{2}\|\bd{m}^c_{x,i}\|^2
\notag \\
&\le 
P(\bd{x}_i)
-\frac{\mu_x}{2}
\|\nabla P(\bd{x}_i)\|^2
-\frac{\mu_x}{2}(1-L_2\mu_x)
\|\bd{m}^c_{x,i}\|^2 \notag \\
&\quad 
+\mu_xL^2_f
\|\bd{y}_i - \bd{y}^o(\bd{x}_i)\|^2
+ \mu_x \|\nabla_x J(\bd{x}_i,\bd{y}_i) - \bd{m}^c_{x,i}\|^2 \notag \\
&\overset{(a)}{\le} 
P(\bd{x}_i)
-\frac{\mu_x}{2}
\|\nabla P(\bd{x}_i)\|^2
-\frac{\mu_x}{2}(1-L_2\mu_x)
\|\bd{m}^c_{x,i}\|^2 \notag \\
&\quad 
+\frac{2\mu_xL^2_f}{\nu}
(P(\bd{x}_i) - J(\bd{x}_i,\bd{y}_i))
+\mu_x \|\nabla_x J(\bd{x}_i, \bd{y}_i) - \bd{m}^c_{x,i}\|^2
\end{align}
where 
$(a)$
follows from the quadratic growth property of $\nu$-PL function. Taking expectation,
moving $P(\bx_i)$
into the left hand side and denoting $\Delta_i \triangleq P(\bx_i) - J(\bx_i,\by_i)$,
we get 
\eqref{ncpl_hcmm1}.\qed
\begin{Lemma}
\label{proof_lemma_hcmm1_optigap}
Under Assumptions \ref{NC_PL}
and 
\ref{LipschitzGradient},
choosing step size 
$\mu_y \le \min\{ \frac{2\kappa^2}{(L_f +L_2)}, \frac{1}{\nu}\}$,
the optimality gap 
$\Delta_i \triangleq P(\bx_i) - J(\bx_i,\by_i)$
produced by running \textbf{HCMM-1}
satisfies 
\begin{align}
&\mbE[\Delta_{i+1} - \Delta_i] \notag \\
&\le   
-\frac{\nu\mu_y}{2}
\mbE\Delta_i
+\frac{2\kappa^2\mu^2_x}{\mu_y}
\mbE\|\bm^c_{x,i}\|^2
-\frac{\mu_y}{2}
(1-L_f\mu_y)\mbE\|\bm^c_{y,i}\|^2
\notag\\
&\quad+\mu_y\mu^2_x L^2_f\mbE\|\bm^c_{x,i}\|^2
+\mu_y
\mbE\|\nabla_y J(\bx_{i}, \by_{i}) - \bm^c_{y,i}\|^2
\end{align}
\end{Lemma}
\noindent
\textbf{Proof:}
Because $-J(\bx_{i+1}, \cdot)$
is $L_f$-smooth, 
we have 
\begin{align}
\label{proof_ncpl_hcmm1_gap_start}
&-J(\bx_{i+1}, \by_{i+1})
\notag \\
&\le 
-J(\bx_{i+1}, \by_{i})
-\mu_y \langle 
\nabla_y J(\bx_{i+1}, \by_{i}),
\bm^c_{y,i}\rangle
+\frac{L_f\mu^2_y}{2}
\|\bm^c_{y,i}\|^2
\notag \\
&\overset{(a)}{\le} 
-J(\bx_{i+1}, \by_{i})
-\frac{\mu_y}{2}\|\nabla_y J(\bx_{i+1}, \by_{i})\|^2-\frac{\mu_y}{2}
\|\bm^c_{y,i}\|^2 \notag \\
&\quad 
+\frac{\mu_y}{2}
\|\nabla_y J(\bx_{i+1}, \by_{i})- 
\nabla_y J(\bx_{i}, \by_{i})+\nabla_y J(\bx_{i}, \by_{i}) \notag \\
&\quad -\bm^c_{y,i}\|^2
+\frac{L_f\mu^2_y}{2}
\|\bm^c_{y,i}\|^2
\notag\\
&
\overset{(b)}{\le} 
-J(\bx_{i+1}, \by_{i})
-\frac{\mu_y}{2}
\|\nabla_y J(\bx_{i+1},\by_{i})\|^2
\notag \\
&\quad 
-\frac{\mu_y}{2}(1-L_f\mu_y)\|\bm^c_{y,i}\|^2
+\mu_y\mu^2_xL^2_f\|\bm^c_{x,i}\|^2
\notag \\
&\quad 
+\mu_y\|\nabla_y J(\bx_{i},\by_{i}) - \bm^c_{y,i}\|^2 
\end{align}
where $(a)$
is derived by rewriting the 
cross-term into squared terms;
$(b)$
follows from 
Jensen's inequality and 
$L_f$-smooth assumption.
By the definition of $\nu$-PL function,
we have 
\begin{align}
-\|\nabla_y J(\bx_{i+1}, \by_i)\|^2
\le -2\nu(P(\bx_{i+1}) - J(\bx_{i+1}, \by_{i}))
\end{align}
Adding $P(\bx_{i+1})$ on both sides of \eqref{proof_ncpl_hcmm1_gap_start},
we can deduce that 
\begin{align}
\label{ncpl_hcmm1_gap}
&P(\bx_{i+1}) - J(\bx_{i+1}, \by_{i+1}) \notag \\
&\le 
(1-\nu\mu_y)
(P(\bx_{i+1}) - J(\bx_{i+1}, \by_{i}))
\notag \\
&\quad 
-\frac{\mu_y}{2}
(1-L_f\mu_y)
\|\bm^c_{y,i}\|^2
+\mu_y\mu^2_xL^2_f\|\bm^c_{x,i}\|^2
\notag \\
&\quad 
+\mu_y\|\nabla_y J(\bx_{i},\by_{i}) - \bm^c_{y,i}\|^2 \notag \\
&
\le 
(1-\nu\mu_y)
(P(\bx_{i+1}) - P(\bx_{i})
+P(\bx_{i}) -
J(\bx_{i},\by_{i})
\notag \\
&\quad 
+J(\bx_{i},\by_{i})
-J(\bx_{i+1}, \by_{i})) \notag 
\\
&\quad 
-\frac{\mu_y}{2}
(1-L_f\mu_y)
\|\bm^c_{y,i}\|^2
+\mu_y\mu^2_xL^2_f\|\bm^c_{x,i}\|^2 \notag \\
&\quad 
+\mu_y\|\nabla_y J(\bx_i,\by_i) - \bm^c_{y,i}\|^2
\end{align}
Note that 
$-J(\cdot, \by_{i})$
is $L_f$-smooth 
and $P(x)$ is $L_2$-smooth,
adding their associated inequality together,
we get 
\begin{align}
&P(\bx_{i+1})
-J(\bx_{i+1}, \by_{i})
\notag \\
&\le 
P(\bx_{i})-J(\bx_i,\by_i)
+\langle
\nabla P(\bx_i)
-\nabla_x J(\bx_{i},\by_{i}),\bx_{i+1}-\bx_{i}
\rangle
\notag \\
&\quad 
+\frac{(L_f+L_2)\mu^2_x}{2}\|\bm^c_{x,i}\|^2 \notag \\
&\overset{(a)}{\le} 
P(\bx_{i})-J(\bx_i,\by_i)
+ \frac{\mu_yL^2_f}{4\kappa^2}\|\by_{i}-\by^o(\bx_{i}) \|^2
\notag \\
&\quad + 
\frac{\kappa^2\mu^2_x}{\mu_y}\|\bm^c_{x,i}\|^2
+\frac{(L_f+L_2)\mu^2_x}{2}\|\bm^c_{x,i}\|^2
\end{align}
where $(a)$ follows from $u^\top v \le \frac{\tau\|u\|^2}{2} + \frac{\|v\|^2}{2\tau}$
and we choose $\tau = \frac{\mu_y}{2\kappa^2}$.
Choosing $\frac{(L_f+L_2)\mu^2_x}{2} \le \frac{\kappa^2\mu^2_x}{\mu_y} \Rightarrow \mu_y \le \frac{2\kappa^2}{(L_2 + L_f)}$
and using the quadratic growth property of the $\nu$-PL function, 
we get 
\begin{align}
&P(\bx_{i+1})  -P(\bx_{i}) 
+J(\bx_{i}, \by_{i})
-J(\bx_{i+1}, \by_{i})
  \notag \\
&\le   \frac{\mu_y L^2_f}{2\kappa^2\nu}
\Delta_i
+ \frac{2\kappa^2\mu^2_x}{\mu_y}
\|\bm^c_{x,i}\|^2 \notag \\
&\le 
\frac{\mu_y\nu}{2}
\Delta_i
+ \frac{2\kappa^2\mu^2_x}{\mu_y}
\|\bm^c_{x,i}\|^2  \quad (\kappa = L_f/\nu)
\end{align}
where we denote $\Delta_i \triangleq P(\bx_{i}) -J(\bx_{i},\by_{i})$.
Combining above results with \eqref{ncpl_hcmm1_gap},
we have 
\begin{align}
&\Delta_{i+1}
\notag \\
&\le 
(1-\nu\mu_y)\Big(
(1+\frac{\mu_y\nu}{2})\Delta_{i}+
\frac{2\kappa^2\mu^2_x}{\mu_y}
\|\bm^c_{x,i}\|^2
\Big)
\notag \\
&\quad 
-\frac{\mu_y}{2}
(1-L_f\mu_y)
\|\bm^c_{y,i}\|^2
+\mu_y\mu^2_xL^2_f\|\bm^c_{x,i}\|^2 \notag \\
&\quad 
+\mu_y\|\nabla_y J(\bx_i,\by_i) - \bm^c_{y,i}\|^2
\notag \\
&\overset{(a)}{\le} 
(1-\frac{\nu\mu_y}{2})
\Delta_i
+ \frac{2\kappa^2\mu^2_x}{\mu_y}
\|\bm^c_{x,i}\|^2
-\frac{\mu_y}{2}
(1-L_f\mu_y)\|\bm^c_{y,i}\|^2
\notag\\
&\quad+\mu_y\mu^2_x L^2_f\|\bm^c_{x,i}\|^2
+\mu_y\|\nabla_y J(\bx_i,\by_i) - \bm^c_{y,i}\|^2
\end{align}
where $(a)$ follows from 
$(1-\nu\mu_y)(1+\frac{\mu_y\nu}{2})
=1-\frac{\mu_y\nu}{2}-\frac{\mu^2_y\nu^2}{2} \le 1-\frac{\mu_y\nu}{2}$ and $\mu_y \le \frac{1}{\nu}$.
Moving 
$\Delta_i$
to the left-hand side and taking expectations, the proof is completed.
\qed

\begin{Lemma}
\label{primal_ncpl}
Under Assumptions  \ref{NC_PL}, and \ref{LipschitzGradient},  the following result holds \cblack{for} {\normalfont \textbf{HCMM-2}:}
\begin{align}
&P(\bd{x}_{i+1}) \notag 
    \\ 
       &\le P(\bd{x}_{i})
      -\frac{\mu_x\|\nabla P(\bd{x}_{i})\|}{3} +
   3\mu_x\|\bd{m}_{x, i} - 
   \nabla_x J(\bd{x}_i, \bd{y}_i)\| \notag \\
   &\quad +3\mu_xL_f\|\bd{y}_i - 
   \bd{y}^o(\bd{x}_i)\| + \frac{L_2\mu^2_x}{2}
   \label{primal_ineq}
\end{align}
\end{Lemma}
\noindent
\textbf{Proof:}
$P(x)$ is $L_2$-smooth, thus
\begin{align}
    &P(\bd{x}_{i+1}) \notag\\
    &\leq P(\bd{x}_{i})
    +\langle \nabla P(\bd{x}_{i}),
   \bd{x}_{i+1} - \bd{x}_i\rangle
    +\frac{L_2}{2}
    \|\bd{x}_{i+1} -\bd{x}_i\|^2 \notag \\
    &\leq
     P(\bd{x}_{i})
    -\mu_x 
    \langle
    \nabla P(\bd{x}_{i}),
     \frac{\bd{m}_{x,i}}{
    \|\bd{m}_{x,i}\|
    }
    \rangle
    +\frac{L_2\mu^2_x}{2} 
\end{align}
For the cross-term, 
we can bound it by considering two cases: \\
Case \textbf{1}:
$\|\nabla P(\bd{x}_i)\| \le
2\|\bd{m}_{x, i} - \nabla P(\bd{x}_i)\|$ 
\begin{align}
   &- 
    \langle
    \nabla P(\bd{x}_{i}),
     \frac{\bd{m}_{x,i}}{
    \|\bd{m}_{x,i}\|
    }
    \rangle \notag \\
   &\overset{(a)}{\le}
   \|\nabla P(\bd{x}_{i})\| \notag \\
   &\le 
   -\frac{\|\nabla P(\bd{x}_{i})\|}{3}
   + \frac{4\|\nabla P(\bd{x}_{i})\|}{3}
   \notag \\
   &\le 
   -\frac{\|\nabla P(\bd{x}_{i})\|}{3}
   + \frac{8\|\bd{m}_{x, i} - \nabla P(\bd{x}_{i})\|}{3}
\end{align}
where 
$(a)$ follows from  \cblack{the} Cauchy-Schwarz
inequality.

\noindent Case \textbf{2}: $\|\nabla P(\bd{x}_i)\| \ge 2\|\bd{m}_{x, i} - \nabla P(\bd{x}_i)\|$
\begin{align}
   &- 
    \langle
    \nabla P(\bd{x}_{i}),
     \frac{\bd{m}_{x,i}}{
    \|\bd{m}_{x,i}\|
    }
    \rangle \notag \\ \notag
    &= 
    - 
    \Big\langle
    \nabla P(\bd{x}_{i}),
     \frac{\bd{m}_{x,i} -\nabla P(\bd{x}_{i}) + 
     \nabla P(\bd{x}_{i})}{
    \|\bd{m}_{x,i}\|
    }
    \Big\rangle \notag \\
    &=
    -\frac{\|P(\bd{x}_{i})\|^2}{\|\bd{m}_{x,i}\|}
    - 
    \Big\langle
    \nabla P(\bd{x}_{i}),
     \frac{\bd{m}_{x,i} -\nabla P(\bd{x}_{i})}{
    \|\bd{m}_{x,i}\|
    }
    \Big\rangle \notag\\
    &\overset{(a)}{\le}
    -\frac{\|\nabla P(\bd{x}_{i})\|^2}{\|\bd{m}_{x,i}\|}
    + \frac{\|\nabla P(\bd{x}_{i})\|\|\bd{m}_{x,i} -\nabla P(\bd{x}_{i})\|}{\|\bd{m}_{x,i}\|} \notag\\
    &\overset{(b)}{\le}
    -\frac{\|\nabla P(\bd{x}_{i})\|^2}{2\|\bd{m}_{x,i}\|} \notag\\
    &\overset{(c)}{\le} 
    -\frac{\|\nabla P(\bd{x}_{i})\|}{3}
\end{align}
where 
$(a)$ follows from \cblack{the} Cauchy-Schwarz
inequality,
$(b)$ follows from $\|\nabla P(\bd{x}_i)\| \ge 2\|\bd{m}_{x, i} - \nabla P(\bd{x}_i)\|$ and 
$(c)$ follows from 
$\|\bd{m}_{x,i}\| \le \|\bd{m}_{x,i}- \nabla P(\bd{x}_i)\|+\|\nabla P(\bd{x}_i)\| \le \frac{3}{2}
\|\nabla P(\bd{x}_i)\|$.
In both cases,
we have 
\begin{align}
    &- 
    \langle
    \nabla P(\bd{x}_{i}),
     \frac{\bd{m}_{x,i}}{
    \|\bd{m}_{x,i}\|
    }
    \rangle \notag\\
    &\le -\frac{\|\nabla P(\bd{x}_{i})\|}{3}
   + \frac{8\|\bd{m}_{x, i} - \nabla P(\bd{x}_{i})\|}{3} \notag\\
   &\overset{(a)}{\le} -\frac{\|\nabla P(\bd{x}_{i})\|}{3} +
   \frac{8\|\bd{m}_{x, i} - 
   \nabla_x J(\bd{x}_i, \bd{y}_i)\|}{3}
   \notag \\
   &\quad +\frac{8\|\nabla_x J(\bd{x}_i, \bd{y}_i)- \nabla P(\bd{x}_{i})\|}{3} \notag\\
    &\overset{(b)}{\le}
    -\frac{\|\nabla P(\bd{x}_{i})\|}{3} +
   3\|\bd{m}_{x, i} - 
   \nabla_x J(\bd{x}_i, \bd{y}_i)\|
   \notag \\
   &\quad +3L_f\|\bd{y}_i - 
   \bd{y}^o(\bd{x}_i)\|
\end{align}
\cblack{where} $(a)$ follows from triangle inequality and 
$(b)$ follows from $L_f$-smooth property of $J(x,y)$.
Putting these results together, we arrive at \eqref{primal_ineq}.

\qed

\begin{Lemma}
\label{deviation_ncpl}
    Under Assumptions \ref{unbiased} and \ref{LipschitzGradient}, choosing $\beta_u \le 1 (u= x \text{ or } y)$ and $\mu_x \le \mu_y$, the following result holds \cblack{for} {\normalfont \textbf{HCMM-2}:}
    \begin{align}
        &\frac{1}{T}\sum_{i=0}^{T-1}\mbE\|\bd{m}_{u,i} - \nabla_u J(\bd{x}_i, \bd{y}_i)\| \notag \\
        &\le \frac{\sigma}{T\beta_u}+\frac{L_h\mu^2_y}{\beta_u} + \frac{2\mu_y\sigma_h}{\sqrt{\beta_u}} +\sigma \sqrt{\beta_u} 
    \end{align}
\end{Lemma}
\noindent\textbf{Proof:}
Inserting the expression of $\bd{m}_{x,i}$ into
$\bd{m}_{x,i} - \nabla_x J(\bd{x}_{i}, \bd{y}_{i})$,
we  get the following expression: 
\begin{align}
        &\bd{m}_{x,i} - \nabla_x J(\bd{x}_{i}, \bd{y}_{i})  \label{mxi_withoutsq}
      \\
    =&(1-\beta_x)\Big[ 
 \bd{m}_{x,i-1} - \nabla_x J(\bd{x}_{i-1}, \bd{y}_{i-1})  +   
  \nabla_x J(\bd{x}_{i-1}, \bd{y}_{i-1})
\notag 
\\
 &-\nabla_x J(\bd{x}_{i}, \bd{y}_{i})
+\nabla^2_{xx} J(\bd{x}_i, \bd{y}_{i})
(\bd{x}_{i} - \bd{x}_{i-1}) +
\nabla^2_{xy}J(\bd{x}_i, \bd{y}_{i})
 \notag \\
&\times (\bd{y}_{i} - \bd{y}_{i-1})+\Big(\nabla^2_{xx}Q(\bd{x}_i, \bd{y}_{i};
 \boldsymbol{\xi}_{i})-\nabla^2_{xx} J(\bd{x}_i, \bd{y}_{i})\Big)
 \notag \\&\times (\bd{x}_{i} - \bd{x}_{i-1})+
\Big(\nabla^2_{xy}Q(\bd{x}_i, \bd{y}_{i};
\boldsymbol{\xi}_{i})-\nabla^2_{xy}J(\bd{x}_i, \bd{y}_{i})\Big)
 \notag \\
 &\times (\bd{y}_{i} - \bd{y}_{i-1})
 \Big] + 
 \beta_x \Big(\nabla_x Q(\bd{x}_i,\bd{y}_i;\bd{\xi}_{i}) - \nabla_x J(\bd{x}_i, \bd{y}_i)\Big) \notag
\end{align}
For notational convenience, we define 
\begin{align}
    \widetilde{\bd{m}}_{x,i} &\triangleq  \bd{m}_{x,i} - \nabla_x J(\bd{x}_{i}, \bd{y}_{i}) \notag\\
Z_{x,i} &\triangleq \nabla_x J(\bd{x}_{i-1}, \bd{y}_{i-1})
-\nabla_x J(\bd{x}_{i}, \bd{y}_{i}) +\nabla^2_{xx} J(\bd{x}_i, \bd{y}_{i}) \notag \\
&\quad  \times 
(\bd{x}_{i} - \bd{x}_{i-1}) +
\nabla^2_{xy}J(\bd{x}_i, \bd{y}_{i})
 (\bd{y}_{i} - \bd{y}_{i-1})  \notag \\
\bd{W}_{x,i} &\triangleq\Big(\nabla^2_{xx}Q(\bd{x}_i, \bd{y}_{i};
 \boldsymbol{\xi}_{i})-\nabla^2_{xx} J(\bd{x}_i, \bd{y}_{i})\Big)
(\bd{x}_{i} - \bd{x}_{i-1}) \notag \\& \quad +
\Big(\nabla^2_{xy}Q(\bd{x}_i, \bd{y}_{i};
\boldsymbol{\xi}_{i})-\nabla^2_{xy}J(\bd{x}_i, \bd{y}_{i})\Big)
 (\bd{y}_{i} - \bd{y}_{i-1})  \notag\\
\bd{s}_{x,i}&\triangleq\nabla_x Q(\bd{x}_i,\bd{y}_i;\bd{\xi}_{i}) - \nabla_x J(\bd{x}_i, \bd{y}_i) \notag
\end{align}
Then we can rewrite the expression \eqref{mxi_withoutsq}
as 
\begin{align}
\label{simplied_mxi}
   \widetilde{\bd{m}}_{x,i} = (1-\beta_x)
   (\widetilde{\bd{m}}_{x,i-1}
   +Z_{x,i}+ \bd{W}_{x,i}) +\beta_x \bd{s}_{x,i}
\end{align}
Iterating \eqref{simplied_mxi} from $i$ to $0$, we get
\begin{align}
\label{iterated_mxi}
\widetilde{\bd{m}}_{x,i}  
=&(1-\beta_x)^i\widetilde{\bd{m}}_{x,0}
+ \sum_{ j=1}^{i}
(1-\beta_x)^j
Z_{x,i-j+1} \notag \\
&+\sum_{j=1}^{i}
(1-\beta_x)^j\bd{W}_{x,i-j+1}
+ \beta_x \sum_{j=0}^{i-1}
(1-\beta_x)^j \bd{s}_{x,i-j}
\end{align}
Taking the $\ell_2$-norm \cblack{of} \eqref{iterated_mxi}
and using the triangle inequality, we get 
\begin{align}
\label{noexpect_mxi}
&\| \widetilde{\bd{m}}_{x,i} \| 
=\underbrace{(1-\beta_x)^i\|\widetilde{\bd{m}}_{x,0}\|}_{\bd{A}}
+ \underbrace{\Big\|\sum_{ j=1}^{i}
(1-\beta_x)^j
Z_{x,i-j+1}\Big\|}_{B} \notag \\
&+\underbrace{\Big\|\sum_{j=1}^{i}
(1-\beta_x)^j\bd{W}_{x,i-j+1}\Big\|}_{\bd{C}}
+ \underbrace{\Big\|\beta_x \sum_{j=0}^{i-1}
(1-\beta_x)^j \bd{s}_{x,i-j}\Big\|}_{\bd{D}}   
\end{align}
Setting $\bd{m}_{x,0} = \nabla_x Q(\bd{x}_0,\bd{y}_0;\bd{\xi}_0)$ and $\bd{m}_{y,0} = \nabla_y Q(\bd{x}_0,\bd{y}_0;\bd{\xi}_0)$,
we can bound $\mbE\bd{A}$ as 
\begin{align}
     &\mbE\bd{A}= \mbE\sqrt{\bd{A}^2} \notag \\
     & \overset{(a)}{\le} (1-\beta_x)^i\sqrt{\mbE\|\bm_{x,0} - \nabla_x J(\bx_{0}, \by_{0})\|^2} \notag \\
    &\le 
    (1-\beta_x)^i \notag \\
    &\quad \times \sqrt{\mbE(\|\bm_{x,0} - \nabla_x J(\bx_{0}, \by_{0})\|^2 + \|\bm_{y,0} - \nabla_y J(\bx_{0}, \by_{0})\|^2)} 
     \notag \\
     &\le (1-\beta_x)^i\sigma
\end{align}
where $(a)$ follows from Jensen's inequality for 
concave function.
Using  \eqref{GradientIncrementalstep4} and $\mu_x \le \mu_y$, we have 
\begin{align}
    \|Z_{x,i-j+1}\| \le  \frac{L_h}{2} \Big\| 
\begin{bmatrix}
    \bd{x}_{i-j+1} - \bd{x}_{i-j} \\
  \bd{y}_{i-j+1} - \bd{y}_{i-j}  
\end{bmatrix}
\Big\|^2 \le L_h \mu^2_y
\end{align}
We then bound $B$ as 
\begin{align}
B \le \sum_{ j=1}^{i}
(1-\beta_x)^j\|
Z_{x,i-j+1}\| &\le \frac{L_h\mu^2_y}{\beta_x}
\end{align}
For $\mbE\bd{C}$, we have
\begin{align}
&\mbE\bd{C}  \notag \\
&\le \mbE\|\sum_{j=1}^{i}
(1-\beta_x)^j\bd{W}_{x,i-j+1}\| \notag \\
&\overset{(a)}{\le} \sqrt{\mbE\|\sum_{j=1}^{i}
(1-\beta_x)^j\bd{W}_{x,i-j+1}\|^2}  \notag \\
&\overset{(b)}{\le} 
\sqrt{\sum_{j=1}^{i}(1-\beta_x)^{2j}\mbE\|
\bd{W}_{x,i-j+1}\|^2} \notag \\
&\overset{(c)}{\le} \sqrt{\sum_{j=1}^{i}(1-\beta_x)^{2j}2\mu^2_y\sigma^2_h} \notag \\
&\le 2\mu_y\sigma_h \sqrt{\frac{1}{1-(1-\beta_x)^2}} \notag \\
&\overset{(d)}{\le} \frac{2\mu_y\sigma_h}{\sqrt{\beta_x}}
\end{align}
where $(a)$ is due to 
Jensen's inequality, \cblack{in}
$(b)$ we expand the squared norm and eliminate the cross-term using the fact \cblack{that} $\{\bd{\xi}_i\}$ is independent over iterations, \cblack{in} $(c)$ we choose $\mu_x \le \mu_y$ and use Assumption \ref{unbiased}, \cblack{and in} $(d)$ we choose $\beta_x \le 1$.
Similarly, we can bound $\mbE\bd{D}$ as 
\begin{align}
&\mbE\bd{D}     \notag \\
&\le \beta_x \sigma \sqrt{\frac{1}{1-(1-\beta_x)^2}} \le \sigma \sqrt{\beta_x}
\end{align}
Finally, it holds that $\forall i \ge 0$
\begin{align}
\label{expected_mxi_re}
    \mbE\| \widetilde{\bd{m}}_{x,i} \| 
    \le (1-\beta_x)^i \sigma + \frac{L_h\mu^2_y}{\beta_x} + \frac{2\mu_y\sigma_h}{\sqrt{\beta_x}} +\sigma \sqrt{\beta_x}
\end{align}
Averaging \eqref{expected_mxi_re} over iterations,
we get 
\begin{align}
    &\frac{1}{T}\sum_{i=0}^{T-1}\mbE\| \widetilde{\bd{m}}_{x,i} \| \notag \\
    &\le \frac{1}{T}\sum_{i=0}^{T-1}(1-\beta_x)^i \sigma + \frac{L_h\mu^2_y}{\beta_x} + \frac{2\mu_y\sigma_h}{\sqrt{\beta_x}} +\sigma \sqrt{\beta_x} \notag \\
    &\overset{(a)}{\le} \frac{\sigma}{T\beta_x}+\frac{L_h\mu^2_y}{\beta_x} + \frac{2\mu_y\sigma_h}{\sqrt{\beta_x}} +\sigma \sqrt{\beta_x} 
\end{align}
where $(a)$ follows from $\sum_{i=0}^{T-1}(1-\beta_x)^i = (1-\beta_x)^0(1-(1-\beta_x^T))/(1-(1-\beta_x)) \le 1/\beta_x$
The result holds similarly for $\frac{1}{T}\sum_{i=0}^{T-1}\mbE\| \widetilde{\bd{m}}_{y,i}\|$.
\qed

\begin{Lemma}
\label{dual_gap_ncpl}
    Under Assumptions \ref{NC_PL} and \ref{LipschitzGradient}, choosing $\mu_x \le \min \{\mu_y, \frac{\mu_y}{6\kappa}\}$, the following result holds \cblack{for} {\normalfont \textbf{HCMM-2}:}
    \begin{align}
    \label{result_dual_gap_ncpl}
   &\frac{1}{T}\sum_{i=0}^{T-1}
   \| \bd{y}_{i} - \bd{y}^{o}(\bd{x}_{i})\|
    \\
   &\le 
\frac{2\|\bd{y}_{0} - \bd{y}^{o}(\bd{x}_{0})\|}{T}
   + \frac{6 \Delta_0}{\mu_y\nu T} \notag \\
&\quad +
   \frac{18}{\nu T}
\sum_{i=0}^{T-1} \|\nabla_y J(\bd{x}_{i}, \bd{y}_{i}) - \bd{m}_{y,i}\| +(\frac{30L_2}{\nu}+2)\mu_y \notag
\end{align}
where $\Delta_0 = P(\bx_0) - J(\bx_0,\by_0)$.
\end{Lemma}
\noindent\textbf{Proof:}
Because $-J(x,y)$ is $L_f$-smooth, we have 
\begin{align}
\label{primal_step1_nc_pl}
    &-J(\bd{x}_{i+1}, \bd{y}_{i+1})
    \notag \\
    &\le  -J(\bd{x}_{i+1}, \bd{y}_{i})
    -\langle \nabla_y J(\bd{x}_{i+1}, \bd{y}_{i}), \mu_y \frac{\bd{m}_{y,i}}{\|\bd{m}_{y,i}\|}\rangle + \frac{L_f\mu^2_y}{2} \notag \\
    &\overset{(a)}{\le} -J(\bd{x}_{i+1}, \bd{y}_{i})
    - \frac{\mu_y\|\nabla_y J(\bd{x}_{i+1}, \bd{y}_i)\|}{3}
    \notag \\
    &\quad + \frac{8\mu_y\|\nabla_y J(\bd{x}_{i+1}, \bd{y}_i) - \bd{m}_{y,i}\|}{3}
    + \frac{L_f \mu^2_y}{2} \notag \\
    & \overset{(b)}{\le }
    -J(\bd{x}_{i+1}, \bd{y}_{i}) 
    - \frac{\mu_y \nu\|\bd{y}_i -\bd{y}^o(\bd{x}_{i+1})\|}{3}
    \notag \\
    &\quad +
    \frac{8\mu_y\|\nabla_y J(\bd{x}_{i+1}, \bd{y}_i) - \nabla_y J(\bd{x}_{i}, \bd{y}_{i})\|}{3} \notag \\
    &\quad + \frac{8\mu_y\|\nabla_y J(\bd{x}_{i}, \bd{y}_{i}) - \bd{m}_{y,i}\|}{3} + \frac{L_f\mu^2_y}{2} \notag \\
    &
    \overset{(c)}{\le} -J(\bd{x}_{i+1}, \bd{y}_{i}) 
    - \frac{\mu_y \nu\|\bd{y}_i -\bd{y}^o(\bd{x}_{i+1})\|}{3} +\frac{8L_f\mu_y\mu_x}{3} \notag \\
    &\quad + \frac{8\mu_y\|\nabla_y J(\bd{x}_{i}, \bd{y}_{i}) - \bd{m}_{y,i}\|}{3} + \frac{L_f\mu^2_y}{2}
\end{align}
where $(a)$ is derived using a similar analysis \cblack{to} Lemma \ref{primal_ncpl},
$(b)$ 
is due to the triangle inequality,
and  
\begin{align}
&\|\nabla_y J(\bd{x}_{i+1}, \bd{y}_i)\| \notag \\
&\ge \sqrt{2\nu(\max_y J(\bd{x}_{i+1}, y) - J(\bd{x}_{i+1}, \bd{y}_i))} \notag \quad \text{ (PL definition) } \\
&\ge \sqrt{\nu^2\|\bd{y}_i - \bd{y}^o(\bd{x}_{i+1})\|^2} \notag \quad \text{ (Quadratic growth) }
\\&\ge
\nu\|\bd{y}_i - \bd{y}^o(\bd{x}_{i+1})\|
\end{align}
$(c)$ of \eqref{primal_step1_nc_pl} is due to $L_f$-smooth assumption.
For simplicity, we denote $\Delta_i \triangleq P(\bd{x}_{i}) - J(\bd{x}_i, \bd{y}_i)$.
Adding $P(\bd{x}_{i+1})$
\cblack{to} both sides of \eqref{primal_step1_nc_pl} and choosing $\mu_x \le \mu_y$, we  deduce that 
\begin{align}
& \Delta_{i+1} \notag \\
& \overset{(a)}{\le} 
P(\bd{x}_{i+1})
-J(\bd{x}_{i+1}, \bd{y}_i)- \frac{\mu_y \nu\|\bd{y}_i -\bd{y}^o(\bd{x}_{i+1})\|}{3}
\notag \\
&\quad + \frac{8\mu_y\|\nabla_y J(\bd{x}_{i}, \bd{y}_{i}) - \bd{m}_{y,i}\|}{3}+4L_f \mu^2_y \notag \\
& \overset{(b)}{\le}
P(\bd{x}_{i+1}) - J(\bd{x}_i, \bd{y}_{i}) + J(\bd{x}_i, \bd{y}_{i})
-J(\bd{x}_{i+1}, \bd{y}_i)+ P(\bd{x}_i)
 \notag \\
&\quad -P(\bd{x}_i)- \frac{\mu_y \nu\|\bd{y}_i -\bd{y}^o(\bd{x}_{i+1})\|}{3}
\notag    \notag \\
&\quad + \frac{8\mu_y\|\nabla_y J(\bd{x}_{i}, \bd{y}_{i}) - \bd{m}_{y,i}\|}{3}+4L_f \mu^2_y \notag \\
& \le 
\Delta_i + P(\bd{x}_{i+1}) -P(\bd{x}_i)
+ J(\bd{x}_i, \bd{y}_i) - J(\bd{x}_{i+1}, \bd{y}_i) \notag \\
&\quad - \frac{\mu_y \nu\|\bd{y}_i -\bd{y}^o(\bd{x}_{i+1})\|}{3}
+ \frac{8\mu_y\|\nabla_y J(\bd{x}_{i}, \bd{y}_{i}) - \bd{m}_{y,i}\|}{3} \notag \\
&\quad +4L_f \mu^2_y  \label{dual_step1_ncpl}
\end{align}
where $(a)$ is due to $\mu_x \le \mu_y$, \cblack{in}
$(b)$ we add and subtract 
$J(\bd{x}_i,\bd{y}_i)$
and $P(\bd{x}_i)$.
\textcolor{black}{Because $P(x)$ is $L_2$-smooth and 
$-J(\cdot, \by_{i})$
is $L_2$-smooth,
for $P(\bx_{i+1})-P(\bx_{i})+J(\bd{x}_i, \bd{y}_{i}) - J(\bd{x}_{i+1}, \bd{y}_i)$,
we can bound it as follows} 
\begin{align}
&P(\bx_{i+1}) -  J(\bx_{i+1}, \by_{i}) \notag \\
&\le 
P(\bx_{i}) - J(\bx_{i}, \by_{i})
+
\langle
\nabla P(\bx_{i}) -  \nabla_x J(\bx_{i}, \by_{i}), \bx_{i+1} - \bx_{i}\rangle
\notag \\
&\quad 
+\frac{(L_2+L_f)\mu^2_x}{2} \notag \\
&\overset{(a)}{\le} 
P(\bx_{i}) - J(\bx_{i}, \by_{i})
+\|\nabla P(\bx_{i}) -  \nabla_x J(\bx_{i}, \by_{i})\|\|\bx_{i+1} -\bx_{i}\|
\notag \\
&\quad +\frac{(L_2+L_f)\mu^2_x}{2}  \notag \\
&\overset{(b)}{\le} 
P(\bx_{i}) - J(\bx_{i}, \by_{i})
+L_f\mu_x\|\by_i -\by^o(\bx_{i})\|
+\frac{(L_2+L_f)\mu^2_x}{2} 
\end{align}
where $(a)$
and $(b)$ follows from Cauchy-Schwarz inequality and $L_f$-smooth property, respectively.
Combining the above results with \eqref{dual_step1_ncpl},
and using the fact that $L_2 > L_f, \mu_y \ge \mu_x$
we get 
\begin{align}
&\Delta_{i+1} \notag  \\
&\le 
\Delta_i + \mu_xL_f\|\bd{y}_i- \bd{y}^o(\bd{x}_i)\|
- \frac{\mu_y \nu\|\bd{y}_i -\bd{y}^o(\bd{x}_{i+1})\|}{3} \notag\\
&\quad 
+ 3\mu_y\|\nabla_y J(\bd{x}_{i}, \bd{y}_{i}) - \bd{m}_{y,i}\|+5L_2\mu^2_y 
\label{dual_step5_ncpl}
\end{align}
Therefore,
\begin{align}
    &\|\bd{y}_i -\bd{y}^o(\bd{x}_{i+1})\| \notag  \\
    &\le  \frac{3(\Delta_i-\Delta_{i+1})}{\mu_y\nu}
    + \frac{3\mu_x\kappa}{\mu_y}\|\bd{y}_i- \bd{y}^o(\bd{x}_i)\| \quad (\kappa = L_f/\nu)\notag \\
   &\quad +\frac{9\|\nabla_y J(\bd{x}_{i}, \bd{y}_{i}) - \bd{m}_{y,i}\|}{\nu}  +\frac{15L_2}{\nu}\mu_y
\end{align}
Furthermore, we have
\begin{align}
   &\| \bd{y}_{i+1} - \bd{y}^{o}(\bd{x}_{i+1})\| \notag \\
& = \| \bd{y}_{i} + \frac{\mu_y \bd{m}_{y,i}}{\|\bd{m}_{y,i}\|} - \bd{y}^{o}(\bd{x}_{i+1})\| \notag \\
&\le \| \bd{y}_{i}  -  \bd{y}^{o}(\bd{x}_{i+1})
\| + \mu_y \notag \\
&\le 
\frac{3(\Delta_i-\Delta_{i+1})}{\mu_y\nu}
    + \frac{3\mu_x\kappa}{\mu_y}\|\bd{y}_i- \bd{y}^o(\bd{x}_i)\| \notag \\
&\quad +\frac{9\|\nabla_y J(\bd{x}_{i}, \bd{y}_{i}) - \bd{m}_{y,i}\|}{\nu}
    +(\frac{15L_2}{\nu}+1)\mu_y
\end{align}
Choosing $\frac{3\mu_x \kappa}{\mu_y} \le \frac{1}{2} \rightarrow \mu_x \le \frac{\mu_y}{6\kappa}$, we have 
\begin{align}
    & \| \bd{y}_{i+1} - \bd{y}^{o}(\bd{x}_{i+1})\| \notag \\
    &\le \| \bd{y}_{i} - \bd{y}^{o}(\bd{x}_{i})\|-\| \bd{y}_{i+1} - \bd{y}^{o}(\bd{x}_{i+1})\| +
    \frac{6(\Delta_i-\Delta_{i+1})}{\mu_y\nu} \notag \\
    &\quad  +\frac{18\|\nabla_y J(\bd{x}_{i}, \bd{y}_{i}) - \bd{m}_{y,i}\|}{\nu}
    +(\frac{30L_2}{\nu}+2)\mu_y
\end{align}
Averaging the above inequality over iterations
and telescoping the terms regarding $\|\by_i - \by^o(\bx_i)\|$ and $\Delta_i$, we arrive at \eqref{result_dual_gap_ncpl}.

\qed

\vspace{-0.5em}
\section{PROOF OF THEOREM \ref{Maintheorem} FOR HCMM-1}
\label{proofMaintheorem}
Subtracting
$\bd{\Omega}_{i}$
from $\bd{\Omega}_{i+1}$,
we \cblack{get} 
\begin{align} 
&\bd{\Omega}_{i+1} - \bd{\Omega}_{i} \notag \\
=&\mathbb{E}
\Big[ \Big(P(\bd{x}_{i+1}) - P(\bd{x}_{i})\Big) 
+ \eta \Big(\|\bd{y}^o(\bd{x}_{i+1}) - \bd {y}_{i+1} \|^2 - \|\bd{y}^o(\bd{x}_i)  \notag \\
 &- \bd {y}_i \|^2\Big) + \gamma
\Big(\|\bd{m}^c_{x,i+1} - \nabla_x J(\bd{x}_{i+1}, \bd{y}_{i+1})\|^2-\|\bd{m}^c_{x,i}      \notag\\
& - \nabla_x J(\bd{x}_{i},\bd{y}_{i})\|^2\Big)  + \gamma \Big(\|\bd{m}^c_{y,i+1}- \nabla_y J(\bd{x}_{i+1}, \bd{y}_{i+1})\|^2 
\notag \\
&   - \|\bd{m}^c_{y,i}  - \nabla_y J(\bd{x}_{i},\bd{y}_{i})\|^2\Big)\Big] \notag 
\\
\label{TheoremStep1}
\overset{(a)}{\le}&
\mbE \Bigg[ \Big(
\mu_xL^2_f\|\bd{y}^o(\bd{x}_i) - \bd{y}_i\|^2 +  \mu_x \| \nabla_x J(\bd{x}_i, \bd{y}_i) -\bd{m}^c_{x,i}\|^2 \notag \\
&  
-\frac{\mu_x}{4}\|\bd{m}^c_{x,i}\|^2 \Big)
+ \eta \Big(
-\frac{\nu\mu_y}{4}
\| \bd{y}_i - \bd{y}^o(\bd{x}_i)\|^2 \notag \\&
-
\pi_1\mu_y
\|\bd{m}^c_{y,i}\|^2 +
\frac{6\mu_y}{\nu} \| \nabla_y J(\bd{x}_i, \bd{y}_i) - \bd{m}^c_{y,i}\|^2
\notag \\
&   
+\frac{6\kappa^2\mu^2_x}{\nu\mu_y}\|\bd{m}^c_{x,i}\|^2 \Big)
+ \gamma
\Big(
-\beta_x\|\bd{m}^c_{x,i} - \nabla_x 
J(\bd{x}_i, \bd{y}_{i})\|^2 \notag 
\\
&
-\beta_y\|\bd{m}^c_{y,i} - \nabla_y 
J(\bd{x}_i, \bd{y}_{i})\|^2
+ (\frac{L^2_h}{2\beta_x} +\frac{L^2_h}{2\beta_y})  \notag \\ &\times
\Big\|\begin{bmatrix}
    \bd{x}_{i+1} - \bd{x}_i \\
    \bd{y}_{i+1} - \bd{y}_i
\end{bmatrix}\Big\|^4
+ 2((1-\beta_x)^2 +(1-\beta_y)^2) \sigma^2_h
\notag \\&
\times 
\Big\|\begin{bmatrix}
    \bd{x}_{i+1} - \bd{x}_i \\
    \bd{y}_{i+1} - \bd{y}_i
\end{bmatrix}\Big\|^2 
+ 2 (\beta^2_x+\beta^2_y) \sigma^2
\Big)
\Bigg] 
\end{align}
where $(a)$ follows from {Lemmas \ref{FunctionIncremental}-\ref{IterateIncremental}.
For brevity,
we choose $\beta_x =\beta_y = \beta$ and denote 
\begin{align}
\widetilde{\bd{b}}_{i} &\triangleq \|\bd{y}^o(\bd{x}_i) - \bd{y}_i\|^2  \\
\widetilde{\bd{m}}^c_{x,i} &\triangleq  \| \nabla_x J(\bd{x}_i, \bd{y}_i) - \bd{m}^c_{x,i}\|^2 \label{proof_hcmm1_definition_mcxi} \\
\widetilde{\bd{m}}^c_{y,i} &\triangleq  \| \nabla_y J(\bd{x}_i, \bd{y}_i) - \bd{m}^c_{y,i}\|^2 \label{proof_hcmm1_definition_mcyi} 
\end{align}
Thus, \cblack{relation}
\eqref{TheoremStep1}
 can be rewritten as 
 \begin{align}\label{TheoremStep2}
&\bd{\Omega}_{i+1} - \bd{\Omega}_{i} \notag \\
=&  
\mbE \Big[ \Big(\mu_x L^2_f -\frac{\eta \mu_y\nu }{4}\Big) \widetilde{\bd{b}}_{i}+ \Big(\mu_x 
- \gamma\beta\Big)\widetilde{\bd{m}}^c_{x,i}
+ \Big(\frac{6\eta\mu_y}{\nu} \notag \\&
-\gamma \beta\Big) \widetilde{\bd{m}}^c_{y,i}
+ \Big( -\frac{\mu_x}{4} + \frac{6 \kappa^2 \eta\mu^2_x}{\nu \mu_y}\Big)\|\bd{m}^c_{x,i}\|^2
- \pi_1\eta\mu_y \notag \\
&\times \|\bd{m}^c_{y,i}\|^2
+ \frac{L^2_h\gamma}{\beta}
\Bigg\|\begin{bmatrix}
    \bd{x}_{i+1} - \bd{x}_i \\
    \bd{y}_{i+1} - \bd{y}_i
\end{bmatrix}\Bigg\|^4 
+ 4 (1-\beta)^2\gamma \sigma^2_h \notag \\
&\times\Bigg\|
\begin{bmatrix}
    \bd{x}_{i+1} - \bd{x}_i \\
    \bd{y}_{i+1} - \bd{y}_i
\end{bmatrix}\Bigg\|^2
+ 4 \gamma\beta^2\sigma^2
\Bigg]
\end{align}
\cblack{where}
\begin{align}
\label{proof_eq_hcmm1_order}
\Bigg\|\begin{bmatrix}
    \bd{x}_{i+1} - \bd{x}_i \\
    \bd{y}_{i+1} - \bd{y}_i
\end{bmatrix}\Bigg\|^2  
&= \mu^2_x \|\bd{m}^c_{x,i}\|^2 
+ \mu^2_y \|\bd{m}^c_{y ,i}\|^2\notag  \\
\Bigg\|\begin{bmatrix}
    \bd{x}_{i+1} - \bd{x}_i \\
    \bd{y}_{i+1} - \bd{y}_i
\end{bmatrix}\Bigg\|^4
 &= \Big(\mu^2_x \|\bd{m}^c_{x,i}\|^2 
+ \mu^2_y \|\bd{m}^c_{y ,i}\|^2\Big)^2 \notag\\
&\le 2\mu^4_x\|\bd{m}^c_{x,i}\|^4 + 2 \mu^4_y
\|\bd{m}^c_{y,i}\|^4
\end{align}
Therefore, choosing $\beta<1$, \cblack{relation} \eqref{TheoremStep2}
can be rewritten as 
\begin{align} \label{TheoremStep3}
&\bd{\Omega}_{i+1} - \bd{\Omega}_{i} \notag \\
\overset{(a)}{\le}&  
\mbE \Big[ \Big(\mu_x L^2_f -\frac{\eta \mu_y\nu }{4}\Big) \widetilde{\bd{b}}_{i}+ \Big(\mu_x 
- \gamma\beta\Big)\widetilde{\bd{m}}^c_{x,i}
+ \Big(\frac{6\eta\mu_y}{\nu} -\gamma \beta\Big)\notag \\&
\times \widetilde{\bd{m}}^c_{y,i}
+ \Big( -\frac{\mu_x}{4} + \frac{6 \kappa^2 \eta\mu^2_x}{\nu \mu_y} +4 \sigma^2_h \mu^2_x\gamma\Big)\|\bd{m}^c_{x,i}\|^2
 \notag \\
&+(- \pi_1\eta\mu_y +4\sigma^2_h\mu^2_y\gamma )\|\bd{m}^c_{y,i}\|^2
+
\frac{2L^2_h\gamma\mu^4_x}{\beta}
\|\bd{m}^c_{x,i}\|^4 \notag\\
&+\frac{2L^2_h\gamma\mu^4_y}{\beta}
\|\bd{m}^c_{y,i}\|^4
+ 4\gamma \beta^2\sigma^2
\Big] 
\end{align}
Due to the clipping procedure, 
we have $\|\bd{m}^c_{x,i}\| \le 
N_1 \Longrightarrow \|\bd{m}^c_{x,i}\|^4 \le N_1^2 \|\bd{m}^c_{x,i}\|^2$.
Choosing $\mu_x \le \mu_y$,
we obtain 
\begin{align} \label{TheoremStep4}
&\bd{\Omega}_{i+1} - \bd{\Omega}_{i} \notag \\
\le&  
\mbE \Big[ \Big(\mu_y L^2_f -\frac{\eta \mu_y\nu }{4}\Big) \widetilde{\bd{b}}_{i}+ \Big(\mu_y 
- \gamma\beta\Big)\widetilde{\bd{m}}^c_{x,i}
+ \Big(\frac{6\eta\mu_y}{\nu}-\gamma \beta\Big) \notag \\&
\times \widetilde{\bd{m}}^c_{y,i}
+ \Big( -\frac{\mu_x}{4} + \frac{6 \kappa^2 \eta\mu^2_x}{\nu \mu_y} +4 \sigma^2_h \mu^2_x\gamma +\frac{2L^2_hN_1^2\gamma\mu^4_x}{\beta}
 \Big)
 \notag \\
&\times \|\bd{m}^c_{x,i}\|^2+(- \pi_1\mu_y\eta +4\sigma^2_h\mu^2_y\gamma +\frac{2L^2_hN_1^2\gamma\mu^4_y}{\beta})\|\bd{m}^c_{y,i}\|^2
 \notag\\
&
+ 4\gamma \beta^2\sigma^2
\Big] 
\end{align}
In the following, we will
link \eqref{TheoremStep4} with Lemma \ref{Metrics}. To do this, we choose
\begin{align}
&(s1) \quad   \mu_y L^2_f -\frac{\eta \mu_y\nu }{4} = -\frac{\mu_yL^2_f}{4} 
  \Longrightarrow 
  \eta =
  \frac{5L^2_f}{\nu} \label{stepsize:choise:start}
    \\
&(s2) \quad  \gamma = \frac{C}{\mu_y}
\end{align}
\cblack{where} $C$ is a {\em constant}  to be determined.
Substituting the 
expressions of $\eta, \gamma$
into \eqref{TheoremStep4},
we obtain
\begin{align} \label{TheoremStep5}
&\bd{\Omega}_{i+1} - \bd{\Omega}_{i} \notag \\
\le&  
\mbE \Big[  -\frac{\mu_y L^2_f}{4} \widetilde{\bd{b}}_{i}+ \Big(\mu_y 
- 
\frac{C\beta}{\mu_y}\Big)\widetilde{\bd{m}}^c_{x,i}
+ \Big(30\mu_y \kappa^2- \frac{C\beta}{\mu_y}\Big)\widetilde{\bd{m}}^c_{y,i} \notag \\& 
+ \Big( -\frac{\mu_x}{4} + 30\kappa^4\frac{\mu^2_x}{\mu_y} + \frac{4 C\sigma^2_h \mu^2_x}{\mu_y}+\frac{2CL^2_hN_1^2\mu^4_x}{\beta\mu_y}
 \Big)\|\bd{m}^c_{x,i}\|^2
 \notag \\
& +(- \frac{5\pi_1\mu_yL^2_f}{\nu} +4C\sigma^2_h\mu_y+\frac{2CL^2_hN_1^2\mu^3_y}{\beta})\|\bd{m}^c_{y,i}\|^2
 \notag\\
&
+ 4\gamma \beta^2\sigma^2
\Big] 
\end{align}
\cblack{We} further choose 
\begin{align}
&(s3) \quad \frac{2CL^2_hN_1^2\mu^3_y}{\beta} \le     4C\sigma^2_h\mu_y \Longrightarrow \mu_y \le \frac{\sigma_h \sqrt{2\beta}}{L_hN_1}   \\
&(s4) \quad 
\frac{5\pi_1\mu_yL^2_f}{\nu}\ge
8C\sigma^2_h\mu_y  \Longrightarrow
C \le 
\frac{5\pi_1L^2_f}{8\nu \sigma^2_h } \\
&(s5) \quad \mu_y - \frac{C\beta}{\mu_y}\le -\mu_y \Longrightarrow  \mu_y \le \sqrt{\frac{C\beta}{2}}\\
&(s6) \quad 
30\mu_y \kappa^2 \le \frac{C\beta}{\mu_y} \Longrightarrow  
\mu_y \le \sqrt{\frac{C \beta}{30\kappa^2}}  
\end{align}
\cblack{so that} 
\eqref{TheoremStep5}
can be written as 
\begin{align} \label{TheoremStep6}
&\bd{\Omega}_{i+1} - \bd{\Omega}_{i} \notag \\
\le&  
\mbE \Big[  -\frac{\mu_yL^2_f}{4} \widetilde{\bd{b}}_{i} -\mu_y \widetilde{\bd{m}}^c_{x,i}- \Big( \frac{\mu_x}{4} - 30\kappa^4\frac{\mu^2_x}{\mu_y}-
  \\& 
 \frac{4 C\sigma^2_h \mu^2_x}{\mu_y}-\frac{2CL^2_hN_1^2\mu^4_x}{\beta\mu_y}
 \Big)\|\bd{m}^c_{x,i}\|^2
+ 4\gamma \beta^2\sigma^2
\Big] 
\end{align}
\cblack{Next}, we deduce \cblack{a} step size condition \cblack{to ensure that} 
\begin{align} \label{stepsize}
   \frac{\mu_x}{4} - 30\kappa^4\frac{\mu^2_x}{\mu_y}- 
\frac{4 C\sigma^2_h \mu^2_x}{\mu_y}-\frac{2CL^2_hN_1^2\mu^4_x}{\beta\mu_y}  \ge \frac{\mu_x}{8}
\end{align}
Equation \eqref{stepsize}
is equivalent to 
\begin{align}
\label{stepcondition2}
    \frac{1}{8}
    &\ge 
    30 \kappa^4 \frac{\mu_x}{\mu_y} + \frac{4C\sigma^2_h \mu_x}{\mu_y} +\frac{2CL^2_hN_1^2\mu^3_x}{\beta\mu_y}  
\end{align}

Since
we  always choose $\mu_y$ so that 
$(s5)$ holds, using 
$
\frac{2\mu^2_y}{C} \le  \beta$,
\cblack{then} condition \eqref{stepcondition2} can be guaranteed by 
letting 
\begin{align}
\label{stepcodnition3}
    \frac{1}{8}
    &\ge 
    30 \kappa^4 \frac{\mu_x}{\mu_y} + \frac{4C\sigma^2_h \mu_x}{\mu_y} +\frac{C^2L^2_hN_1^2\mu^3_x}{\mu^3_y}  
\end{align}
The right-hand side  (RHS)  of 
\eqref{stepcodnition3}
is an upper bound for the RHS of 
\eqref{stepcondition2}.
It is \cblack{observed} that \eqref{stepcodnition3} is easily satisfied by
choosing $\mu_x \ll \mu_y$.
To move forward, we let 
\begin{align}
 &(s7) \quad \frac{C^2L^2_hN_1^2\mu^3_x}{\mu^3_y}  \le 
 \frac{4C\sigma^2_h \mu^2_x}{\mu^2_y}  \Longrightarrow
 C \le \frac{4\sigma^2_h \mu_y}{N_1^2 L^2_h \mu_x}   \\
 &(s8) \quad
 30 \kappa^4 \frac{\mu_x}{\mu_y} \le \frac{1}{16}
 \Longrightarrow \mu_x \le  \frac{\mu_y}{480\kappa^4}  
\end{align}
Since $\mu_y \ge \mu_x$,
$(s7)$ can be guaranteed by having
$C \le \frac{4\sigma^2_h}{M^2L^2_h}$. Therefore,
\eqref{stepcodnition3}
can be guaranteed by letting 
\begin{align}
\label{stepcodnition4}
(s9) \quad \frac{1}{16} \ge  \frac{8C \sigma^2_h \mu_x}{\mu_y} \Longrightarrow
    C \le \frac{\mu_y}{128 \sigma^2_h\mu_x}
\end{align}
Since $\mu_y \ge \mu_x$, condition
\eqref{stepcodnition4}
is guaranteed by having
$C \le \frac{1}{128 \sigma^2_h}$.
In order to guarantee the aforementioned conditions, we
choose the constant $C$  as follows:
\begin{align}
(s10) \quad     C = \min \Big\{\frac{5\pi_1L^2_f}{8\nu \sigma^2_h }, \frac{4\sigma^2_h}{N_1^2L^2_h}, \frac{1}{128 \sigma^2_h}\Big\} \label{stepsize:choice:last}
\end{align}
Note that $C$ can be chosen smaller than the values specified above; however, the stability range of  $\mu_y$ explicitly depends on $C$, 
as shown in conditions $(s5)$--$(s6)$. Therefore, to maximize the allowable range of step sizes, it is preferable to set  $C$ as large as possible.
Using $(s8)$ and $(s10)$ 
and $-\mu_y \le -\mu_x$, \cblack{relation}
\eqref{TheoremStep6} becomes
\begin{align} \label{TheoremStep7}
&\bd{\Omega}_{i+1} - \bd{\Omega}_{i} \notag \\
\le&  
\mbE \Big[  -\frac{\mu_xL^2_f}{4} \widetilde{\bd{b}}_{i} -\mu_x \widetilde{\bd{m}}^c_{x,i}- \frac{\mu_x}{8}\|\bd{m}^c_{x,i}\|^2
+ 4\gamma \beta^2\sigma^2
\Big] 
\end{align}
Now we have established the descent relation 
for the potential function $\bd{\Omega}_{i+1}$.
Averaging \cblack{the} above inequality over time we deduce that 
\begin{align}
\label{potentialresult}
& \frac{1}{T}\sum_{i=0}^{T-1}\mbE \Big[  \frac{L^2_f}{4} \widetilde{\bd{b}}_{i} + \widetilde{\bd{m}}^c_{x,i}+ \frac{1}{8}\|\bd{m}^c_{x,i}\|^2
\Big]  \notag  \\  
&\le \frac{1}{T\mu_x}\sum_{i=0}^{T-1}
(\bd{\Omega}_{i} - \bd{\Omega}_{i+1})+ \frac{4\gamma \beta^2\sigma^2}{\mu_x} \notag \\
&\overset{(a)}{\le} \frac{\bd{\Omega}_{0}
-P^\star}{T\mu_x} + \frac{4\gamma \beta^2\sigma^2}{\mu_x}
\end{align}
where $(a)$ follows from Assumption \ref{boundPhi} and telescoping results of the potential function and $-\bd{\Omega}_T \le -P(\bx_{T}) \le -P^\star$.
Finally, we have
\begin{align}
\label{finalPbound}
    &\frac{1}{T}
    \sum_{i=0}^{T-1}
    \mbE\|\nabla P(\bd{x}_i)\| \notag\\
           \overset{(a)}{\le}&  \frac{1}{T}
    \sum_{i=0}^{T-1}
    \mbE[L_f \| \bd{y}^o(\bd{x}_i) - \bd{y}_i\|
    +\|\nabla_x J(\bd{x}_i, \bd{y}_i) - \bd{m}^c_{x,i}\| 
   \notag  
    \\
   & +\|\bd{m}^c_{x, i}\|]  \notag 
\\
\le& 
\sqrt{
\Big(\frac{1}{T}
    \sum_{i=0}^{T-1}
    \mbE[L_f \| \bd{y}^o(\bd{x}_i) - \bd{y}_i\|
    +\|\nabla_x J(\bd{x}_i, \bd{y}_i) - \bd{m}^c_{x,i}\| }
   \notag   \\
   & \overline{+\|\bd{m}^c_{x, i}\|]\Big)^2}
 \notag \\
\overset{(b)}{\le}&
\sqrt{
\frac{1}{T}
    \sum_{i=0}^{T-1}\Big(
    \mbE\Big[L_f \| \bd{y}^o(\bd{x}_i) - \bd{y}_i\|
    +\|\nabla_x J(\bd{x}_i, \bd{y}_i) - \bd{m}^c_{x,i}\| }
   \notag   \\
   & \overline{+\|\bd{m}^c_{x, i}\|\Big]\Big)^2}
\notag \\
\overset{(c)}{\le}&
\sqrt{
\frac{1}{T}
    \sum_{i=0}^{T-1} 3
    ( \mbE L_f \| \bd{y}^o(\bd{x}_i) - \bd{y}_i\|)^2}
   \notag   \\
   &  \overline{+3(\mbE\|\nabla_x J(\bd{x}_i, \bd{y}_i) - \bd{m}^c_{x,i}\|)^2+3(\mbE\|\bd{m}^c_{x, i}\|)^2}
\notag\\
  \overset{(d)}{\le}&
   \sqrt{ 
   \frac{3}{T}
    \sum_{i=0}^{T-1}
    \mbE[L^2_f\widetilde{\bd{b}}_{i} +  \widetilde{\bd{m}}^c_{x,i}+\|\bd{m}^c_{x, i}\|^2]
   } \notag    
 \\
   \le&
    \sqrt{ 
   \frac{1}{T}
    \sum_{i=0}^{T-1}
    \mbE[6L^2_f\widetilde{\bd{b}}_{i} + 24 
\widetilde{\bd{m}}^c_{x,i}+3\|\bd{m}^c_{x, i}\|^2]
   } \notag \\
   \overset{(e)}{\le}&
  \sqrt{ \frac{24(\bd{\Omega}_{0}
-P^\star)}{T\mu_x} + \frac{96\gamma \beta^2\sigma^2}{\mu_x}} \notag \\
 \le&  \sqrt{ \frac{24(\bd{\Omega}_{0}
-P^\star)}{T\mu_x}} +\sqrt{\frac{96\gamma \beta^2\sigma^2}{\mu_x}}
\end{align}
where $(a)$ follows from Lemma \ref{Metrics}, $(b)$, $(c)$ and $(d)$ follow from Jensen's inequality and convexity of the quadratic function,
$(e)$ follows from \eqref{potentialresult}.
To summarize,
the stability condition for the hyperparameters in \textbf{HCMM-1} is given by
\begin{align}
 \label{proof_hyperparamter_condition1}
\beta_x &= \beta_y \le \frac{1}{2} 
\\
 \mu_y &\le  \min \Big\{ \frac{\sigma_h \sqrt{2\beta_y}}{L_h N_1}, \sqrt{\frac{C\beta_y}{2}}, \sqrt{\frac{C\beta_y}{30\kappa^2}}, \frac{2}{\nu}, \pi_1\Big\}, 
  \label{proof_hyperparamter_condition2}\\
\mu_x &\le  \min \Big\{\mu_y, \frac{1}{480\kappa^4}\mu_y, \frac{1}{2L_1}\Big\}  
 \label{proof_hyperparamter_condition3}
\end{align}
where 
$\kappa = \frac{L_f}{\nu}, L_1 =L_f +\kappa L_f$, while
 $C, \pi_1$ are constants given by
\begin{align}
C = \min \Big\{\frac{5\pi_1L^2_f}{8\nu \sigma^2_h }, \frac{4\sigma^2_h}{N_1^2L^2_h}, \frac{1}{128 \sigma^2_h}\Big\}, \quad \pi_1 = \frac{1}{2L_f+\nu} \notag
\end{align} 
We further choose the smoothing factors as $\beta_x = \beta_y = \mathcal{O}(\frac{1}{T^{2/3}})$, and 
$\mu_x = c_1\sqrt{\beta_x},  \mu_y = c_2\sqrt{\beta_y}$
for some small constants $c_1 < c_2$. Accordingly, all the aforementioned conditions can be satisfied for {\em sufficiently} large $T$. 
Initializing $\bm^c_{x,0}$ using a minibatch such that $\bd{\Omega}_0 = \mathcal{O}(1)$ and
the convergence rate of \textbf{HCMM-1} is given by 
\begin{align}
   \frac{1}{T}
    \sum_{i=0}^{T-1}
    \mb{E}\|\nabla P(\bd{x}_i)\| 
    \le \mc{O}\Big(\frac{1}{T^{1/3}}\Big)
\end{align}
\qed
\vspace{-1em}
\section{PROOF OF COROLLARY 1 FOR HCMM-1}
\label{appendices:corollary}
Similar to the proof of  
Theorem \ref{Maintheorem},
we construct a new potential function as follows:
\begin{align} \label{potential2}
\bd{\Omega}_{i}
&= \mathbb{E}
\Big[P(\bd{x}_i) 
+ \eta \Delta_i + \gamma
\|\bd{m}^c_{x,i} - \nabla_x J(\bd{x}_{i}, \bd{y}_{i})\|^2
\notag \\&  \quad  +\gamma\|\bd{m}^c_{y,i} - \nabla_y J(\bd{x}_{i}, \bd{y}_{i})\|^2
\Big]
\end{align}
where $\Delta_i \triangleq P(\bx_{i}) -J(\bx_{i}, \by_{i}) \ge0$.
Subtracting $\bd{\Omega}_{i}$
from $\bd{\Omega}_{i+1}$
and recall the definitions 
\eqref{proof_hcmm1_definition_mcxi},
\eqref{proof_hcmm1_definition_mcyi} 
for 
$\widetilde{\bd{m}}^c_{x,i}$
and 
$\widetilde{\bd{m}}^c_{y,i}$,
we get 
\begin{align}
&\bd{\Omega}_{i+1}
-\bd{\Omega}_{i} \notag \\
&=\mbE
\Big[
(P(\bx_{i+1}) - P(\bx_{i}))
+\eta(\Delta_{i+1} - \Delta_{i})
\notag \\
&\quad +\gamma(\widetilde{\bm}^c_{x,i+1} - \widetilde{\bm}^c_{x,i})+\gamma(\widetilde{\bm}^c_{y,i+1} - \widetilde{\bm}^c_{y,i})
\Big] \notag \\
&\overset{(a)}{\le}\mbE \Big[
-\frac{\mu_x}{2}
\|\nabla P(\bd{x}_i)\|^2
-\frac{\mu_x}{2}(1-L_2\mu_x)
\|\bd{m}^c_{x,i}\|^2 \notag \\
&\quad 
+\frac{2\mu_xL^2_f}{\nu} \Delta_i
+ \mu_x \widetilde{\bm}^c_{x,i}
+\eta \Big(
-\frac{\nu\mu_y}{2}
\Delta_i
+ \frac{2\kappa^2\mu^2_x}{\mu_y}
\|\bm^c_{x,i}\|^2
\notag \\
&\quad 
-\frac{\mu_y}{2}
(1-L_f\mu_y)\|\bm^c_{y,i}\|^2+\mu_y\mu^2_x L^2_f\|\bm^c_{x,i}\|^2 + \mu_y \widetilde{\bm}^c_{y,i}\Big)
\notag\\
&\quad+ \gamma
\Big(
-\beta_x\widetilde{\bm}^c_{x,i}
-\beta_y \widetilde{\bm}^c_{y,i}
+ (\frac{L^2_h}{2\beta_x} +\frac{L^2_h}{2\beta_y})  \notag \\ &\times
\Big\|\begin{bmatrix}
    \bd{x}_{i+1} - \bd{x}_i \\
    \bd{y}_{i+1} - \bd{y}_i
\end{bmatrix}\Big\|^4
+ 2((1-\beta_x)^2 +(1-\beta_y)^2) \sigma^2_h
\notag \\&
\times 
\Big\|\begin{bmatrix}
    \bd{x}_{i+1} - \bd{x}_i \\
    \bd{y}_{i+1} - \bd{y}_i
\end{bmatrix}\Big\|^2 
+ 2 (\beta^2_x+\beta^2_y) \sigma^2
\Big)
\Big]
\end{align}
where $(a)$
follows from Lemma 
\ref{GradientIncremental},
\ref{proof_hcmm1_costvalue_gap}, and \ref{proof_lemma_hcmm1_optigap}.
Setting 
$\beta_x = \beta_y = \beta$
and using the relation 
\eqref{proof_eq_hcmm1_order},
we can simplify the above inequality as 
\begin{align}
&\bd{\Omega}_{i+1}
-\bd{\Omega}_{i} \notag \\
&\le\mbE \Big[
-\frac{\mu_x}{2}
\|\nabla P(\bd{x}_i)\|^2
-\frac{\mu_x}{2}(1-L_2\mu_x)
\|\bd{m}^c_{x,i}\|^2 \notag \\
&\quad 
+\frac{2\mu_xL^2_f}{\nu} \Delta_i
+ \mu_x \widetilde{\bm}^c_{x,i}
+\eta \Big(
-\frac{\nu\mu_y}{2}
\Delta_i
+ \frac{2\kappa^2\mu^2_x}{\mu_y}
\|\bm^c_{x,i}\|^2
\notag \\
&\quad 
-\frac{\mu_y}{2}
(1-L_f\mu_y)\|\bm^c_{y,i}\|^2+\mu_y\mu^2_x L^2_f\|\bm^c_{x,i}\|^2+\mu_y \widetilde{\bm}^c_{y,i}\Big)
\notag\\
&\quad+ \gamma
\Big(
-\beta\widetilde{\bm}^c_{x,i}
-\beta \widetilde{\bm}^c_{y,i}
+ \frac{L^2_h}{\beta}
(2\mu^4_x \|\bm^c_{x,i}\|^4 + 2\mu^4_y\|\bm^c_{y,i}\|^4)\notag \\ &
\quad + 4(1-\beta)^2 \sigma^2_h(\mu^2_x\|\bm^c_{x,i}\|^2+\mu^2_y\|\bm^c_{y,i}\|^2)
+ 4 \beta^2 \sigma^2
\Big)
\Big] \notag \\
&\le 
\mbE \Big[
-\frac{\mu_x}{2}
\|\nabla P(\bx_{i})\|^2
+\Big( 
\frac{2\mu_x L^2_f}{\nu}
-\frac{\eta\nu\mu_y}{2}
\Big)\Delta_i \notag \\
&\quad +
(\mu_x - \gamma\beta)\widetilde{\bm}^c_{x,i}+
(\eta\mu_y - \gamma \beta)
\widetilde{\bm}^c_{y,i}  
\notag \\
&\quad -\Big(\frac{\mu_x}{2}- \frac{L_2\mu^2_x}{2}-\frac{2\eta\kappa^2\mu^2_x}{\mu_y}-\eta\mu_y\mu^2_xL^2_f -\frac{2L^2_h\gamma\mu^4_xN^2_1}{\beta}
\notag 
\\ 
&\quad -4\gamma\mu^2_x \sigma^2_h\Big)
\|\bm^c_{x,i}\|^2 - \Big(
\frac{\eta\mu_y}{2}(1-L_f\mu_y)
-\frac{2L^2_h\gamma\mu^4_yN^2_1}{\beta}
\notag \\
&\quad -4\gamma\mu^2_y\sigma^2_h\Big)
\|\bm^c_{y,i}\|^2
+4\beta^2\gamma\sigma^2
\Big]
\end{align}
In the following, 
we choose appropriate 
coefficients 
$\eta, \gamma$
to cancel out these stochastic quantities.
First, choosing $\mu_x \le \mu_y$
and knowing the fact that $\Delta_i, \widetilde{\bm}^c_{x,i}, \widetilde{\bm}^c_{y,i}$
are nonnegative,
we  get 
\begin{align}
&\bd{\Omega}_{i+1}
-\bd{\Omega}_{i} \notag \\
&\le 
\mbE \Big[
-\frac{\mu_x}{2}
\|\nabla P(\bx_{i})\|^2
+\Big( 
\frac{2\mu_y L^2_f}{\nu}
-\frac{\eta\nu\mu_y}{2}
\Big)\Delta_i \notag \\
&\quad +
(\mu_y - \gamma\beta)\widetilde{\bm}^c_{x,i}+
(\eta\mu_y - \gamma \beta)
\widetilde{\bm}^c_{y,i}  
\notag \\
&\quad -\Big(\frac{\mu_x}{2}- \frac{L_2\mu^2_x}{2}-\frac{2\eta\kappa^2\mu^2_x}{\mu_y}-\eta\mu_y\mu^2_xL^2_f -\frac{2L^2_h\gamma\mu^4_xN^2_1}{\beta} 
\notag 
\\ 
&\quad -4\gamma\mu^2_x \sigma^2_h\Big)
\|\bm^c_{x,i}\|^2 - \Big(
\frac{\eta\mu_y}{2}(1-L_f\mu_y)
-\frac{2L^2_h\gamma\mu^4_yN^2_1}{\beta}
\notag \\
&\quad -4\gamma\mu^2_y\sigma^2_h\Big)
\|\bm^c_{y,i}\|^2
+4\beta^2\gamma\sigma^2
\Big]
\end{align}
We proceed by choosing
\begin{align}
&(s1) \quad       \frac{2\mu_yL^2_f}{\nu} - \frac{\eta \nu \mu_y}{2} = 0 \Longrightarrow \eta = 4\kappa^2
\end{align}
The above inequality becomes
\begin{align}
&\bd{\Omega}_{i+1} - \bd{\Omega}_{i} \notag \\
&\le 
\mbE
\Big[
-\frac{\mu_x}{2}
\|\nabla P(\bx_i)\|^2
+(\mu_y - \gamma \beta)
\widetilde{\bm}^c_{x,i}
+(4\mu_y\kappa^2-\gamma\beta)
\notag \\
&\times \widetilde{\bm}^c_{y,i} 
-\Big(\frac{\mu_x}{2}- \frac{L_2\mu^2_x}{2}-\frac{8\kappa^4\mu^2_x}{\mu_y}-4\mu_y\mu^2_x\kappa^2L^2_f -\frac{2L^2_h\gamma\mu^4_xN^2_1}{\beta}
\notag \\
&-4\gamma\mu^2_x \sigma^2_h\Big) \|\bm^c_{x,i}\|^2
-\Big(
2\mu_y\kappa^2(1-L_f\mu_y)
-\frac{2L^2_h\gamma\mu^4_yN^2_1}{\beta}
\notag \\
&-4\gamma\mu^2_y\sigma^2_h
\Big)\|\bm^c_{y,i}\|^2
+4\beta^2\gamma\sigma^2
\Big 
]
\end{align}
Moreover,
we choose 
\begin{align}
(s2)
\quad \gamma = \frac{C}{\mu_y}
\end{align}
where $C$ is a constant to be determined later.
To cancel out 
$\widetilde{\bm}^c_{x,i}$
and 
$\widetilde{\bm}^c_{y,i}$, 
the step size $\mu_x$ needs to satisfy
\begin{align}
&(s3) \quad 
\mu_y - \frac{C \beta}{\mu_y} \le 0 \Longrightarrow
\mu_y \le  \sqrt{C\beta}  \\
&(s4) \quad 
4\mu_y\kappa^2 -\frac{C\beta}{\mu_y} \le 0 \Longrightarrow\mu_y \le \frac{\sqrt{C\beta}}{2\kappa}
\end{align}
We then get 
\begin{align}
&\bd{\Omega}_{i+1} - \bd{\Omega}_{i} \notag \\
&\le 
\mbE
\Big[
-\frac{\mu_x}{2}
\|\nabla P(\bx_i)\|^2-\Big(\frac{\mu_x}{2}- \frac{L_2\mu^2_x}{2}-\frac{8\kappa^4\mu^2_x}{\mu_y}
\notag \\ 
& \quad 
-4\mu_y\mu^2_x\kappa^2L^2_f -\frac{2L^2_hC\mu^4_xN^2_1}{\beta\mu_y}-\frac{4C\mu^2_x \sigma^2_h}{\mu_y}\Big) \|\bm^c_{x,i}\|^2
\notag \\
&\quad
-\Big(
2\mu_y\kappa^2(1-L_f\mu_y)
-\frac{2L^2_h C\mu^3_yN^2_1}{\beta}-4C\mu_y\sigma^2_h
\Big)\|\bm^c_{y,i}\|^2
\notag \\
& \quad 
+4\beta^2\gamma\sigma^2
\Big 
]
\end{align}
Furthermore, 
we need to choose step sizes $\mu_x$ and $\mu_y$
such that 
\begin{align}
&\frac{\mu_x}{2}
-\frac{L_2\mu^2_x}{2}
-\frac{8\kappa^4\mu^2_x}{\mu_y}
\notag \\
&-4\mu_y\mu^2_x\kappa^2L^2_f-\frac{2L^2_hC\mu^4_xN^2_1}{\beta\mu_y} - \frac{4C\mu^2_x\sigma^2_h}{\mu_y}\ge 0
\label{proof_ncpl_hcmm1_inequality1}
\end{align}
and
\begin{align}
&2\kappa^2\mu_y-2L_f\kappa^2\mu^2_y
-\frac{2L^2_h C\mu^3_yN^2_1}{\beta} - 4C\mu_y\sigma^2_h \ge 0
\label{proof_ncpl_hcmm1_inequality2}
\end{align}
It is evident that 
the above two inequalities can be satisfied by choosing sufficiently small step sizes $\mu_x, \mu_y$
and a small enough constant $C$.
We first establish the first inequality by choosing  appropriate step sizes 
to upper bound the
terms that appear after $\frac{\mu_x}{2}$
\begin{align}
&(s5) \quad 
\frac{L_2\mu^2_x}{2} \le \frac{\mu_x}{8} \Longrightarrow
\mu_x \le \frac{1}{4L_2}
\\
&(s7) \quad 
\frac{8\kappa^4\mu^2_x}{\mu_y}
\le \frac{\mu_x}{8} \Longrightarrow
\frac{\mu_x}{\mu_y} \le \frac{1}{64\kappa^4}  \\
&(s8) \quad 
4\mu_y\mu^2_x\kappa^2L^2_f \le \frac{\mu_x}{16} \Longrightarrow
\mu_y  \le \frac{1}{8\kappa L_f}
\\
&(s9) \quad \frac{2L^2_hC\mu^4_xN^2_1}{\beta\mu_y} \le \frac{4C\mu^2_x\sigma^2_h}{\mu_y} \Longrightarrow
\mu_x \le  \frac{\sqrt{2\beta\sigma^2_h}}{L_hN_1}  
\end{align}
Therefore, 
the first inequality 
is satisfied when $\frac{\mu_x}{2}$
is greater than the upper bound of the subtrahend, i.e.,
\begin{align}
    \frac{\mu_x}{2}
    \ge  \frac{5\mu_x}{16} + \frac{8C\mu^2_x\sigma^2_h}{\mu_y}
\end{align}
the above relation is satisfies 
when 
\begin{align}
C \le \frac{3\mu_y}{128\mu_x\sigma^2_h}
\end{align}
since $\mu_y \ge \mu_x$,
the above condition is guaranteed by letting
\begin{align}
(s10) \quad C \le \frac{3}{128\sigma^2_h}
\end{align}
We proceed to establish the second inequality.
To upper bound the subtrahend 
appear in 
\eqref{proof_ncpl_hcmm1_inequality2},
we choose 
\begin{align}
&(s11) \quad 2L_f\kappa^2\mu^2_y \le \frac{\kappa^2\mu_y}{2} \Longrightarrow \mu_y \le \frac{1}{4L_f} \\
&(s12) \quad 
\frac{2L^2_hC\mu^3_yN^2_1}{\beta} \le 4C\mu_y\sigma^2_h \Longrightarrow \mu_y  \le \frac{\sqrt{2\beta\sigma^2_h}}{L_hN_1}
\end{align}
Therefore, the second inequality is satisfied when 
$2\kappa^2\mu_y$
is greater than the upper bound of the subtrahend, i.e., 
\begin{align}
(s13) 
\quad 2\kappa^2\mu_y \ge 
    \frac{\kappa^2\mu_y}{2} + 8C\mu_y\sigma^2_h 
&\Longrightarrow
C \le \frac{3\kappa^2}{16\sigma^2_h} 
\end{align}
Finally,
$C$ can be chosen as 
\begin{align}
C = \min \Big\{ \frac{3\kappa^2}{16\sigma^2_h},
\frac{3}{128\sigma^2_h}\Big\}
\end{align}
we then obtain 
\begin{align}
&\boldsymbol{\Omega}_{i+1} -\bd{\Omega}_{i}
\notag \\
&\le 
\mbE\Big[
-\frac{\mu_x}{2}
\|\nabla P(\bx_i)\|^2
+4\beta^2\gamma\sigma^2
\Big]
\end{align}
Rearrange the above term and averaging the inequality over iterations,
we get 
\begin{align}
    &\frac{1}{T}\sum_{i=0}^{T-1}\mbE\|\nabla P(\bx_i)\|^2 \le \frac{2(\bd{\Omega}_0 - P^\star)}{\mu_x T}
+ 8\beta^2\gamma\sigma^2
\end{align}
Using Jensen's inequality
for quadratic and square root function, we get 
\begin{align}
&\frac{1}{T}\sum_{i=0}^{T-1}
\mbE\|\nabla P(\bx_i)\| \notag \\
&= 
\frac{1}{T}\sum_{i=0}^{T-1}
\sqrt{(\mbE\|\nabla P(\bx_i)\|)^2}
\le 
\frac{1}{T}\sum_{i=0}^{T-1}
\sqrt{\mbE\|\nabla P(\bx_i)\|^2} \notag \\
&\le 
\sqrt{\frac{1}{T}\sum_{i=0}^{T-1}\mbE\|\nabla P(\bx_i)\|^2} \le \sqrt{\frac{2(\bd{\Omega}_0 - P^\star)}{\mu_x T}
+ 8\beta^2\gamma\sigma^2} \notag \\
&\le \sqrt{\frac{2(\bd{\Omega}_0 - P^\star)}{\mu_x T}}
+ \sqrt{8\beta^2\gamma\sigma^2}
\end{align}
To summarize, the stability condition for the hyperparameters in \textbf{HCMM-1} is given by
and 
\begin{align}
 \mu_y &\le  \min \Big\{ \frac{\sqrt{C\beta}}{2\kappa}, \frac{2\kappa^2}{L+L_f}, \frac{\sqrt{2\beta\sigma^2_h}}{L_hN_1}, \frac{1}{8\kappa L_f}, \frac{1}{4L_f}, \frac{1}{\nu}\Big\},  \label{nonconvexsc_bound1}\\
\mu_x &\le  \min \Big\{\mu_y, \frac{\mu_y}{64\kappa^4}, \frac{1}{4L_2}, \frac{\sqrt{2\beta\sigma^2_h}}{L_h N_1}\Big\}  
\label{nonconvexsc_bound2}
\end{align}
We further choose the smoothing factors as $\beta_x = \beta_y = \mathcal{O}(\frac{1}{T^{2/3}})$, and 
$\mu_x = c_1\sqrt{\beta_x},  \mu_y = c_2\sqrt{\beta_y}$
for some small constants $c_1 < c_2$. Accordingly, all the aforementioned conditions can be satisfied for {\em sufficiently} large $T$. Initializing $\bm^c_{x,0}$ using a minibatch such that $\bd{\Omega}_0 = \mathcal{O}(1)$ and
the convergence rate of \textbf{HCMM-1} is given by 
\begin{align}
   \frac{1}{T}
    \sum_{i=0}^{T-1}
    \mb{E}\|\nabla P(\bd{x}_i)\| 
    \le \mc{O}\Big(\frac{1}{T^{1/3}}\Big)
\end{align}
\qed
\vspace{-2em}
\section{PROOF OF THEOREM \ref{Maintheorem2} FOR HCMM-2}
\label{proof_Maintheorem2}
From \eqref{primal_ineq},
we have 
\begin{align}
   &\|\nabla P(\bd{x}_i)\| \le 
   \frac{(P(\bd{x}_i) - P(\bd{x}_{i+1}))}{3\mu_x} + 9\|\bd{m}_{x,i} - \nabla_x J(\bd{x}_i, \bd{y}_i)\| \notag \\
   &\quad + 9L_f\|\bd{y}_i - \bd{y}^o(\bd{x}_i)\| +\frac{3L_2\mu_x}{2}
\end{align}
Averaging \cblack{the} above inequality over iterations and taking expectation, we get 
\begin{align}
\label{average_P_ncpl}
    &\frac{1}{T}\sum_{i=0}^{T-1}\mbE\|\nabla P(\bd{x}_i)\| \notag \\
   &\le 
   \frac{(P(\bd{x}_0) - P^\star)}{3\mu_xT} + \frac{9}{T}\sum_{i=0}^{T-1}\mbE\|\bd{m}_{x,i} - \nabla_x J(\bd{x}_i, \bd{y}_i)\| \notag \\
   &\quad + \frac{9L_f}{T}\sum_{i=0}^{T-1}\mbE\|\bd{y}_i - \bd{y}^o(\bd{x}_i)\| +\frac{3L_2\mu_x}{2}
\end{align}
Invoking Lemmas \ref{deviation_ncpl}
and \ref{dual_gap_ncpl}, we get 
\begin{align}
    &\frac{1}{T}\sum_{i=0}^{T-1} \mbE\|\nabla P(\bd{x}_i)\| \notag \\
   &\le  
   \frac{P(\bd{x}_0) - P^\star}{3\mu_xT}
   +(162\kappa+9)(\frac{\sigma}{T\beta_x}+\frac{L_h\mu^2_y}{\beta_x} \notag \\
   &\quad + \frac{2\mu_y\sigma_h}{\sqrt{\beta_x}} +\sigma \sqrt{\beta_x})+
      \frac{18L_f\|\bd{y}_{0} - \bd{y}^{o}(\bd{x}_{0})\|}{T} \notag \\
   &\quad + \frac{54\kappa \Delta_0}{\mu_y T} +(270\kappa L_2+18L_f)\mu_y +\frac{3L_2\mu_x}{2}
\end{align}
We further choose the smoothing factors 
as $\beta_x =\beta_y = \mathcal{O}\Big(\frac{1}{T^{2/3}}\Big)$
and $\mu_y = \mathcal{O}\Big(\frac{1}{T^{2/3}}\Big), \mu_x = c_3 \mu_y$ for a small constant $c_3 < 1$.
we get 
\begin{align}
    &\frac{1}{T}\sum_{i=0}^{T-1}\mbE\|\nabla P(\bd{x}_i)\| \le \mc{O}\Big( \frac{1}{T^{1/3}}\Big)
    + \mc{O}\Big( \frac{1}{T}\Big) +\mc{O}\Big( \frac{1}{T^{2/3}}\Big)
\end{align}
Therefore, the convergence rate is dominated by 
$\mc{O}(\frac{1}{T^{1/3}})$.

\qed
\vspace{-2em}
\bibliographystyle{IEEEbib}
\bibliography{refs}
\end{document}